\documentclass[11pt, a4paper, logo, copyright]{hail_lab}

\usepackage[utf8]{inputenc} 
\usepackage[T1]{fontenc}    
\usepackage{url}            
\usepackage{booktabs}       
\usepackage{amsfonts}       
\usepackage{nicefrac}       
\usepackage{microtype}      
\usepackage{xcolor}         

\usepackage{amsmath,amsfonts,bm}

\def\eqref#1{equation~\ref{#1}}

\def\1{\bm{1}}

\DeclareMathAlphabet{\mathsfit}{\encodingdefault}{\sfdefault}{m}{sl}
\SetMathAlphabet{\mathsfit}{bold}{\encodingdefault}{\sfdefault}{bx}{n}

\usepackage{microtype}
\usepackage{graphicx}
\usepackage{subfig}
\usepackage{booktabs}
\usepackage{amsmath,bm}
\usepackage{amsthm}
\usepackage{cite}
\usepackage{natbib}
\usepackage{enumerate}
\usepackage{wrapfig}
\usepackage{color, soul}
\usepackage{psfrag}
\usepackage{adjustbox}
\usepackage{xspace}
\usepackage{amssymb}
\usepackage{multirow}
\usepackage{afterpage}
\usepackage{colortbl}
\usepackage{float}
\usepackage{textcomp}
\usepackage{siunitx}
\usepackage{mdframed}
\usepackage{makecell}
\usepackage[super]{nth}

\usepackage{enumitem}
\usepackage{bbding}
\usepackage{pifont}
\usepackage{bbm}
\usepackage{xcolor}
\usepackage{pdflscape}
\usepackage{theoremref}
\usepackage{pifont}
\usepackage{adjustbox}
\usepackage{xcolor}
\usepackage{mathtools}
\usepackage{subcaption}
\usepackage{pbox}
\usepackage{longtable}
\usepackage{listings}
\usepackage{lmodern}

\colorlet{darkgreen}{green!65!black}
\colorlet{darkblue}{blue!75!black}
\colorlet{darkred}{red!80!black}
\definecolor{statistical}{HTML}{8c564b}
\definecolor{structural}{HTML}{0070C0}
\definecolor{semantic}{HTML}{008080}
\definecolor{yellow}{HTML}{f7c600}
\definecolor{lightblue}{HTML}{0071bc}
\definecolor{lightgreen}{HTML}{39b54a}
\definecolor{deemph}{gray}{0.55}

\definecolor{baselinecolor}{gray}{.95}
\definecolor{graycolor}{gray}{.95}
\newcommand{\grayrow}{\rowcolor[gray]{.95}}

\usepackage[pagebackref=false, breaklinks=true, colorlinks, citecolor=lightblue, urlcolor=lightblue, bookmarks=false]{hyperref}

\newlength\savewidth
\newcolumntype{x}[1]{>{\centering\arraybackslash}p{#1pt}}
\newcolumntype{y}[1]{>{\raggedright\arraybackslash}p{#1pt}}
\newcolumntype{z}[1]{>{\raggedleft\arraybackslash}p{#1pt}}

\newcommand{\inc}[1]{\textcolor{darkgreen}{\textbf{\scriptsize{(+#1)}}}}
\newcommand{\dec}[1]{\textcolor{darkblue}{\textbf{\scriptsize{(-#1)}}}}

\newcommand{\simclr}{{SimCLR}~\cite{chen2020simple}}
\newcommand{\dino}{{DINO}~\cite{oquab2023dinov2}}

\newcommand{\ours}{\textsc{OSF}\xspace}
\newcommand{\benchname}{\texttt{SleepBench}\xspace}

\definecolor{textgreen}{RGB}{57, 172, 57}
\definecolor{textred}{RGB}{200, 10, 10}
\definecolor{boxyellow}{HTML}{FAF5E6}
\definecolor{frameyellow}{HTML}{B7950B}
\definecolor{boxpurple}{HTML}{F4EFF6}
\definecolor{framepurple}{HTML}{6C3483}
\definecolor{boxblue}{HTML}{EEF4F8}
\definecolor{frameblue}{HTML}{2874A6}
\definecolor{boxgray}{HTML}{F0F2F3}
\definecolor{framegray}{HTML}{5D6D7E}
\definecolor{boxgreen}{HTML}{EAFaf1}
\definecolor{framegreen}{HTML}{196F3D}
\usepackage{pifont} 
\newcommand{\cmark}{\textcolor{textgreen}{\ding{51}}}
\newcommand{\xmark}{\textcolor{textred}{\ding{55}}}

\usepackage[most]{tcolorbox}
\tcbset{
    center title,
    left=0pt,
    right=0pt,
    top=0pt,
    bottom=0pt,
    colframe=black,
    colback=gray!10,
    width=\linewidth,
    enlarge left by=0mm,
    boxsep=5pt,
    boxrule=1pt,
    arc=0pt,outer arc=0pt,
}

\newtcolorbox{promptbox}[1][]{
    enhanced,
    colback=white,
    colframe=black,
    fonttitle=\bfseries,
    title=Prompt,
    attach boxed title to top left={xshift=10pt, yshift*=-\tcboxedtitleheight/2},
    boxed title style={colback=black},
    top=12pt, bottom=10pt, left=10pt, right=10pt,
    #1
}

\tcbset{
    fancybox/.style={
        enhanced,
        breakable,
        arc=1mm,
        outer arc=1mm,
        boxrule=0pt,
        leftrule=1mm,
        toptitle=0.5mm,
        bottomtitle=0.5mm,
        fonttitle=\bfseries\sffamily\small,
        fontupper=\scriptsize,
        coltitle=white,
        drop shadow
    }
}

\newtcolorbox{thoughtbox}{
    fancybox,
    colback=boxyellow,
    colframe=frameyellow,
    coltitle=black,
    title=Thought
}
\newtcolorbox{userbox}{
    fancybox,
    colback=boxpurple,
    colframe=framepurple,
    title=User
}
\newtcolorbox{agentbox}{
    fancybox,
    colback=boxblue,
    colframe=frameblue,
    title=Agent
}
\newtcolorbox{outputbox}{
    fancybox,
    colback=boxgray, 
    colframe=framegray,
    coltitle=black,
    title=Execution Output
}
\newtcolorbox{solutionbox}{
    fancybox,
    colback=boxgreen,
    colframe=framegreen,
    title=Solution
}

\definecolor{codegreen}{rgb}{0.0, 0.5, 0.0}
\definecolor{codegray}{rgb}{0.4, 0.4, 0.4}
\definecolor{codepurple}{rgb}{0.50, 0, 0.50}
\definecolor{backcolour}{rgb}{0.97, 0.97, 0.97}

\lstdefinestyle{mystyle}{
    backgroundcolor=\color{backcolour},
    commentstyle=\color{codegreen},
    keywordstyle=\color{magenta},
    stringstyle=\color{codepurple},
    basicstyle=\ttfamily\scriptsize, 
    breakatwhitespace=false,
    breaklines=true,
    captionpos=b,
    keepspaces=true,
    numbers=none,              
    showspaces=false,
    showstringspaces=false,
    showtabs=false,
    tabsize=2,
    frame=single,
    rulecolor=\color{black!10}, 
    frameround=fttt,            
    upquote=true
}
\newcommand{\finding}[2]{%
  \begin{tcolorbox}[
    enhanced,
    breakable,
    width=\linewidth,
    colback=white!90!gray,
    colframe=white,        
    boxrule=0pt,           
    arc=5pt,
    boxsep=5pt,
    left=5pt,right=5pt,top=3pt,bottom=3pt,
    before skip=0.6\baselineskip,
    after skip=0.6\baselineskip,
  ]
  \noindent\textbf{\textit{Finding~#1:}}~#2
  \end{tcolorbox}%
}
\newcommand{\takeaway}[2]{%
  \begin{tcolorbox}[
    enhanced,
    breakable,
    width=\linewidth,
    colback=white!90!gray,
    colframe=white,        
    boxrule=0pt,           
    arc=5pt,
    boxsep=5pt,
    left=5pt,right=5pt,top=3pt,bottom=3pt,
    before skip=0.6\baselineskip,
    after skip=0.6\baselineskip,
  ]
  \noindent\textbf{\textit{Takeaway~#1:}}~#2
  \end{tcolorbox}%
}
\newcommand{\sleepfm}{{SleepFM}~\cite{thapa2026multimodal}}

\title{
\ours: On Pre-training and Scaling of Sleep Foundation Models
}

\author[1]{Zitao Shuai}
\author[1]{Zongzhe Xu}
\author[2]{David Yang}
\author[1]{Wei Wang}
\author[1$\dagger$]{Yuzhe Yang}

\affil[1]{University of California, Los Angeles}
\affil[2]{Emory University}
\correspondingauthor{ $^\dagger$Correspondence to: yuzhey@ucla.edu.}

\codelink{https://github.com/yang-ai-lab/OSF-Open-Sleep-FM}
\projecturl{https://yang-ai-lab.github.io/osf}

\begin{abstract}

Polysomnography (PSG) provides the gold standard for sleep assessment but suffers from substantial heterogeneity across recording devices and cohorts.
There have been growing efforts to build general-purpose foundation models (FMs) for sleep physiology, but lack an in-depth understanding of the \textit{pre-training} process and \textit{scaling} patterns that lead to more generalizable sleep FMs.
To fill this gap, we curate a massive corpus of 166,500 hours of sleep recordings from nine public sources and establish \benchname, a comprehensive, fully open-source benchmark.
Leveraging \benchname, we systematically evaluate four families of self-supervised pre-training objectives and uncover three critical findings:
(1) existing FMs fail to generalize to missing channels at inference;
(2) channel-invariant feature learning is essential for pre-training;
and (3) scaling sample size, model capacity, and multi-source data mixture consistently improves downstream performance.
With an enhanced pre-training and scaling recipe, we introduce \ours, a family of sleep FMs that achieves state-of-the-art performance across nine datasets on diverse sleep and disease prediction tasks.
Further analysis of \ours also reveals intriguing properties in sample efficiency, hierarchical aggregation, and cross-dataset scaling. Codes are available at: \url{https://github.com/yang-ai-lab/OSF-Open-Sleep-FM}.

\end{abstract}

\begin{document}

\maketitle

\newenvironment{Itemize}{
    \begin{itemize}[leftmargin=*]
    \setlength{\itemsep}{0pt}
    \setlength{\topsep}{0pt}
    \setlength{\partopsep}{0pt}
    \setlength{\parskip}{1pt}}
{\end{itemize}}
\setlength{\leftmargini}{9pt}

\vspace{-10pt}
\section{Introduction}
\label{sec:intro}
\vspace{-2pt}

\begin{wrapfigure}{r}{0.44\textwidth}

\centering
\includegraphics[width=\linewidth]{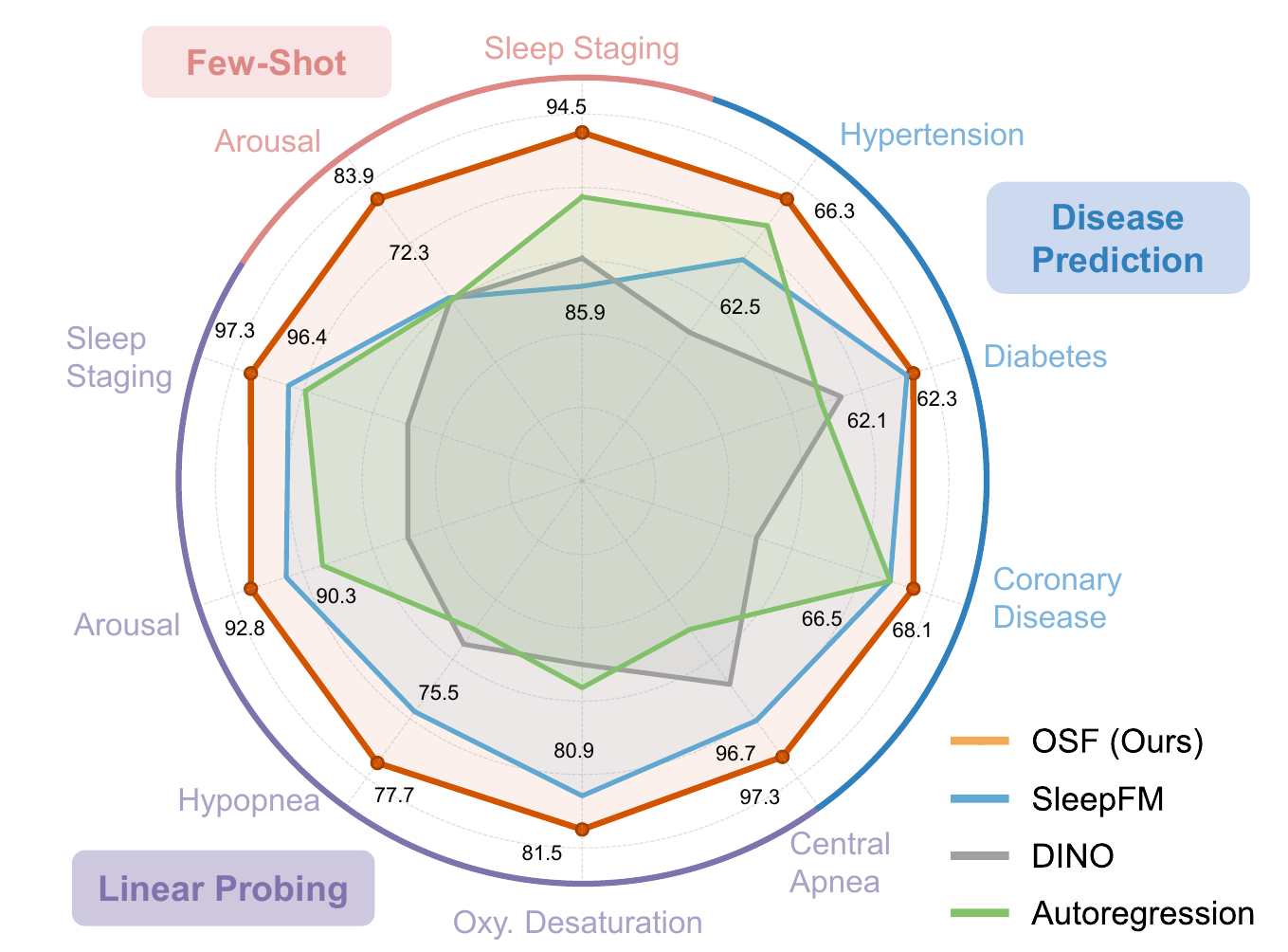}
\caption{\textbf{Performance comparison across downstream tasks.} \ours consistently achieves state-of-the-art on downstream tasks.}
\label{fig:radar_graph}
\vspace{-16pt}

\end{wrapfigure}

Sleep is a fundamental physiological process for human health \citep{thapa2026multimodal, xu2026sleeplm, yang2022artificial}. Assessing sleep quality \citep{perslev2021u} and detecting sleep disorder events \citep{sands2018quantifying} often rely on polysomnography (PSG). PSG typically includes multiple complementary signals that capture brain activity, respiratory effort, muscle movement, and cardiac activity. In practice, these recordings differ across patient cohorts and devices \citep{li2026hearts, jia2021multi}, and the available channel set is often inconsistent due to differences in acquisition protocols and occasional channel dropouts. For example, home studies often lack brain signals \citep{ayappa2008validation,zhang2018national}, and sensors can become dislodged during the night. Together, these challenges limit the transferability of sleep analysis models \citep{zhai2024challenges}. 

Recent foundation models (FMs)~\citep{thapa2026multimodal} have shown rapid progress in modeling sleep recordings. They learn generalizable representations via self-supervised learning (SSL) on large-scale 30-second PSG epochs. These representations support efficient adaptation to diverse epoch-level tasks (e.g., sleep staging, arousal detection) and can be aggregated over time to predict patient-level outcomes. Prior work \citep{thapa2026multimodal} has primarily focused on contrastive approaches, where sleep signals are grouped into multiple modalities and their embeddings are aligned in a shared latent space. While contrastive learning has achieved strong results, the broader design space for pre-training sleep FMs remains largely under-explored.

In this work, we move beyond reporting gains from a single pre-training recipe and ask what actually drives transfer in sleep FMs. Existing sleep FMs differ in their pre-training objectives, but it is unclear which of these choices reliably improve downstream performance across cohorts, devices, and channel sets. A central question arises:
\vspace{-5pt}
\begin{center}
\emph{Which \textbf{pre-training} and \textbf{scaling} design choices truly improve the generalization of sleep FMs, especially under cohort shift and missing-channel inference?}
\end{center}
\vspace{-5pt}
To answer this question with controlled evaluation, we conduct a systematic study that quantifies how different pre-training design choices affect downstream performance. We also construct \benchname, the largest fully open, multi-source sleep benchmark, aggregated from nine publicly available datasets with diverse demographics, comprising 166,500 hours of recordings from over 21,000 sleep studies, and spanning a broad set of PSG channels.

Leveraging \benchname, we identify three main findings.
\textit{First}, existing sleep FMs are not reliable under inference-time input incompleteness, showing large performance drops under realistic missing-channel settings.
\textit{Second}, through a controlled, step-by-step study of pre-training design choices, we find that learning channel-invariant features is critical for invariance-driven methods to learn stronger representations.
\textit{Third}, we observe consistent scaling trends in sleep: increasing training data and model capacity, together with multi-source data mixing, improves downstream performance across tasks.
Guided by these insights, we propose \textbf{O}pen \textbf{S}leep \textbf{F}oundation Model (\textbf{\ours}), a family of sleep foundation models that achieves state-of-the-art across diverse sleep-related tasks and datasets (Fig.~\ref{fig:radar_graph}).

\textbf{Our contribution.} In summary, we present a systematic study of how to build generalizable sleep FMs. We curate a large corpus of 166,500 hours of sleep recordings from nine public sources and establish \benchname, a comprehensive, fully open-source benchmark. Leveraging \benchname, we evaluate four families of self-supervised pre-training objectives and identify three key findings: 
\ding{182} existing sleep FMs fail to generalize to missing channels at inference;
\ding{183} channel-invariant feature learning is essential for strong pre-training; and
\ding{184} scaling sample size, model capacity, and multi-source data mixture consistently improves downstream performance. 
Guided by these results, we introduce \ours, a family of sleep FMs trained with an improved pre-training and scaling recipe, achieving state-of-the-art performance across nine datasets on diverse sleep and disease prediction tasks. Further analysis of \ours reveals strong sample efficiency, effective hierarchical aggregation, and consistent scaling across datasets.

\vspace{-5pt}
\section{Fully Open Sleep Benchmarking at Scale}
\label{dataset}

Our benchmark aggregates nine publicly available datasets hosted by the National Sleep Research Resource (NSRR) \citep{zhang2018national}:  SHHS~\cite{quan1997sleep}, NCHSDB~\cite{lee2022large}, CFS~\cite{redline1995familial}, CCSHS~\cite{rosen2003prevalence}, WSC~\cite{young2009burden}, MROS~\cite{blackwell2011associations}, MESA~\cite{chen2015racial}, CHAT~\cite{marcus2013randomized}, and SOF~\cite{spira2008sleep}. In total, it contains \textbf{166,500} hours of PSG recordings spanning more than \textbf{21,000} nights. We further partition the data into an in-domain cohort for pre-training and an external cohort for out-of-domain (OOD) evaluation. Each cohort is split into training, validation, and test sets. All samples in the external cohort are held out from pre-training and used only for downstream evaluation. For the in-domain cohort, we use the training split for pre-training and also for downstream fine-tuning in the in-domain setting. Detailed statistics across cohorts, datasets, and splits are provided in Appendix \ref{appendix:data stats}.

\begin{wrapfigure}{r}{0.45\textwidth}
\centering
\includegraphics[width=\linewidth, trim={0cm 0cm 0cm 0cm}, clip]{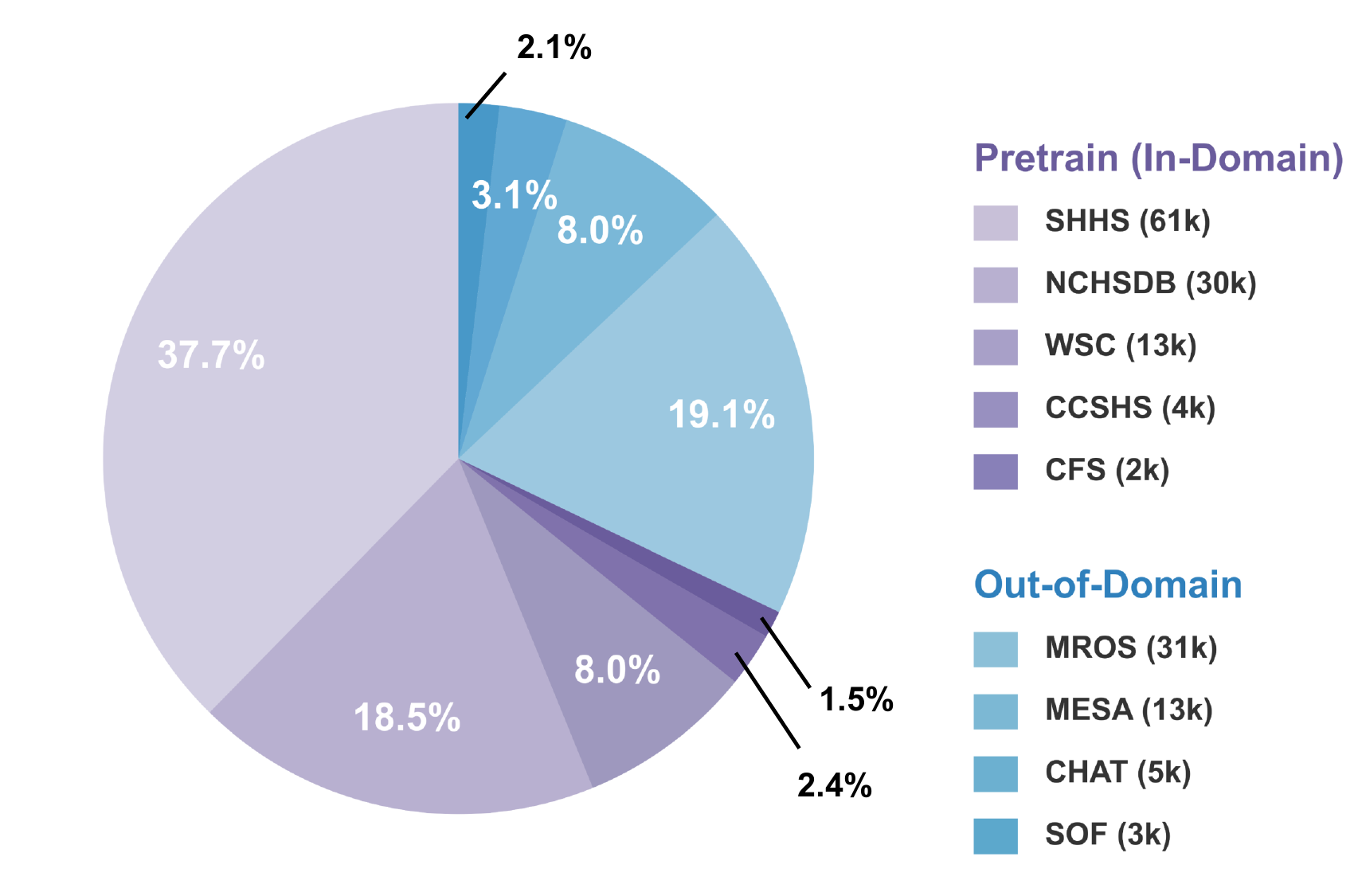}
\vspace{-8pt}
\caption{\textbf{Distribution of our established \benchname.}}
\label{fig:data_dist}
\vspace{-10pt}
\end{wrapfigure}

We focus on a shared subset of channels that is available across most cohorts. Specifically, we utilize a standardized 12-channel montage grouped by physiological modality:
\ding{182} \textit{Brain (EEG/EOG):} \texttt{C3-A2}, \texttt{C4-A1}, \texttt{E1-A2}, \texttt{E2-A1},
\ding{183} \textit{Respiration:} \texttt{Abdominal}, \texttt{Thorax}, \texttt{Nasal Pressure}, \texttt{Snore},
\ding{184} \textit{Cardiac:} \texttt{ECG}, and
\ding{185} \textit{Somatic:} \texttt{EMG-Chin}, \texttt{EMG-LLeg}, \texttt{EMG-RLeg}.
Detailed settings and preprocessing steps are provided in Appendix~\ref{appendix:channel details}.

We process each night's recording with a standardized pipeline to reduce cross-dataset heterogeneity. We first perform manual quality control to trim prolonged wake periods at the beginning and end of the night, removing extreme artifacts from sensor setup and removal. Next, we apply per-night z-score normalization to each channel. For respiratory channels, we additionally apply area-dependent z-score normalization to reduce amplitude drift and improve cross-night consistency. 

We follow prior work and segment each night into non-overlapping 30-second epochs, yielding roughly \textbf{20 million} epochs for self-supervised pre-training. For segmented samples, we resample all channels to 64 Hz and zero-pad missing channels. Epoch-level statistics are provided in Table \ref{tab:epoch_stats}. To the best of our knowledge, this corpus constitutes the \textit{largest} fully public benchmark for pre-training and evaluating sleep foundation models across diverse channel types.
We compare our \benchname other published benchmarks in Appendix~\ref{appendix:data stats}. 
With large-scale sleep data spanning diverse cohorts, \benchname enables controlled benchmarking of different pre-training methods on sleep data, and supports systematic studies of how performance scales with pre-training sample size and how multi-source data blending affects generalization.

We acknowledge ongoing community efforts to harmonize public sleep datasets. 
Rather than treating \benchname as a standalone contribution, we view it as a methodological infrastructure that enables controlled multi-source pre-training and evaluation under a unified setting. 
We will open-source the codebase and processing pipeline, and maintain \benchname as a living resource that can be expanded with additional PSG datasets and community contributions.

\vspace{-5pt}
\section{On Pre-training Sleep FM}
\label{sec:methods}
\vspace{-5pt}

In this section, we study practical design choices for self-supervised pre-training of sleep foundation models. Following \citep{thapa2026multimodal}, we conduct pre-training and downstream evaluation on 30-second epochs segmented from full-night recordings. 

\vspace{-5pt}
\subsection{Generalization Breaks Under Missing-Channel Inference}
\vspace{-5pt}

We first assess whether existing sleep FMs can cope with real-world deployment challenges.
Real-world sleep recordings exhibit heterogeneous channel availability across cohorts, devices, and study protocols, such that channels present in the pre-training corpus may be absent at test time. For instance, home-based recordings \citep{gleason2014challenges} sometimes do not collect EEG and EOG channels, whereas EEG-centric sleep architecture studies \citep{gong2025leveraging} may not include breathing signals. This input mismatch motivates a key research question: \textit{Do current sleep FMs generalize under missing-channel settings?}

\begin{figure}[!t]
\centering
\begin{minipage}[t]{0.42\columnwidth}
  \centering
  \includegraphics[width=\linewidth, trim={0cm 0cm 3.7cm 0cm}, clip]{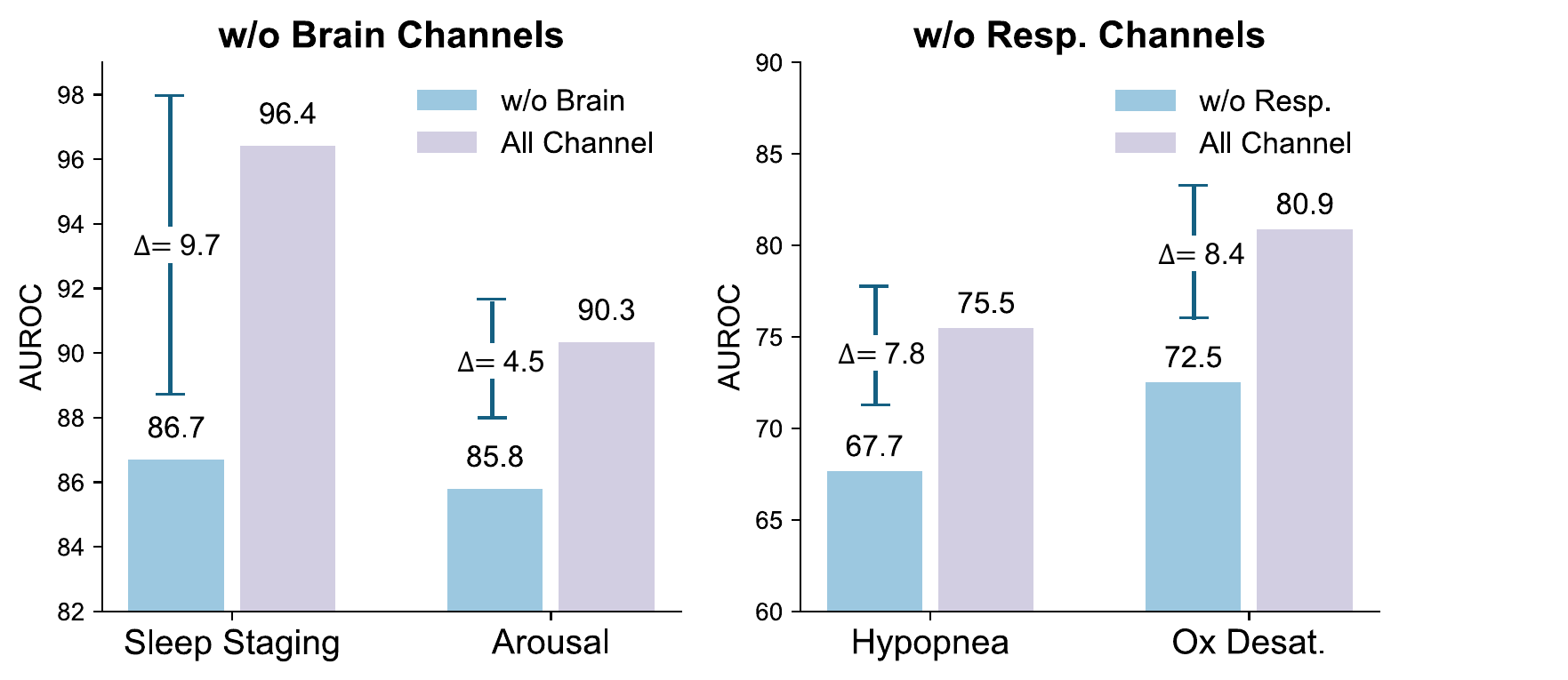}
  \caption{\textbf{Inference with full versus missing channels.} Existing sleep FM fails to generalize to missing channel samples.}
  \label{fig:missing_inference}
\end{minipage}\hfill
\begin{minipage}[t]{0.56\columnwidth}
  \centering
  \includegraphics[width=\linewidth, trim={0cm 0cm 0cm 0cm}, clip]{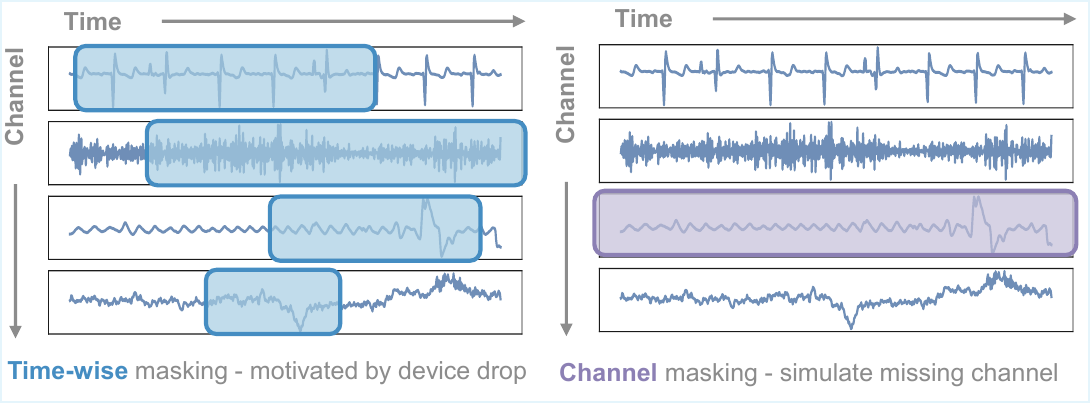}
  \caption{\textbf{Illustration of considered augmentations.} We consider time-wise masking and channel masking strategies.}
  \label{fig:aug_approach}
\end{minipage}
\vspace{-8pt}
\end{figure}

To this end, we simulate realistic missing-channel cases by zero-masking a subset of channels at inference time. We group the 12 channels into three groups: (1) \textit{brain activity} channels, (EEG and EOG), which provide key cues for sleep staging; (2) \textit{respiratory} channels, including breathing effort and airflow waveforms that characterize respiratory physiology; and (3) all remaining channels (ECG and EMG). We focus on two realistic settings: \textit{in-home studies} that do not provide brain signals, and \textit{micro-event studies} where breathing channels are not collected.

We evaluate current sleep FMs under full-channel input and the two missing-channel settings described above.
As shown in Fig. \ref{fig:missing_inference}, removing brain activity channels leads to a substantial drop in sleep staging performance. Similarly, when respiratory channels are removed, hypopnea detection performance degrades significantly.
These results match \textit{clinical intuition}: hypopnea is driven by respiration, and sleep staging depends primarily on brain signals.

\finding{1}{Existing sleep FMs fail to generalize under missing-channel inference, motivating pre-training designs that explicitly handle channel incompleteness.}

\subsection{Channel-Invariant Pre-training Improves Robustness and Transfer}
\label{sec:what ssl works}

\begin{wraptable}{r}{0.55\textwidth}

\centering
\caption{\textbf{Comparisons of different masking strategies.} Channel masking consistently improves downstream performance, and combining time and channel masking yields the best results across pre-training methods.}
\label{tab:motivation_augmentation}

\begin{adjustbox}{width=\linewidth}
\begin{tabular}{l cc  cccc}
\toprule[1.5pt]
\multirow{2.5}{*}{\textbf{Model}} & \multicolumn{2}{c}{\textbf{Mask Strategy}} &
\multicolumn{2}{c}{\textbf{Sleep Staging}} &
\multicolumn{2}{c}{\textbf{Hypopnea}}\\
\cmidrule(lr){2-3}\cmidrule(lr){4-5} \cmidrule(lr){6-7}
& Time & Channel
& AUC$^\uparrow$ & AUPRC$^\uparrow$ & AUC$^\uparrow$ & AUPRC$^\uparrow$ \\
\midrule
\midrule

\multirow{3}{*}{\textsc{SimCLR}}
  & \cmark & \xmark & 81.4 & 56.8 & 58.4 & 53.1 \\
& \xmark & \cmark & 94.8 & 84.3 & 73.3 & 59.2 \\
\grayrow
& \cmark & \cmark & \textbf{96.7} &  \textbf{89.0} & \textbf{75.6} & \textbf{60.9} \\
\midrule

\multirow{3}{*}{\textsc{DINO}}
  & \cmark & \xmark & 93.6 & 81.4 & 72.7 & 59.5 \\
& \xmark & \cmark & 96.4 & 87.9 & 77.1 & 61.4 \\
\grayrow
& \cmark & \cmark & \textbf{97.3} & \textbf{90.4}  & \textbf{77.7} &  \textbf{62.2}\\
\bottomrule[1.5pt]
\end{tabular}
\end{adjustbox}

\end{wraptable}

The failures of existing sleep FMs under missing-channel inference suggest that \textit{pre-training should encourage the model to learn representations that are invariant to the available channel subset.} 

To test this hypothesis, we perform a controlled, step-by-step study using \simclr while varying its augmentation strategy. We focus on SimCLR because it learns by aligning two augmented views of the same input, and its objective is largely shaped by the augmentation design. Specifically, following prior work on self-supervised learning for time series \citep{liu2023self}, we use time-wise block masking as the default augmentation. We compare it against an enhanced strategy that additionally masks a randomly selected subset of input channels.

The augmentation strategies are illustrated in Fig. \ref{fig:aug_approach}. 
For block masking, given an input $\mathbf{x}\in\mathbb{R}^{C\times T}$ with $C$ channels and sequence length $T$, we apply masking independently to each channel. 
For each channel, we sample a masking ratio $r\in[0.3,0.6]$ and a start index $p\sim \mathcal{U}\{0,\ldots,\lfloor(1-r)T\rfloor\}$, and set the contiguous segment $[p,\,p+\lfloor rT\rfloor)$ to zero. For channel masking, we randomly drop 50\% of the channels by setting all time steps in the selected channels to zero.

As shown in Table \ref{tab:motivation_augmentation}, SimCLR pre-trained with channel masking consistently outperforms the default time-only masking by a large margin across multiple downstream tasks. To isolate the contribution of channel masking, we further train a SimCLR variant that uses only channel masking, while keeping all other hyperparameters and masking ratios the same. Table \ref{tab:motivation_augmentation} confirms that this channel-only variant still significantly outperforms the default setting, suggesting that SimCLR learns more robust features when the two views differ in channel availability.

To test whether this effect extends beyond contrastive learning, we conduct an analogous controlled study with \dino, a distillation-based method. As shown in Table \ref{tab:motivation_augmentation}, we observe a similar pattern: although DINO performs strongly with time masking alone, adding channel masking yields a clear improvement. Based on these results, we summarize our second finding as follows:

\finding{2}{Explicitly encouraging channel-invariant feature learning during pre-training improves robustness and downstream transfer, particularly for contrastive and distillation-based methods.}

\vspace{-5pt}
\subsection{Scaling Laws Emerge in Sleep FM Pre-training}
\label{sec:method_scale}

\begin{wrapfigure}{r}{0.48\textwidth}
\centering
\includegraphics[width=\linewidth, trim={0cm 0cm 2.4cm 0cm}, clip]{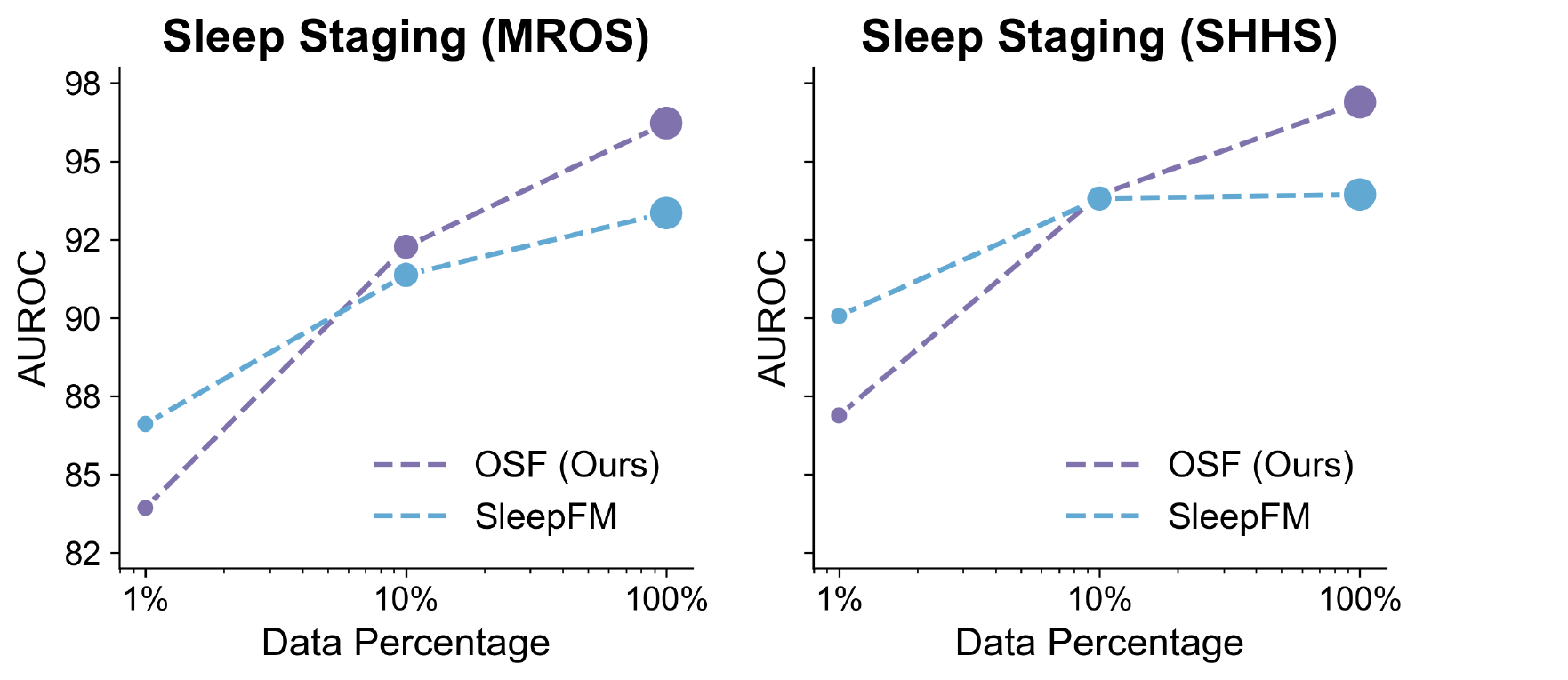}
\caption{\textbf{Comparison of scaling behavior.} Current sleep FM is less scalable. In contrast, \ours better utilizes training data.}
\vspace{-10pt}
\label{fig:scale_1M_4M}
\end{wrapfigure}

Recent sleep FMs are pre-trained on millions of 30-second epochs, yet it remains unclear whether current designs continue to improve as we scale up the amount of pre-training data. We therefore ask \textit{whether sleep FM performance improves consistently as the pre-training sample size increases.}

We start by pre-training the prior sleep FM \citep{thapa2026multimodal} on our pre-training cohort, matching the original parameter budget. As shown in Fig. \ref{fig:scale_1M_4M}, sleep staging performance on SHHS improves when increasing the pre-training data from 1\% to 10\% of the corpus, but shows little or no gain when further scaling from 10\% to 100\%. We observe a similar saturation trend on MROS. These results suggest that the baseline pre-training design may not fully benefit from large-scale sleep data. We then repeat the same data-scaling study using the pre-training recipe identified in Sec. \ref{sec:what ssl works}, while keeping the model size fixed. As shown in Fig. \ref{fig:scale_1M_4M}, performance improves consistently as we scale up the pre-training sample size, indicating that data scaling can hold for sleep FM pre-training with appropriate SSL designs.

We further analyze scaling and multi-source data mixture in more detail. We find that (1) consistent gains can be observed when scaling both sample size and model size, and (2) increasing data diversity via multi-source blending improves generalization relative to training on a single source. Results are presented in Sec. \ref{sec:scaling}. Taken together, these results motivate our final pre-training recipe: we scale model capacity and pre-train on large-scale, multi-source data, while explicitly encouraging the model to learn channel-invariant representations.

\finding{3}{Baseline sleep FMs can saturate as pre-training data grows; with channel-invariant SSL, performance scales more consistently with larger data size, larger models, and multi-source data mixtures.}

\section{\ours: Open Sleep Foundation Models}
\label{sec:main}
\begin{table*}[t]
\centering
\caption{\textbf{Sleep staging and sleep event detection.} \ours achieves the best overall performance among all compared methods. We report macro-AUC and macro-AUPRC.}
\label{tab:mros_dense_task}

\begin{adjustbox}{width=\textwidth}
\footnotesize
\setlength{\tabcolsep}{5pt}
\begin{tabular}{l *{5}{cc}}
\toprule[1.5pt]
\multirow{2}{*}{\textbf{Method}} &
\multicolumn{2}{c}{\textbf{Sleep Staging}} &
\multicolumn{2}{c}{\textbf{Arousal}} &
\multicolumn{2}{c}{\textbf{Hypopnea}} &
\multicolumn{2}{c}{\textbf{Ox. Desat.}} &
\multicolumn{2}{c}{\textbf{Central Apnea}} \\
\cmidrule(lr){2-3}\cmidrule(lr){4-5}\cmidrule(lr){6-7}\cmidrule(lr){8-9}\cmidrule(lr){10-11}
& AUC$^\uparrow$& AUPRC$^\uparrow$
& AUC$^\uparrow$ & AUPRC$^\uparrow$
& AUC$^\uparrow$& AUPRC$^\uparrow$
& AUC$^\uparrow$ & AUPRC$^\uparrow$
& AUC$^\uparrow$ & AUPRC$^\uparrow$ \\
\midrule
\midrule
\multicolumn{11}{l}{\textit{\textbf{Supervised:}}} \\
\textsc{ViT} \cite{dosovitskiy2020image} & 96.9 & 89.9 & 88.4 & 83.0 & 81.3 & 66.2 & 81.8 & 81.6 & 97.3 & 70.5 \\

\midrule
\multicolumn{11}{l}{\textit{\textbf{Linear Probing:}}} \\
\textsc{SleepFM}~\citep{thapa2026multimodal} & 96.4 & 87.7 & \underline{90.3} & \underline{84.9} & 75.5 & 60.6 & 80.9 & 80.3 & \underline{96.7} & 70.3 \\
\textsc{SimCLR}~\citep{chen2020simple} & 81.4 & 56.8 & 73.6 & 67.8 & 58.4 & 53.1 & 66.8 & 65.0 & 79.4 & 53.6 \\
\textsc{DINO}~\citep{oquab2023dinov2} & 93.6 & 81.4 & 82.2 & 75.7 & 72.7 & 59.5 & 78.6 & 78.2 & 96.1 & 66.8 \\
\textsc{VQ-VAE}~\citep{van2017neural} & 95.8 & 86.7 & 86.2 & 80.3 & 75.0 & 60.8 & 79.8 & 79.5 & 96.7 & 68.8 \\
\textsc{MAE}~\citep{he2022masked} & \underline{96.5} & \underline{88.7} & 89.7 & 84.3 & \underline{77.6} & \underline{61.9} & \underline{81.2} & \underline{80.8} & \textbf{97.3} & \textbf{71.7} \\
\textsc{Autoregression}~\citep{radford2019language} & 96.0 & 87.6 & 87.8 & 81.9 & 72.1 & 58.9 & 79.0 & 78.5 & 95.2 & 65.7 \\

\rowcolor{gray!15}
\textsc{\ours} & \textbf{97.3} & \textbf{90.4} & \textbf{92.8} & \textbf{88.3} & \textbf{77.7} & \textbf{62.2} & \textbf{81.5} & \textbf{81.0} & \textbf{97.3} & \underline{70.7} \\

\midrule
\multicolumn{11}{l}{\textit{\textbf{Full Fine-tuning:}}} \\

\textsc{SleepFM}~\citep{thapa2026multimodal} & 97.0 & 89.5 & \underline{93.7} & \underline{89.6} & \underline{85.0} & 69.0 & 82.8 & 82.4 & 97.7 & 71.0 \\
\textsc{SimCLR}~\citep{chen2020simple} & 95.2 & 85.5 & 84.0 & 77.4 & 65.9 & 56.2 & 80.5 & 80.1 & 96.8 & 69.3 \\
\textsc{DINO}~\citep{oquab2023dinov2} & 97.0 & 90.2 & 92.3 & 88.0 & 84.5 & 68.9 & 82.6 & 82.3 & 97.9 & 75.2 \\
\textsc{VQ-VAE}~\citep{van2017neural} & \underline{97.4} & \underline{90.8} & 93.5 & 89.5 & 81.8 & 65.8 & 82.4 & 82.1 & \underline{98.0} & 75.2 \\
\textsc{MAE}~\citep{he2022masked} & 97.3 & 90.6 & 93.3 & 89.3 & \textbf{85.1} & \textbf{69.3} & \underline{83.3} & \underline{83.0} & \textbf{98.1} & \textbf{76.0} \\
\textsc{Autoregression}~\citep{radford2019language} & 97.1 & 90.0 & 93.0 & 88.9 & 84.6 & 68.9 & 82.8 & 82.4 & 97.9 & \underline{75.6} \\

\rowcolor{gray!15}
\textsc{\ours} & \textbf{97.9} & \textbf{92.0} & \textbf{94.6} & \textbf{91.0} & \underline{85.0} & \underline{69.2} & \textbf{83.5} & \textbf{83.1} & \textbf{98.1} & 75.5 \\

\bottomrule[1.5pt]
\end{tabular}
\end{adjustbox}

\end{table*}

\subsection{Experimental Setup}
\textbf{Pre-training Setup.}
To rigorously benchmark self-supervised pre-training methods on sleep data, in addition to \sleepfm, we adapt four representative method families:
\ding{182} \textit{Contrastive Learning}: \simclr;
\ding{183} \textit{Reconstruction-based Method}: MAE \citep{he2022masked} and VQ-VAE \citep{van2017neural};
\ding{184} \textit{Autoregressive Modeling}: Autoregression \citep{radford2019language}; and
\ding{185} \textit{Self-Distillation}: \dino.
In our implementation, we use DINO as the base method and incorporate our pre-training recipes, yielding \ours.
Unless otherwise specified, all methods reported in the main tables use a Transformer \citep{dosovitskiy2020image} encoder with 85M parameters.
All models are pre-trained on the same split of the pre-training cohorts in \benchname for up to 30 epochs, with early stopping upon convergence.
We use AdamW \citep{loshchilov2017decoupled} with linear warmup (10\% of total steps) followed by cosine annealing, and we tune the learning rate and batch size for each method. 

\textbf{Evaluation Setup.}
For main experiments, we evaluate pre-trained models across five \textit{epoch-level} tasks and three \textit{patient-level} disease prediction tasks.
The epoch-level tasks include (1) 4-class sleep staging, and (2) four binary sleep-event detection tasks: Arousal, Hypopnea, Oxygen Desaturation (Ox. Desat.), and Central Apnea.
We also evaluate three patient-level disease classifications: Coronary Disease, Diabetes, and Hypertension.
We report AUROC and AUPRC due to label imbalance.
For epoch-level tasks, we consider three transfer protocols: (1) linear probing, (2) full fine-tuning, and (3) few-shot adaptation.
For disease classification, we segment each recording into 30-second epochs, extract an embedding for each epoch using the pre-trained encoder, and then aggregate epoch embeddings for patient-level prediction.
We use MROS as the primary OOD evaluation cohort and SHHS as the primary in-domain cohort.
Unless otherwise specified, we report results on MROS in the main text.
All pre-training and evaluation configurations, as well as implementation details, are provided in Appendix \ref{appendix:impl_details}.

\begin{table*}[t]
\centering
\caption{\textbf{Linear probing sleep staging across diverse cohorts.} \ours achieves the best performance across all datasets.}
\vspace{-2pt}
\label{tab:sweep_stage_multicohort}
\begin{adjustbox}{width=\textwidth}
\begin{tabular}{l *{4}{cc} | *{3}{cc}}
\toprule[1.5pt]
\multirow{2}{*}{\textbf{Method}} 
& \multicolumn{2}{c}{\textbf{CCSHS}} &
  \multicolumn{2}{c}{\textbf{CFS}} &
  \multicolumn{2}{c}{\textbf{NCHSDB}} &
  \multicolumn{2}{c}{\textbf{WSC}} &
  \multicolumn{2}{c}{\textbf{CHAT}} &
  \multicolumn{2}{c}{\textbf{MESA}} &
  \multicolumn{2}{c}{\textbf{SOF}} \\
\cmidrule(lr){2-3}\cmidrule(lr){4-5}\cmidrule(lr){6-7}\cmidrule(lr){8-9}\cmidrule(lr){10-11}\cmidrule(lr){12-13}\cmidrule(lr){14-15}
& AUC$^\uparrow$ & AUPRC$^\uparrow$
& AUC$^\uparrow$ & AUPRC$^\uparrow$
& AUC$^\uparrow$ & AUPRC$^\uparrow$
& AUC$^\uparrow$ & AUPRC$^\uparrow$
& AUC$^\uparrow$ & AUPRC$^\uparrow$
& AUC$^\uparrow$ & AUPRC$^\uparrow$
& AUC$^\uparrow$ & AUPRC$^\uparrow$ \\

\midrule
\midrule
& \multicolumn{8}{l}{\textit{\textbf{In-Domain Datasets}}} & \multicolumn{6}{l}{\textit{\textbf{Out-of-Domain Datasets}}} \\
\textsc{SleepFM}           & 97.8 & 94.9 & 97.4 & 93.4 & 95.2 & 89.0 & 97.7 & 89.2 & 97.8 & 95.1 & 94.6 & 82.8 & 96.2 & 89.7 \\
\textsc{SimCLR}            & 84.5 & 65.4 & 83.7 & 64.5 & 80.3 & 56.5 & 84.2 & 56.3 & 87.9 & 70.2 & 73.8 & 46.6 & 79.5 & 56.4 \\
\textsc{DINO}              & 95.6 & 88.9 & 95.9 & 89.4 & 91.2 & 78.8 & 96.2 & 85.3 & 96.5 & 90.7 & 89.4 & 71.5 & 94.3 & 85.0 \\
\textsc{VQ-VAE}            & 97.4 & 93.6 & 97.1 & 92.5 & 93.8 & 85.2 & 97.2 & 88.1 & 97.6 & 93.8 & 93.0 & 79.0 & 95.0 & 86.3 \\
\textsc{MAE}               & 97.7 & 94.8 & 97.4 & 93.5 & 95.0 & 88.3 & 97.7 & 89.1 & 97.9 & 94.9 & 92.3 & 78.4 & 95.8 & 88.8 \\
\textsc{AR}                & 97.6 & 94.4 & 97.3 & 93.2 & 94.7 & 87.4 & 97.6 & 89.3 & 97.8 & 94.5 & 90.4 & 73.7 & 95.7 & 88.6 \\
\rowcolor{gray!15}
\textsc{\ours}     & \textbf{98.6} & \textbf{96.7} & \textbf{97.9} & \textbf{94.9} & \textbf{95.8} & \textbf{90.1} & \textbf{98.1} & \textbf{90.6} & \textbf{98.4} & \textbf{96.2} & \textbf{95.9} & \textbf{85.7} & \textbf{97.0} & \textbf{91.8} \\
\bottomrule[1.5pt]
\end{tabular}
\end{adjustbox}

\end{table*}

\begin{figure*}
\centering
\includegraphics[width=1\linewidth, trim={0cm 0cm 0cm 0cm}, clip]{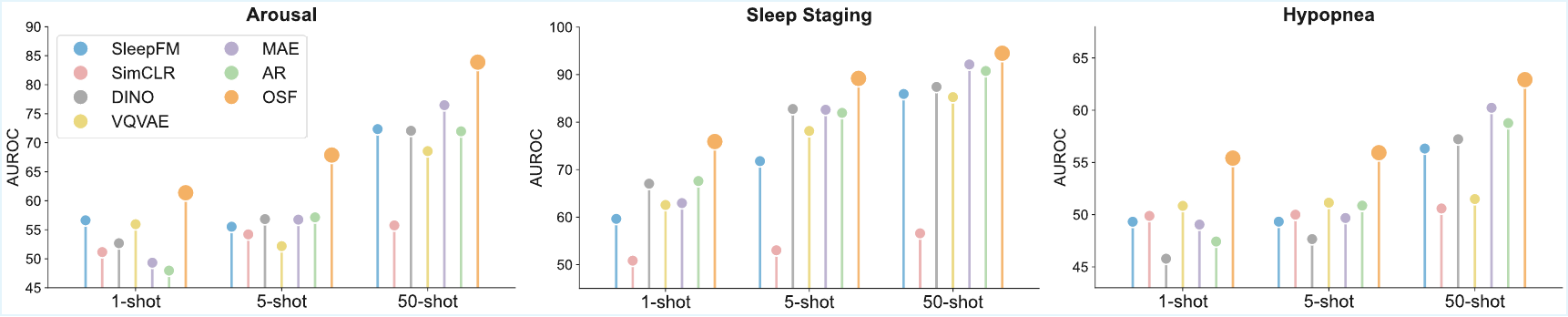}
\caption{\textbf{Few-shot adaptation to downstream tasks.} \ours achieves the best performance across all tasks at every shot count.}
\label{fig:main_few_shot}
\vspace{-6pt}
\end{figure*}

\subsection{Main Results}

\textbf{\ours achieves state-of-the-art on sleep analysis tasks.}
We evaluate linear-probing performance for sleep staging across all cohorts in \benchname.
As shown in Table \ref{tab:sweep_stage_multicohort}, \ours achieves the best performance across all cohorts.
We further evaluate models on four sleep event detection tasks.
On MROS, Table \ref{tab:mros_dense_task} shows that \ours achieves state-of-the-art performance on both sleep staging and event detection under linear probing and fine-tuning.
We observe similar trends on SHHS (Appendix \ref{appendix:main_downstream}).

\begin{wrapfigure}{r}{0.5\textwidth}
\centering
\includegraphics[width=\linewidth, trim={0cm 0cm 0cm 0cm}, clip]{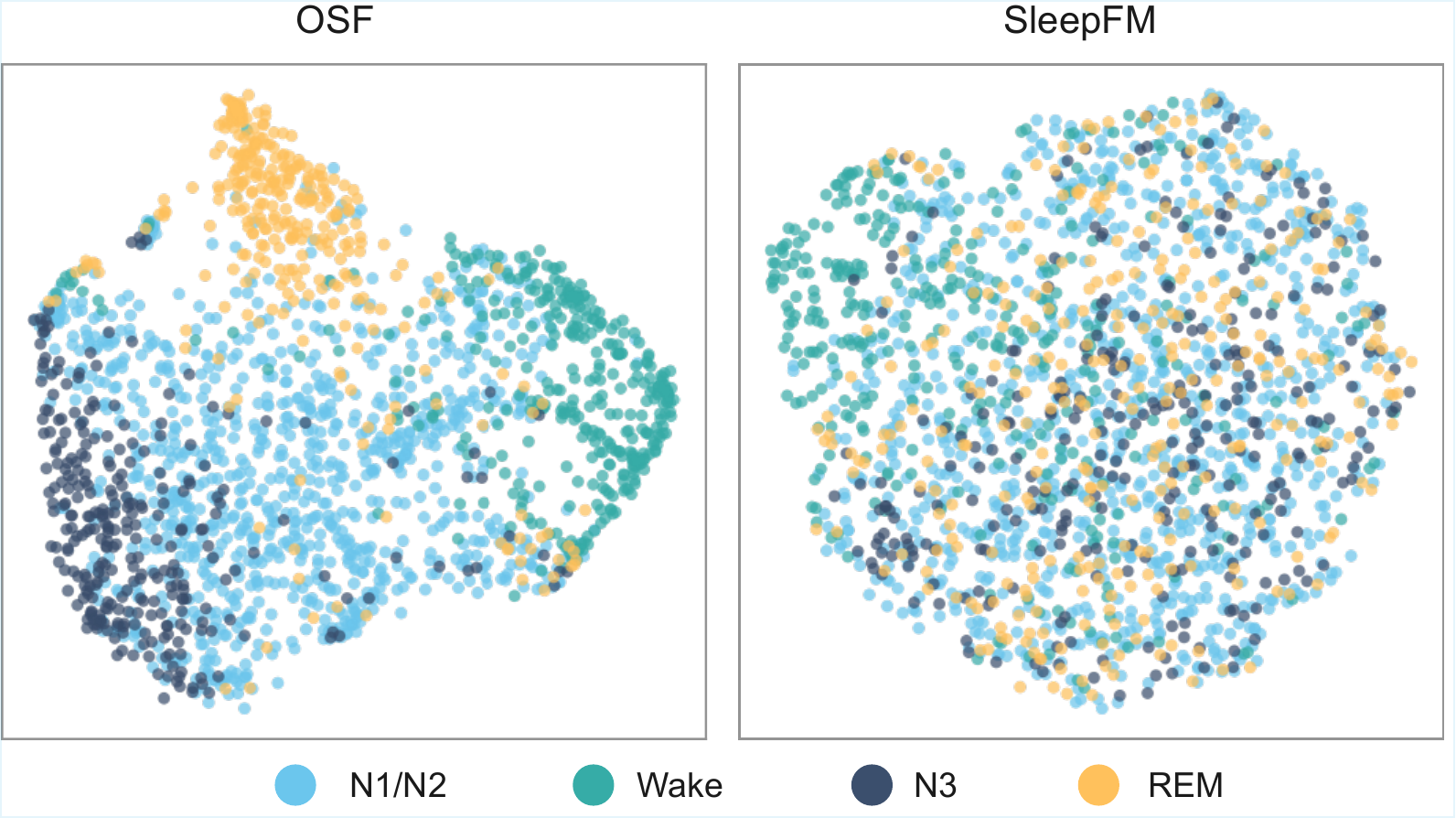}
\caption{\textbf{UMAP visualization of learned representations.} \ours produces more stage-separable clusters, indicating stronger alignment between the embedding geometry and sleep stage structure.}
\label{fig:placeholder}
\vspace{-10pt}
\end{wrapfigure}

\textbf{\ours adapts better in few-shot settings.}
To measure adaptation under limited supervision, we evaluate all models in a few-shot setting on the out-of-domain MROS cohort.
We consider $K \in \{1, 5, 50\}$ labeled examples per class, train a linear classifier on the $K$-shot training subset, and evaluate on the same held-out test set.
As shown in Fig. \ref{fig:main_few_shot}, \ours consistently outperforms all other methods across three sleep analysis tasks for all shot budgets.
Detailed results are provided in Appendix \ref{appendix:main few-shot}.

\begin{wraptable}{r}{0.5\textwidth}
  \centering
  \caption{\textbf{Comparison with domain-specific pretraining methods.} OSF outperforms baselines tailored for physiological signals.}
  \label{tab:mros_domain_specific_baseline}
  \vspace{-2mm}
  \begin{adjustbox}{width=\linewidth}
  \footnotesize
  \setlength{\tabcolsep}{5pt}
  \begin{tabular}{l *{4}{c}}
    \toprule[1.5pt]
    \multirow{2}{*}{\textbf{Method}} &
    \multicolumn{2}{c}{\textbf{Sleep Staging}} &
    \multicolumn{2}{c}{\textbf{Hypopnea}} \\
    \cmidrule(lr){2-3}\cmidrule(lr){4-5}
    & AUC$^\uparrow$ & AUPRC $^\uparrow$ & AUC $^\uparrow$ & AUPRC $^\uparrow$ \\
    \midrule
    \midrule
    \textsc{TS2Vec}~\citep{yue2022ts2vec} & 86.9 & 67.3 & 61.9 & 54.5 \\
    \textsc{ST-MEM}~\citep{na2024guiding} & 96.1 & 87.5 & 75.4 & 60.0 \\
    \textsc{LSM-2}~\citep{xu2025lsm} & 95.8 & 86.7 & 74.6 & 60.1 \\
    \textsc{PedSleepMAE}~\citep{pandey2024pedsleepmae} & 84.0 & 59.0 & 59.4 & 53.4 \\
    \textsc{SleepGPT}~\citep{huang2026unified} & 95.4 & 85.7 & 65.1 & 55.7 \\
    \rowcolor{gray!15}
    \textsc{\ours} & \textbf{97.3} & \textbf{90.4} & \textbf{77.7} & \textbf{62.2} \\
    \bottomrule[1.5pt]
  \end{tabular}
  \end{adjustbox}
  \vspace{-2mm}
\end{wraptable}

\textbf{OSF outperforms physiological-signal-specific pre-training methods.}
We further compare \ours with five recent domain-specific baselines: TS2Vec~\citep{yue2022ts2vec}, ST-MEM~\citep{na2024guiding}, LSM-2~\citep{xu2025lsm}, PedSleepMAE~\citep{pandey2024pedsleepmae}, and SleepGPT~\citep{huang2026unified}.
We pre-train PedSleepMAE on \benchname, evaluate SleepGPT from its released checkpoint, and adapt the remaining methods to our benchmark. Table~\ref{tab:mros_domain_specific_baseline} verifies that \ours achieves the strongest performance across all downstream tasks on MROS in this comparison. More results are in the Appendix~\ref{appendix:main_downstream}.

\textbf{\ours learns clinically useful features for disease prediction.}
We evaluate patient-level disease prediction on MROS using the same embedding aggregation for all pre-trained models.
As shown in Table \ref{tab:ssl_compare_disease}, representations learned by \ours transfer better to unseen data and consistently outperform \sleepfm.

\textbf{The embedding space of \ours shows a more structured pattern.}
We further study what properties of the learned representation space may contribute to the strong performance of \ours.
We visualize embeddings from \ours and \sleepfm \ using UMAP~\citep{mcinnes2018umap} on the SHHS pre-training cohort.
Embeddings from \ours form clearer clusters aligned with sleep stage labels, while embeddings from \sleepfm\ appear less structured.
This suggests that \ours learns more separable representations, which may help downstream adaptation.

Overall, \ours provides a practical recipe for building sleep FMs and achieves strong performance, improved sample efficiency, and effective hierarchical aggregation.

\section{On Scaling Sleep FM}
\label{sec:scaling}

\begin{wraptable}{r}{0.5\textwidth}
\centering
\caption{\textbf{Disease prediction results.} \ours achieves the best overall performance.}
\label{tab:ssl_compare_disease}
\begin{adjustbox}{width=\linewidth}
\begin{tabular}{l *{3}{cc}}
\toprule[1.5pt]
\multirow{2}{*}{\textbf{Model}} &
\multicolumn{2}{c}{\textbf{Coronary Dis.}} &
\multicolumn{2}{c}{\textbf{Diabetes}} &
\multicolumn{2}{c}{\textbf{Hypertension}} \\
\cmidrule(lr){2-3}\cmidrule(lr){4-5}\cmidrule(lr){6-7}
& AUC$^\uparrow$ & AUPRC$^\uparrow$
& AUC$^\uparrow$ & AUPRC$^\uparrow$
& AUC$^\uparrow$ & AUPRC$^\uparrow$ \\

\midrule
\midrule
\textsc{End2End ViT}    & 61.4 & 60.1 & 55.8 & 52.5 & 57.7 & 57.4 \\
\midrule
\textsc{SleepFM}            & \underline{66.5} & \underline{65.5} & \underline{62.1} & \underline{54.9} & 62.5 & 61.4 \\
\textsc{SimCLR}         & 60.4 & 60.0 & 60.4 & 54.8 & 58.8 & 57.9 \\
DINO           & 58.0 & 57.3 & 60.1 & 54.3 & 58.2 & 56.1 \\
VQ-VAE         & 65.5 & 64.4 & 61.8 & 54.6 & 62.1 & 59.7 \\
MAE            & 65.4 & 65.6 & 57.6 & 53.5 & 64.4 & \underline{63.9} \\
AR             & 66.5 & 65.3 & 59.5 & 54.4 & \underline{64.6} & 63.2 \\
\grayrow
\ours   & \textbf{68.1} & \textbf{67.4} & \textbf{62.3} & \textbf{56.3} & \textbf{66.3} & \textbf{66.1} \\
\bottomrule[1.5pt]
\end{tabular}
\end{adjustbox}
\end{wraptable}

In this section, we study how multi-source data mixtures affect downstream performance, and we examine the scaling behavior of sleep FM pre-training with respect to sample size and model capacity.

\begin{figure}[!t]
\centering
\begin{minipage}[t]{0.46\columnwidth}
  \centering
  \includegraphics[width=\linewidth, trim={0cm 0cm 2.4cm 0cm}, clip]{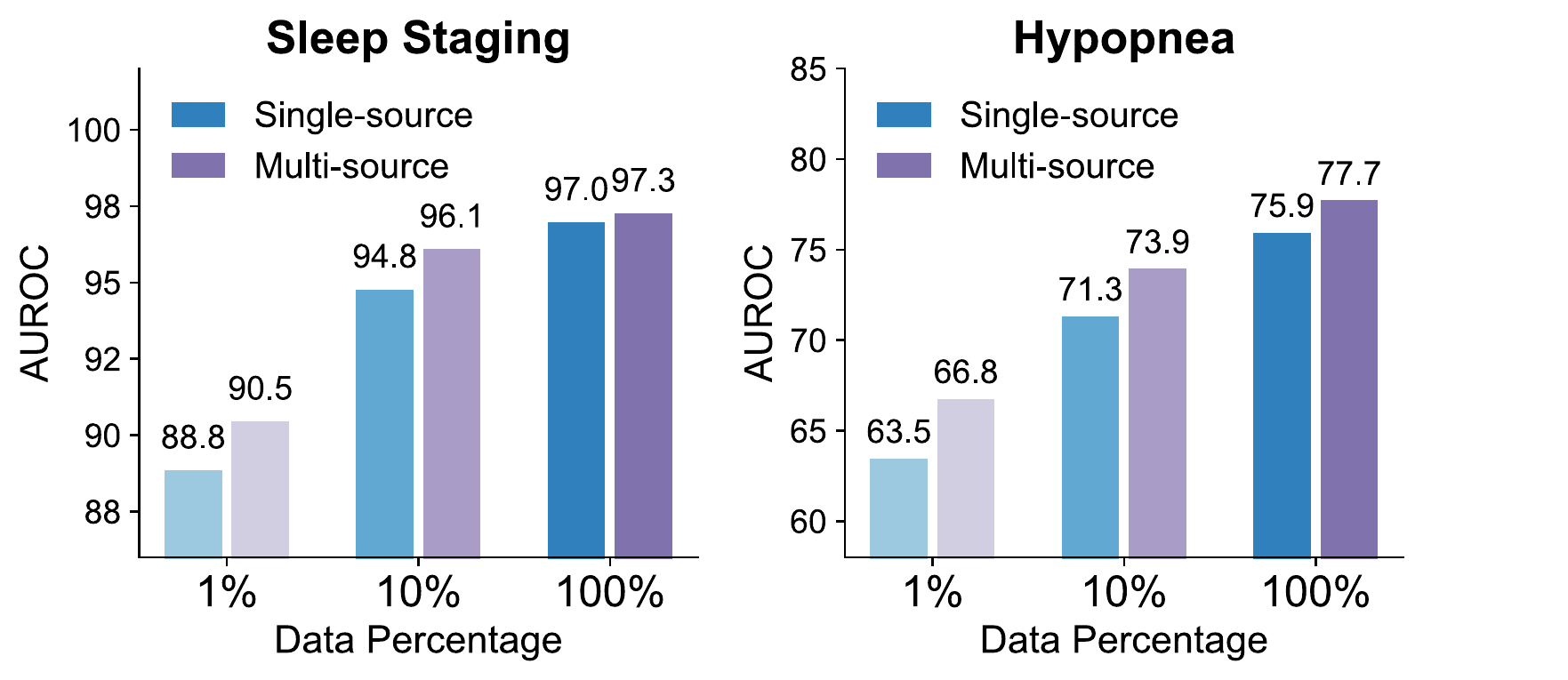}
  \caption{\textbf{Single-source vs. multi-source pretraining.} Multi-source pre-training yields consistently better downstream performance.}
  \label{fig:multi_single}
\end{minipage}\hfill
\begin{minipage}[t]{0.52\columnwidth}
  \centering
  \includegraphics[width=\linewidth, trim={0cm 0cm 0cm 0cm}, clip]{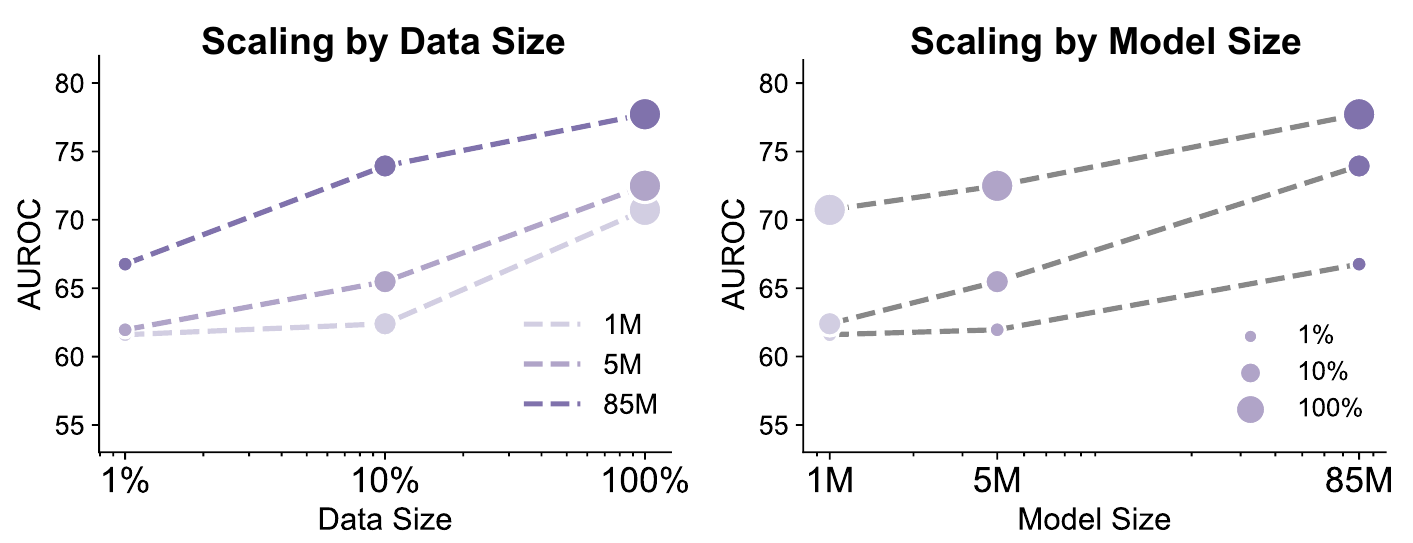}
  \caption{\textbf{Scaling behavior.} Linear probing results on hypopnea detection show that \ours improves with both model capacity and pre-training sample size.}
  \label{fig:scaling-ours}
\end{minipage}
\vspace{-8pt}
\end{figure}

\textbf{Does multi-source data mixture help pre-training?}
Prior work often pre-trains sleep FMs on multi-source corpora. However, cohorts differ in acquisition protocols and population characteristics, and mixing datasets introduces distribution shift.
To examine whether multi-source data mixture helps pre-training, we train two variants of \ours: one pre-trained on \textit{SHHS only} and the other pre-trained on the \textit{full} multi-cohort corpus. 
As shown in Fig. \ref{fig:multi_single}, multi-source pre-training yields consistent improvements across all tasks on the out-of-domain MROS dataset. Moreover, when we vary the pre-training sample size and train both single-source and multi-source variants under identical settings, multi-source pre-training consistently outperforms single-source pre-training. Additional results are provided in Appendix \ref{appendix:multi-source}.

\takeaway{1}{Multi-source pre-training improves generalization by increasing data diversity.}

\textbf{Does scaling model size help pre-training on sleep data?}
To answer this question, we pre-train \ours with ViT encoders of increasing capacity, ranging from 1M and 5M to 85M parameters, and evaluate them via linear probing on hypopnea detection. As shown in Fig. \ref{fig:scaling-ours}, larger models consistently achieve better performance. 
We observe a similar trend for sleep staging (Appendix \ref{appendix:scaling}). These results suggest that, given the richness and diversity of epoch-level sleep signals, increasing model capacity does not lead to obvious overfitting in our setting and translates into stronger transferable representations. Additional details are provided in Appendix \ref{appendix:model scaling}.

\begin{wraptable}{r}{0.5\textwidth}
\centering
\caption{\textbf{Robustness of pre-training with fewer channels.} We pre-train both models using only ECG and respiratory signals and evaluate downstream performance. \ours makes better use of this limited channel set.}
\label{tab:monitor_type_ablation}
\vspace{-2mm}
\begin{adjustbox}{width=\linewidth}
\footnotesize
\setlength{\tabcolsep}{5pt}
\begin{tabular}{l cc cc}
\toprule[1.5pt]

\multirow{2}{*}{\textbf{Method}} &
\multicolumn{2}{c}{\textbf{Hypopnea}} &
\multicolumn{2}{c}{\textbf{Ox.\ Desat.}} \\
\cmidrule(lr){2-3}\cmidrule(lr){4-5}
&
AUC$^\uparrow$ & AUPRC$^\uparrow$ &
AUC$^\uparrow$ & AUPRC$^\uparrow$ \\
\midrule
\midrule

\textsc{SleepFM} & 71.3 & 58.6 & 79.7 & 78.9 \\
\grayrow
\textsc{\ours} & \textbf{77.6} & \textbf{62.4} & \textbf{80.8} & \textbf{80.1} \\

\bottomrule[1.5pt]
\end{tabular}
\end{adjustbox}
\vspace{-2mm}
\end{wraptable}

\textbf{Does scaling sample size help pre-training on sleep data?}
Beyond the results in Sec. \ref{sec:methods}, we further test data scaling for \ours on harder settings and larger model sizes. Specifically, we pre-train \ours with ViT-1M, 5M, and 85M encoders using 1\%, 10\%, and 100\% of the multi-source corpus, and evaluate via linear probing on MROS. As shown in Fig. \ref{fig:scaling-ours}, hypopnea detection performance improves consistently as the pre-training sample size increases. This indicates that pre-training performance scales positively with data size in our setting. Results on additional tasks and cohorts are provided in Appendix \ref{appendix:data scaling}.

\takeaway{2}{Scaling laws emerge in sleep data; jointly scaling model and data size yields the strongest gains.}

\section{Discussion and Analysis}
\label{sec:discussion}

\subsection{On Pre-training and Inference with Incomplete Channels}
\label{sec:missing channel}
\textbf{Evaluation under missing-channel inference.}
We evaluate pre-trained models under realistic missing-channel settings.
Motivated by real-world deployment scenes, we consider four common study configurations: 
\ding{182} \textit{\textbf{head-band device only}} \citep{arnal2020dreem}; 
\ding{183} \textit{\textbf{sleep disorder studies}} \citep{gleason2014challenges}; 
\ding{184} \textit{\textbf{sleep micro-event studies}} \citep{gong2025leveraging}; and
\ding{185} \textit{\textbf{in-home studies}} \citep{redline2014sleep} that provide breathing signals only. We report linear probing results on sleep analysis tasks on MROS. As shown in Table \ref{tab:ablation_channel_setting_mros}, \ours consistently outperforms SleepFM across these missing-channel settings. Specifically, (1) \ours \textit{makes better use of the available channels.} With brain-activity channels only, it achieves stronger sleep staging and arousal detection, suggesting stronger brain-related representations. Similarly, with respiratory channels only, it achieves stronger performance on hypopnea and oxygen desaturation. (2) \ours \textit{is more robust when key modalities are missing.} When respiratory signals are removed, both methods degrade on hypopnea and oxygen desaturation, but \ours remains consistently better. Conversely, when brain-related channels are unavailable, sleep staging becomes much harder for both models; nevertheless, \ours better uses the remaining channels and yields stronger performance.

\begin{table*}[!t]
  \centering
  \caption{\textbf{Linear probing results under realistic missing-channel settings.} \textsc{\ours} is more robust to missing channels.} 
  \label{tab:ablation_channel_setting_mros}
  \vspace{-2mm}
  \begin{adjustbox}{width=\textwidth}
  \footnotesize
  \setlength{\tabcolsep}{3pt}
  \begin{tabular}{l ccc l *{4}{ll}}
    \toprule[1.5pt]
    \multirow{2.5}{*}{\textbf{Realistic Settings Ref.}} & \multirow{2.5}{*}{\textbf{Brain}} & \multirow{2.5}{*}{\textbf{Resp.}} & \multirow{2.5}{*}{\textbf{ECG}} & \multirow{2.5}{*}{\textbf{Method}} &
    \multicolumn{2}{c}{\textbf{Sleep Staging}} &
    \multicolumn{2}{c}{\textbf{Arousal}} &
    \multicolumn{2}{c}{\textbf{Hypopnea}} &
    \multicolumn{2}{c}{\textbf{Ox. Desat.}} \\
    \cmidrule(lr){6-7}\cmidrule(lr){8-9}\cmidrule(lr){10-11}\cmidrule(lr){12-13}
    & & & & & AUC$^\uparrow$ & AUPRC$^\uparrow$
      & AUC$^\uparrow$ & AUPRC$^\uparrow$
      & AUC$^\uparrow$ & AUPRC$^\uparrow$
      & AUC$^\uparrow$ & AUPRC$^\uparrow$ \\
    \midrule
    \midrule

    \multirow{1}{*}{\textbf{Head-Band Data}} &\multirow{2}{*}{\textbf{\cmark}} & \multirow{2}{*}{\textbf{\xmark}} & \multirow{2}{*}{\textbf{\xmark}} &
      \textsc{SleepFM} & 96.6 & 88.7 & 90.8 & 85.5 & 67.5 & 56.7 & 69.9 & 68.9 \\
   \citep{arnal2020dreem} & & & & \textsc{\ours} & \textbf{97.2}~\inc{0.6} & \textbf{90.0}~\inc{1.3} & \textbf{92.2}~\inc{1.4} & \textbf{87.6}~\inc{2.1} & \textbf{68.8}~\inc{1.3} & \textbf{57.1}~\inc{0.4} & \textbf{70.7}~\inc{0.8} & \textbf{69.9}~\inc{1.0} \\
    \midrule

    \multirow{1}{*}{\textbf{Sleep Disorder Study}} & \multirow{2}{*}{\textbf{\xmark}} & \multirow{2}{*}{\textbf{\cmark}} & \multirow{2}{*}{\textbf{\cmark}} &
      \textsc{SleepFM} & 86.7 & 64.9 & 85.8 & 79.8 & 75.7 & 60.8 & 80.9 & 80.1 \\
    \citep{gleason2014challenges} & & & & \textsc{\ours} & \textbf{86.8}~\inc{0.1} & \textbf{65.3}~\inc{0.4} & \textbf{86.4}~\inc{0.6} & \textbf{80.5}~\inc{0.7} & \textbf{77.6}~\inc{1.9} & \textbf{62.2}~\inc{1.4} & \textbf{81.1}~\inc{0.2} & \textbf{80.4}~\inc{0.3} \\
    \midrule

    \multirow{1}{*}{\textbf{Sleep Micro-Event. Study}} & \multirow{2}{*}{\textbf{\cmark}} & \multirow{2}{*}{\textbf{\xmark}} & \multirow{2}{*}{\textbf{\cmark}} &
      \textsc{SleepFM} & 96.2 & 87.2 & 89.5 & 83.7 & 67.7 & 56.7 & \textbf{72.5} & \textbf{71.8} \\
    \citep{gong2025leveraging}& & & & \textsc{\ours} & \textbf{97.1}~\inc{0.9} & \textbf{89.9}~\inc{2.7} & \textbf{92.3}~\inc{2.8} & \textbf{87.5}~\inc{3.8} & \textbf{68.7}~\inc{1.0} & \textbf{57.1}~\inc{0.4} & 71.6~\dec{0.9} & 70.7~\dec{1.1} \\
    \midrule

    \multirow{1}{*}{\textbf{In-Home Study}} & \multirow{2}{*}{\textbf{\xmark}} & \multirow{2}{*}{\textbf{\cmark}} & \multirow{2}{*}{\textbf{\xmark}} &
      \textsc{SleepFM} & 86.7 & 65.2 & 85.7 & 79.7 & 77.4 & 62.0 & 81.1 & 80.5 \\
    \citep{redline2014sleep}& & & & \textsc{\ours} & \textbf{87.4}~\inc{0.7} & \textbf{65.9}~\inc{0.7} & \textbf{86.6}~\inc{0.9} & \textbf{80.8}~\inc{1.1} & \textbf{81.0}~\inc{3.6} & \textbf{65.1}~\inc{3.1} & \textbf{81.9}~\inc{0.8} & \textbf{81.3}~\inc{0.8} \\

    \bottomrule[1.5pt]
  \end{tabular}
  \end{adjustbox}
  \vspace{-3mm}
\end{table*}

\begin{wraptable}{r}{0.5\textwidth}
\centering
\caption{\textbf{Inference with different types of channel corruption.} \ours demonstrates stronger robustness across practical channel corruption scenarios.}
\label{tab:ch_corruption}
\vspace{-2mm}
\begin{adjustbox}{width=\linewidth}
\footnotesize
\setlength{\tabcolsep}{5pt}
\begin{tabular}{l l cc cc}
\toprule[1.5pt]
\multirow{2}{*}{\textbf{Setting}} & \multirow{2}{*}{\textbf{Method}} &
\multicolumn{2}{c}{\textbf{Sleep Staging}} &
\multicolumn{2}{c}{\textbf{Hypopnea}} \\
\cmidrule(lr){3-4}\cmidrule(lr){5-6}
& & AUC$^\uparrow$ & AUPRC$^\uparrow$ & AUC$^\uparrow$ & AUPRC$^\uparrow$ \\
\midrule
\midrule
\multirow{2}{*}{Random Noise}
& \textsc{SleepFM} & 96.1 & 87.1 & \textbf{73.1} & \textbf{59.7} \\

& \textsc{\ours}     & \textbf{97.1} & \textbf{90.0} & 69.4 & 57.5 \\
\midrule
\multirow{2}{*}{Temporal Detach}

& \textsc{SleepFM} & 96.1 & 87.0 & 69.1 & 57.5 \\

& \textsc{\ours}     & \textbf{97.3} & \textbf{90.4} & \textbf{76.2} & \textbf{61.2} \\
\bottomrule[1.5pt]
\end{tabular}
\end{adjustbox}
\vspace{-2mm}
\end{wraptable}

\textbf{Evaluation under more types of channel corruption.}
To further evaluate robustness beyond missing-channel settings, we consider two additional corruption scenarios applied to non-brain-related channels: \textit{noise interference} and \textit{temporal channel detachment}.
We show linear probing results on MROS dataset in Table~\ref{tab:ch_corruption}, \ours achieves the best performance in most settings and tasks, suggesting that its robustness extends beyond channel removal to other practically relevant forms of channel corruption.
The advantage is particularly clear under temporal detachment, which is consistent with our use of channel masking and time-wise masking during pre-training.

\textbf{Pre-training with fewer channels.}
Brain-related channels provide strong cues for many sleep analysis tasks, but they are inconvenient for patients to collect. We therefore study a practical pre-training setup that uses only ECG and respiratory channels \citep{gleason2014challenges}. We evaluate on hypopnea and oxygen desaturation, since these tasks are primarily driven by respiratory and cardiac physiology. As shown in Table \ref{tab:monitor_type_ablation}, \ours makes better use of the available channels and consistently outperforms SleepFM by a large margin on both hypopnea and oxygen desaturation.

Overall, \ours better addresses practical constraints in inference and pre-training with incomplete channels. However, a meaningful performance gap remains when task-critical channels are absent. We encourage future work to further improve robustness under channel missingness.

\subsection{Ablation on Design Variations}

\begin{wraptable}{r}{0.53\textwidth}
\centering
\caption{\textbf{Comparisons of different embedding aggregation methods.} Top-$k$ selection consistently outperforms mean pooling for patient-level prediction.}
\vspace{-4pt}
\label{tab:aggregation_ablation}
\begin{adjustbox}{width=\linewidth}
\begin{tabular}{l *{3}{cc}}
\toprule[2pt]
\multirow{2}{*}{\textbf{Aggregation}} &
\multicolumn{2}{c}{\textbf{Coronary Dis.}} &
\multicolumn{2}{c}{\textbf{Diabetes}} &
\multicolumn{2}{c}{\textbf{Hypertension}} \\
\cmidrule(lr){2-3}\cmidrule(lr){4-5}\cmidrule(lr){6-7}
& AUC$^\uparrow$ & AUPRC$^\uparrow$
& AUC$^\uparrow$ & AUPRC$^\uparrow$
& AUC$^\uparrow$ & AUPRC$^\uparrow$ \\

\midrule
\midrule
\textsc{Avg Pool}          & 65.4 & 64.6 & \underline{60.0} & 55.4 & \textbf{64.2} & \textbf{64.1} \\
\textsc{LSTM}           & 63.5 & 63.0 & 59.9 & \underline{55.9} & 61.8 & 60.9 \\
\textsc{MIL~\citep{ilse2018attention}}               & \underline{66.1} & \underline{65.5} & 58.6 & 53.7 & 62.3 & 61.4 \\
\grayrow
\textsc{Top-$k$ Selection} & \textbf{66.8} & \textbf{66.3} & \textbf{62.8} & \textbf{56.3} & \underline{64.1} & \underline{63.5} \\
\bottomrule[2pt]
\end{tabular}
\end{adjustbox}
\vspace{-4pt}
\end{wraptable}

\textbf{How to aggregate epoch-level embeddings for patient-level tasks?}
Patient-level disease prediction requires aggregating epoch-level representations into a single patient-level feature. Given a full-night recording, we segment it into $N$ non-overlapping 30-second epochs, feed each epoch into the pre-trained encoder, and obtain a sequence of embeddings $\{\mathbf{z}_i\}_{i=1}^{N}$ where $\mathbf{z}_i \in \mathbb{R}^{D}$. Prior work~\citep{thapa2026multimodal} typically uses simple pooling or a learnable temporal aggregator (e.g., an LSTM \citep{hochreiter1997long}) to summarize this sequence. However, downstream datasets often contain only a limited number of patient-level recordings, which can increase the risk of overfitting. This motivates us to test whether a learnable aggregation module is necessary.

We hypothesize that disease-related signals concentrate in specific sleep periods, and thus simple average pooling may be suboptimal for aggregating epoch embeddings. To test this idea, we evaluate multiple-instance learning (MIL) \citep{ilse2018attention} and an LSTM-based aggregator. We also test a selection module trained with straight-through estimation to select and average the top-$k$ disease-relevant epoch embeddings. We apply each method to embeddings extracted from multiple pre-trained models and average performance across models. As shown in Table \ref{tab:aggregation_ablation}, top-$k$ selection outperforms mean pooling overall.

\takeaway{3}{Patient-level prediction benefits from better joint modeling of epoch-level embedding.}

\begin{wraptable}{r}{0.5\textwidth}
\centering
\caption{\textbf{Comparison of augmentation strategies.} We report linear-probing performance (AUC$^\uparrow$). Standard vision augmentations offer limited additional benefit beyond channel masking.}
\vspace{-3pt}
\label{tab:ablation_on_aug}
\begin{adjustbox}{width=\linewidth}
\begin{tabular}{ccc | cccc}
\toprule[1.5pt]
\ \textbf{Crop} & \textbf{Time} & \textbf{Channel}
& \textbf{Sleep Staging} & \textbf{Arousal} & \textbf{Hypopnea} & \textbf{Ox.\ Desat.} \\
\midrule
\midrule
  \cmark & \xmark & \xmark & 95.4 & 85.7 & 77.2 & 80.6 \\
  \xmark & \xmark & \cmark & 96.4 & 90.5 & 77.1 & 80.3 \\
  \cmark & \cmark & \cmark & 96.6 & 89.3 & 77.3 & 81.0 \\
\grayrow
  \xmark & \cmark & \cmark & \textbf{97.3} & \textbf{92.8} & \textbf{77.7} & \textbf{81.5} \\
\bottomrule[1.5pt]
\end{tabular}
\end{adjustbox}
\vspace{-10pt}
\end{wraptable}

\label{sec:ablation_aug}
\textbf{Vision-style augmentations transfer to sleep data, but are not necessary for strong performance.}
To test whether standard vision augmentations (e.g., cropping) can be adapted to sleep data, we implement temporal cropping by randomly selecting a contiguous segment of each input signal, where the crop ratio $r$ is sampled uniformly from $[0.25, 0.75]$. We pre-train models under the same configuration using (1) crop-only augmentation and (2) cropping combined with our channel-masking augmentation. As shown in Table \ref{tab:ablation_on_aug}, cropping alone yields reasonable performance but remains worse than channel masking alone. Moreover, adding cropping on top of our augmentation does not yield consistent gains. These results suggest that additional augmentations can help in some cases, but are not required to obtain strong performance in our setting.

\begin{wraptable}{r}{0.48\textwidth}
\centering
\caption{\textbf{Ablation on channel masking ratios.} A moderate masking ratio works the best.}
\label{tab:main_masking_ratio}
\vspace{-2mm}
\begin{adjustbox}{width=\linewidth}
\footnotesize
\setlength{\tabcolsep}{5pt}
\begin{tabular}{l cc cc}
\toprule[1.5pt]
\multirow{2}{*}{\textbf{Masking Ratio}} &
\multicolumn{2}{c}{\textbf{Sleep Staging}} &
\multicolumn{2}{c}{\textbf{Hypopnea}} \\
\cmidrule(lr){2-3}\cmidrule(lr){4-5}
& AUC$^\uparrow$ & AUPRC$^\uparrow$
& AUC$^\uparrow$ & AUPRC$^\uparrow$ \\
\midrule
\midrule
0.1 & 95.9 & 87.2 & 77.4 & 61.5 \\
\grayrow
0.5  & \textbf{97.3} & \textbf{90.4} & \textbf{77.7} & \textbf{62.2} \\
0.9 & 95.3 & 85.9 & 75.0 & 60.3 \\
\bottomrule[1.5pt]
\end{tabular}
\end{adjustbox}
\vspace{-2mm}
\end{wraptable}

\textbf{Ablation on channel masking ratio.} 
We further ablate the channel masking ratio by pre-training two additional variants with masking ratios of 0.1 and 0.9, while keeping all other settings unchanged.
As shown in Table~\ref{tab:main_masking_ratio}, the default ratio of 0.5 achieves the best overall linear probing performance across tasks, suggesting that moderate masking better balances pretext-task difficulty and input preservation for downstream transfer. More results are in Appendix~\ref{appendix:missing channel inference}.

\vspace{-2pt}
\section{Related Work}
\label{sec:related-work}
\vspace{-1pt}

\textbf{Physiological Foundation Models.}
Foundation models (FMs) have emerged as a strong approach for analyzing physiological time-series data \citep{huang2026unified, xu2026sleeplm, yuan2024self}. 
Building on this line of work, \cite{thapa2026multimodal} propose SleepFM for sleep physiology, showing strong transfer performance on downstream tasks such as sleep staging and disease prediction. SleepFM partitions channels into modality groups and applies contrastive objectives to align their embeddings. Despite these advances, the sleep domain still lacks a clear understanding of which pre-training design choices drive strong representations, and it remains unclear whether scaling trends reported for other physiological FMs \citep{narayanswamy2024scaling, zhang2025sensorlm} reliably hold for sleep data. Our work fills this gap by establishing a fully open multi-source benchmark \benchname and conducting a controlled study of pre-training objectives and scaling behavior for sleep foundation models.

\textbf{Self-Supervised Learning.}
Self-supervised learning is widely used to build FMs across domains. Invariance-based methods, including contrastive learning \citep{chen2020simple, yangsimper} and self-distillation \citep{oquab2023dinov2}, have achieved strong results, while reconstruction-based methods such as MAE \citep{he2022masked} and VQ-VAE \citep{van2017neural} are also effective. In language and time-series modeling, autoregressive pre-training has been especially successful \citep{radford2019language, ansari2024chronos, cohen2025time}. However, existing sleep-focused studies cover only a subset of this design space \citep{thapa2026multimodal, pandey2024pedsleepmae} and lack a unified framework for evaluating and comparing methods. As a result, the field still lacks a systematic analysis of how pre-training design choices affect downstream performance for sleep FMs. In contrast, we benchmark representative objectives from these SSL families under a unified protocol and derive practical design guidelines for scalable sleep FM pre-training.

\section{Conclusion}

In this paper, we present a systematic study of key design choices for building sleep FMs through controlled experiments on a fully open, multi-source corpus. 
Through open benchmarking, we identify three main findings and establish best practices for pre-training sleep FMs. Guided by these insights, we scale both pre-training data and model capacity, explicitly encourage channel-invariant feature learning, and develop the \ours family of sleep FMs, which consistently outperforms existing models. Extensive evaluations across nine datasets and eight downstream tasks demonstrate the effectiveness, robustness, and intriguing behaviors of \ours.

\section*{Acknowledgments}
We gratefully acknowledge the support by Amazon Science Hub and UCLA DataX. Any opinions, findings, conclusions, or recommendations expressed in this material are those of the author(s) and do not necessarily reflect the views of the funders.

\bibliography{ref}

@inproceedings{chen2020simple,
  title={A simple framework for contrastive learning of visual representations},
  author={Chen, Ting and Kornblith, Simon and Norouzi, Mohammad and Hinton, Geoffrey},
  booktitle={International conference on machine learning},
  pages={1597--1607},
  year={2020},
  organization={PmLR}
}

@inproceedings{he2022masked,
  title={Masked autoencoders are scalable vision learners},
  author={He, Kaiming and Chen, Xinlei and Xie, Saining and Li, Yanghao and Doll{\'a}r, Piotr and Girshick, Ross},
  booktitle={Proceedings of the IEEE/CVF conference on computer vision and pattern recognition},
  pages={16000--16009},
  year={2022}
}

@article{van2017neural,
  title={Neural discrete representation learning},
  author={Van Den Oord, Aaron and Vinyals, Oriol and others},
  journal={Advances in neural information processing systems},
  volume={30},
  year={2017}
}

@article{radford2019language,
  title={Language models are unsupervised multitask learners},
  author={Radford, Alec and Wu, Jeffrey and Child, Rewon and Luan, David and Amodei, Dario and Sutskever, Ilya and others},
  journal={OpenAI blog},
  volume={1},
  number={8},
  pages={9},
  year={2019}
}

@article{oquab2023dinov2,
  title={Dinov2: Learning robust visual features without supervision},
  author={Oquab, Maxime and Darcet, Timoth{\'e}e and Moutakanni, Th{\'e}o and Vo, Huy and Szafraniec, Marc and Khalidov, Vasil and Fernandez, Pierre and Haziza, Daniel and Massa, Francisco and El-Nouby, Alaaeldin and others},
  journal={arXiv preprint arXiv:2304.07193},
  year={2023}
}

@article{mcinnes2018umap,
  title={Umap: Uniform manifold approximation and projection for dimension reduction},
  author={McInnes, Leland and Healy, John and Melville, James},
  journal={arXiv preprint arXiv:1802.03426},
  year={2018}
}

@article{loshchilov2017decoupled,
  title={Decoupled weight decay regularization},
  author={Loshchilov, Ilya and Hutter, Frank},
  journal={arXiv preprint arXiv:1711.05101},
  year={2017}
}

@article{dosovitskiy2020image,
  title={An image is worth 16x16 words: Transformers for image recognition at scale},
  author={Dosovitskiy, Alexey},
  journal={arXiv preprint arXiv:2010.11929},
  year={2020}
}

@article{yang2022artificial,
  title={Artificial intelligence-enabled detection and assessment of Parkinson's disease using nocturnal breathing signals},
  author={Yang, Yuzhe and Yuan, Yuan and Zhang, Guo and Wang, Hao and Chen, Ying-Cong and Liu, Yingcheng and Tarolli, Christopher G and Crepeau, Daniel and Bukartyk, Jan and Junna, Mithri R and others},
  journal={Nature Medicine},
  volume={28},
  number={10},
  pages={2207-2215},
  year={2022},
  publisher={Nature Publishing Group}
}

@article{narayanswamy2024scaling,
  title={Scaling wearable foundation models},
  author={Narayanswamy, Girish and Liu, Xin and Ayush, Kumar and Yang, Yuzhe and Xu, Xuhai and Liao, Shun and Garrison, Jake and Tailor, Shyam and Sunshine, Jake and Liu, Yun and others},
  journal={arXiv preprint arXiv:2410.13638},
  year={2024}
}

@inproceedings{yangsimper,
  title={SimPer: Simple Self-Supervised Learning of Periodic Targets},
  author={Yang, Yuzhe and Liu, Xin and Wu, Jiang and Borac, Silviu and Katabi, Dina and Poh, Ming-Zher and McDuff, Daniel},
  booktitle={The Eleventh International Conference on Learning Representations},
  year={2023}
}

@article{yuan2024self,
  title={Self-supervised learning for human activity recognition using 700,000 person-days of wearable data},
  author={Yuan, Hang and Chan, Shing and Creagh, Andrew P and Tong, Catherine and Acquah, Aidan and Clifton, David A and Doherty, Aiden},
  journal={NPJ digital medicine},
  volume={7},
  number={1},
  pages={91},
  year={2024},
  publisher={Nature Publishing Group UK London}
}

@article{thapa2026multimodal,
  title={A multimodal sleep foundation model for disease prediction},
  author={Thapa, Rahul and Kjaer, Magnus Ruud and He, Bryan and Covert, Ian and Moore IV, Hyatt and Hanif, Umaer and Ganjoo, Gauri and Westover, M Brandon and Jennum, Poul and Brink-Kjaer, Andreas and others},
  journal={Nature Medicine},
  pages={1--11},
  year={2026},
  publisher={Nature Publishing Group US New York}
}

@inproceedings{carter2025wav2sleep,
  title={wav2sleep: A Unified Multi-Modal Approach to Sleep Stage Classification from Physiological Signals},
  author={Carter, Jonathan F and Tarassenko, Lionel},
  booktitle={Machine Learning for Health (ML4H)},
  pages={186--202},
  year={2025},
  organization={PMLR}
}

@article{zhang2025sensorlm,
  title={SensorLM: Learning the Language of Wearable Sensors},
  author={Zhang, Yuwei and Ayush, Kumar and Qiao, Siyuan and Heydari, A Ali and Narayanswamy, Girish and Xu, Maxwell A and Metwally, Ahmed A and Xu, Shawn and Garrison, Jake and Xu, Xuhai and others},
  journal={arXiv preprint arXiv:2506.09108},
  year={2025}
}

@inproceedings{pandey2024pedsleepmae,
  title={PedSleepMAE: Generative Model for Multimodal Pediatric Sleep Signals},
  author={Pandey, Saurav Raj and Saeed, Aaqib and Lee, Harlin},
  booktitle={2024 IEEE EMBS International Conference on Biomedical and Health Informatics (BHI)},
  pages={1--8},
  year={2024},
  organization={IEEE}
}

@article{ansari2024chronos,
  title={Chronos: Learning the language of time series},
  author={Ansari, Abdul Fatir and Stella, Lorenzo and Turkmen, Caner and Zhang, Xiyuan and Mercado, Pedro and Shen, Huibin and Shchur, Oleksandr and Rangapuram, Syama Sundar and Arango, Sebastian Pineda and Kapoor, Shubham and others},
  journal={arXiv preprint arXiv:2403.07815},
  year={2024}
}

@article{cohen2025time,
  title={This Time is Different: An Observability Perspective on Time Series Foundation Models},
  author={Cohen, Ben and Khwaja, Emaad and Doubli, Youssef and Lemaachi, Salahidine and Lettieri, Chris and Masson, Charles and Miccinilli, Hugo and Ram{\'e}, Elise and Ren, Qiqi and Rostamizadeh, Afshin and others},
  journal={arXiv preprint arXiv:2505.14766},
  year={2025}
}

@article{ayappa2008validation,
  title={Validation of a self-applied unattended monitor for sleep disordered breathing},
  author={Ayappa, Indu and Norman, Robert G and Seelall, Vijay and Rapoport, David M},
  journal={Journal of Clinical Sleep Medicine},
  volume={4},
  number={1},
  pages={26--37},
  year={2008},
  publisher={American Academy of Sleep Medicine}
}

@article{zhang2018national,
  title={The National Sleep Research Resource: towards a sleep data commons},
  author={Zhang, Guo-Qiang and Cui, Licong and Mueller, Remo and Tao, Shiqiang and Kim, Matthew and Rueschman, Michael and Mariani, Sara and Mobley, Daniel and Redline, Susan},
  journal={Journal of the American Medical Informatics Association},
  volume={25},
  number={10},
  pages={1351--1358},
  year={2018},
  publisher={Oxford University Press}
}

@inproceedings{ilse2018attention,
  title={Attention-based deep multiple instance learning},
  author={Ilse, Maximilian and Tomczak, Jakub and Welling, Max},
  booktitle={International conference on machine learning},
  pages={2127--2136},
  year={2018},
  organization={PMLR}
}

@article{perslev2021u,
  title={U-Sleep: resilient high-frequency sleep staging},
  author={Perslev, Mathias and Darkner, Sune and Kempfner, Lykke and Nikolic, Miki and Jennum, Poul J{\o}rgen and Igel, Christian},
  journal={NPJ digital medicine},
  volume={4},
  number={1},
  pages={72},
  year={2021},
  publisher={Nature Publishing Group UK London}
}

@article{sands2018quantifying,
  title={Quantifying the arousal threshold using polysomnography in obstructive sleep apnea},
  author={Sands, Scott A and Terrill, Philip I and Edwards, Bradley A and Taranto Montemurro, Luigi and Azarbarzin, Ali and Marques, Melania and De Melo, Camila M and Loring, Stephen H and Butler, James P and White, David P and others},
  journal={Sleep},
  volume={41},
  number={1},
  pages={zsx183},
  year={2018},
  publisher={Oxford University Press US}
}

@article{gong2025leveraging,
  title={Leveraging clinical sleep data across multiple pediatric cohorts for insights into neurodevelopment: the retrospective analysis of sleep in Pediatric (RASP) cohorts study},
  author={Gong, Naihua N and Mahat, Aditya and Ahmad, Samya and Glaze, Daniel and Maletic-Savatic, Mirjana and McGinley, Matthew and Morse, Anne Marie and Rodriguez, Alcibiades J and Thurm, Audrey and Redline, Susan and others},
  journal={Sleep},
  pages={zsaf157},
  year={2025},
  publisher={Oxford University Press}
}

@article{zhai2024challenges,
  title={Challenges and opportunities of deep learning for wearable-based objective sleep assessment},
  author={Zhai, Bing and Elder, Greg J and Godfrey, Alan},
  journal={npj Digital Medicine},
  volume={7},
  number={1},
  pages={85},
  year={2024},
  publisher={Nature Publishing Group UK London}
}

@article{jia2021multi,
  title={Multi-view spatial-temporal graph convolutional networks with domain generalization for sleep stage classification},
  author={Jia, Ziyu and Lin, Youfang and Wang, Jing and Ning, Xiaojun and He, Yuanlai and Zhou, Ronghao and Zhou, Yuhan and Lehman, Li-wei H},
  journal={IEEE Transactions on Neural Systems and Rehabilitation Engineering},
  volume={29},
  pages={1977--1986},
  year={2021},
  publisher={IEEE}
}

@article{kjaer2025stanford,
  title={Stanford Sleep Bench: Evaluating Polysomnography Pre-training Methods for Sleep Foundation Models},
  author={Kjaer, Magnus Ruud and Thapa, Rahul and Ganjoo, Gauri and Moore IV, Hyatt and Jennum, Poul Joergen and Westover, Brandon M and Zou, James and Mignot, Emmanuel and He, Bryan and Brink-Kjaer, Andreas},
  journal={arXiv preprint arXiv:2512.09591},
  year={2025}
}

@article{hochreiter1997long,
  title={Long short-term memory},
  author={Hochreiter, Sepp and Schmidhuber, J{\"u}rgen},
  journal={Neural computation},
  volume={9},
  number={8},
  pages={1735--1780},
  year={1997},
  publisher={MIT press}
}

@article{liu2023self,
  title={Self-supervised contrastive learning for medical time series: A systematic review},
  author={Liu, Ziyu and Alavi, Azadeh and Li, Minyi and Zhang, Xiang},
  journal={Sensors},
  volume={23},
  number={9},
  pages={4221},
  year={2023},
  publisher={MDPI}
}

@article{arnal2020dreem,
  title={The Dreem Headband compared to polysomnography for electroencephalographic signal acquisition and sleep staging},
  author={Arnal, Pierrick J and Thorey, Valentin and Debellemaniere, Eden and Ballard, Michael E and Bou Hernandez, Albert and Guillot, Antoine and Jourde, Hugo and Harris, Mason and Guillard, Mathias and Van Beers, Pascal and others},
  journal={Sleep},
  volume={43},
  number={11},
  pages={zsaa097},
  year={2020},
  publisher={Oxford University Press US}
}

@article{redline2014sleep,
  title={Sleep-disordered breathing in Hispanic/Latino individuals of diverse backgrounds. The Hispanic community health study/study of Latinos},
  author={Redline, Susan and Sotres-Alvarez, Daniela and Loredo, Jose and Hall, Martica and Patel, Sanjay R and Ramos, Alberto and Shah, Neomi and Ries, Andrew and Arens, Raanan and Barnhart, Janice and others},
  journal={American journal of respiratory and critical care medicine},
  volume={189},
  number={3},
  pages={335--344},
  year={2014},
  publisher={American Thoracic Society}
}

@article{gleason2014challenges,
  title={Challenges in recruitment to a randomized controlled study of cardiovascular disease reduction in sleep apnea: an analysis of alternative strategies},
  author={Gleason, Kevin and Shin, Donghoon and Rueschman, Michael and Weinstock, Tanya and Wang, Rui and Ware, James H and Mittleman, Murray A and Redline, Susan},
  journal={Sleep},
  volume={37},
  number={12},
  pages={2035--2038},
  year={2014},
  publisher={Oxford University Press}
}

@article{quan1997sleep,
  title={The sleep heart health study: design, rationale, and methods},
  author={Quan, Stuart F and Howard, Barbara V and Iber, Conrad and Kiley, James P and Nieto, F Javier and O'Connor, George T and Rapoport, David M and Redline, Susan and Robbins, John and Samet, Jonathan M and others},
  journal={Sleep},
  volume={20},
  number={12},
  pages={1077--1085},
  year={1997},
  publisher={Oxford University Press}
}

@article{lee2022large,
  title={A large collection of real-world pediatric sleep studies},
  author={Lee, Harlin and Li, Boyue and DeForte, Shelly and Splaingard, Mark L and Huang, Yungui and Chi, Yuejie and Linwood, Simon L},
  journal={Scientific Data},
  volume={9},
  number={1},
  pages={421},
  year={2022},
  publisher={Nature Publishing Group UK London}
}

@article{young2009burden,
  title={Burden of sleep apnea: rationale, design, and major findings of the Wisconsin Sleep Cohort study},
  author={Young, Terry and Palta, Mari and Dempsey, Jerome and Peppard, Paul E and Nieto, F Javier and Hla, K Mae},
  journal={WMJ: official publication of the State Medical Society of Wisconsin},
  volume={108},
  number={5},
  pages={246},
  year={2009}
}

@article{rosen2003prevalence,
  title={Prevalence and risk factors for sleep-disordered breathing in 8-to 11-year-old children: association with race and prematurity},
  author={Rosen, Carol L and Larkin, Emma K and Kirchner, H Lester and Emancipator, Judith L and Bivins, Sarah F and Surovec, Susan A and Martin, Richard J and Redline, Susan},
  journal={The Journal of pediatrics},
  volume={142},
  number={4},
  pages={383--389},
  year={2003},
  publisher={Elsevier}
}

@article{redline1995familial,
  title={The familial aggregation of obstructive sleep apnea.},
  author={Redline, Susan and Tishler, Peter V and Tosteson, Tor D and Williamson, John and Kump, Kenneth and Browner, Ilene and Ferrette, Veronica and Krejci, Patrick},
  journal={American journal of respiratory and critical care medicine},
  volume={151},
  number={3},
  pages={682--687},
  year={1995},
  publisher={American Public Health Association}
}

@article{blackwell2011associations,
  title={Associations between sleep architecture and sleep-disordered breathing and cognition in older community-dwelling men: the osteoporotic fractures in men sleep study},
  author={Blackwell, Terri and Yaffe, Kristine and Ancoli-Israel, Sonia and Redline, Susan and Ensrud, Kristine E and Stefanick, Marcia L and Laffan, Alison and Stone, Katie L and Osteoporotic Fractures in Men Study Group},
  journal={Journal of the American Geriatrics Society},
  volume={59},
  number={12},
  pages={2217--2225},
  year={2011},
  publisher={Wiley Online Library}
}

@article{chen2015racial,
  title={Racial/ethnic differences in sleep disturbances: the Multi-Ethnic Study of Atherosclerosis (MESA)},
  author={Chen, Xiaoli and Wang, Rui and Zee, Phyllis and Lutsey, Pamela L and Javaheri, Sogol and Alc{\'a}ntara, Carmela and Jackson, Chandra L and Williams, Michelle A and Redline, Susan},
  journal={Sleep},
  volume={38},
  number={6},
  pages={877--888},
  year={2015},
  publisher={Oxford University Press}
}

@article{marcus2013randomized,
  title={A randomized trial of adenotonsillectomy for childhood sleep apnea},
  author={Marcus, Carole L and Moore, Rene{\'e} H and Rosen, Carol L and Giordani, Bruno and Garetz, Susan L and Taylor, H Gerry and Mitchell, Ron B and Amin, Raouf and Katz, Eliot S and Arens, Raanan and others},
  journal={New England Journal of Medicine},
  volume={368},
  number={25},
  pages={2366--2376},
  year={2013},
  publisher={Mass Medical Soc}
}

@article{spira2008sleep,
  title={Sleep-disordered breathing and cognition in older women},
  author={Spira, Adam P and Blackwell, Terri and Stone, Katie L and Redline, Susan and Cauley, Jane A and Ancoli-Israel, Sonia and Yaffe, Kristine},
  journal={Journal of the American Geriatrics Society},
  volume={56},
  number={1},
  pages={45--50},
  year={2008},
  publisher={Wiley Online Library}
}

@article{xu2026sleeplm,
  title={SleepLM: Natural-Language Intelligence for Human Sleep},
  author={Xu, Zongzhe and Shuai, Zitao and Mozaffari, Eideen and Aysola, Ravi S and Kumar, Rajesh and Yang, Yuzhe},
  journal={arXiv preprint arXiv:2602.23605},
  year={2026}
}

@article{huang2026unified,
  title={A unified time-frequency foundation model for sleep decoding},
  author={Huang, Weixuan and Wang, Yan and Cheng, Hanrong and Xu, Wei and Li, Tingyue and Wu, Xiuwen and Xu, Hui and Liao, Pan and Cui, Zaixu and Zou, Qihong and others},
  journal={Nature Communications},
  year={2026},
  publisher={Nature Publishing Group UK London}
}

@article{li2026hearts,
  title={HEARTS: Benchmarking LLM Reasoning on Health Time Series},
  author={Li, Sirui and Xiao, Shuhan and Joshi, Mihir and Metwally, Ahmed and McDuff, Daniel and Wang, Wei and Yang, Yuzhe},
  journal={arXiv preprint arXiv:2603.06638},
  year={2026}
}

@inproceedings{yue2022ts2vec,
  title={Ts2vec: Towards universal representation of time series},
  author={Yue, Zhihan and Wang, Yujing and Duan, Juanyong and Yang, Tianmeng and Huang, Congrui and Tong, Yunhai and Xu, Bixiong},
  booktitle={Proceedings of the AAAI conference on artificial intelligence},
  volume={36},
  pages={8980--8987},
  year={2022}
}

@article{na2024guiding,
  title={Guiding masked representation learning to capture spatio-temporal relationship of electrocardiogram},
  author={Na, Yeongyeon and Park, Minje and Tae, Yunwon and Joo, Sunghoon},
  journal={arXiv preprint arXiv:2402.09450},
  year={2024}
}

@article{xu2025lsm,
  title={Lsm-2: Learning from incomplete wearable sensor data},
  author={Xu, Maxwell A and Narayanswamy, Girish and Ayush, Kumar and Spathis, Dimitris and Liao, Shun and Tailor, Shyam A and Metwally, Ahmed and Heydari, A Ali and Zhang, Yuwei and Garrison, Jake and others},
  journal={arXiv preprint arXiv:2506.05321},
  year={2025}
}
\bibliographystyle{plain}

\newpage
\appendix
\section{Implementation Details}
\label{appendix:impl_details}
\subsection{Detailed Pre-training Setups}
\label{appendix:pretrain_setup}

In this section, we provide implementation details of our pre-training configurations for both the benchmarked baselines and our method. For our explorations, we aim to directly adapt well-established self-supervised pre-training methods, and benchmark their performance on the sleep data. In addition to the existing pre-training method \sleepfm, we have considered four representative families of SSL objectives:

(1) \textbf{Contrastive learning}: SimCLR~\citep{chen2020simple};

(2) \textbf{Reconstruction-based methods}: MAE~\citep{he2022masked} and VQ-VAE~\citep{van2017neural};

(3) \textbf{Autoregressive modeling}: Autoregression~\citep{radford2019language};

(4) \textbf{Self-distillation}: DINO~\citep{oquab2023dinov2}.

For \ours model family, we use DINO as a specification and incorporate it with our identified pre-training recipes, and form \ours used in the main experiments. We have also incorporated \ours with SimCLR, whose performance in the main evaluation tasks is shown in Table~\ref{tab:simclr_ours_dense_task}. We consider them because these invariance-based methods are a natural testbed for applying our pre-training recipes. Specifically, contrastive learning and self-distillation-based methods are invariance-based methods, which aim to align two (or more) augmented views of the same input in the latent space. As a result, their downstream performance usually depends on the approach we use to augment the raw input data.

\begin{wraptable}{r}{0.5\textwidth}
  \centering
  \caption{\textbf{Encoder backbone configurations used in our experiments.}}
  \label{tab:vit_backbones}
  \vspace{-2mm}
  \begin{adjustbox}{width=\linewidth}
  \footnotesize
  \setlength{\tabcolsep}{8pt}
  \begin{tabular}{@{}l rrrr r@{}}
    \toprule[1.5pt]
    \multirow{2}{*}{\textbf{Model}} &
    \multicolumn{4}{c}{\textbf{Architecture}} &
    \multirow{2}{*}{\textbf{Params (M)}} \\
    \cmidrule(lr){2-5}
    & Width & Depth & Heads & MLP dim & \\
    \midrule
    \textsc{ViT-1M}   & 128 & 6  & 4  & 512  &  1.7M \\
    \textsc{ViT-5M}   & 192 & 12 & 3  & 768  &  5.5M \\
    \textsc{ViT-85M}   & 768 & 12 & 12 & 3072 & 85.5M \\
    \bottomrule[1.5pt]
  \end{tabular}
  \end{adjustbox}
  \vspace{-2mm}
\end{wraptable}

\textbf{Backbone and training protocol.}
Unless otherwise specified, all methods reported in the main tables use the same ViT-85M Transformer backbone~\citep{dosovitskiy2020image}. We replace the patchify layer in the original ViT with a simple convolution layer to project the input signal into tokens, and the architecture details are in Table~\ref{tab:vit_backbones}. We pre-train all models on the same split of the pre-training cohorts in our \benchname. Each run is trained for up to 30 epochs, with early stopping when we observe training converges, based on the loss curves.

\textbf{Optimization and scheduling.}
We use AdamW~\citep{loshchilov2017decoupled} for all methods. 
For the scheduler of learning rate, we follow a standard configuration of self-supervised pre-training, and use a linear warmup for the first 10\% of total training steps, and then apply cosine annealing decay. We have tuned the learning rate and batch size for each method independently to ensure a fair comparison under its best-performing regime. We first search batch size, for the maximized usage of GPU memory; and then optimize learning rate choices based on the performance.

\textbf{Augmentations for \ours.}
For \ours, as discussed in Sec.~\ref{sec:methods}, we adopt a two-stage masking augmentation. We first apply channel masking by randomly dropping 50\% of input channels. We then apply block-wise temporal masking, where the temporal masking ratio is independently sampled for each training example from a uniform distribution over $[0.3, 0.6]$.

The complete pre-training configuration, including detailed hyper-parameters and method-specific setup, is summarized in Table~\ref{tab:pretrain_config}.

\subsection{Detailed Downstream Evaluation Setups}
\label{appendix:eval_setup}

\textbf{Evaluation setup.}
We evaluate pre-trained models on three types of downstream tasks under three standard transfer learning protocols. For epoch-level evaluation, we consider a 4-class sleep staging task, and four sleep-event detection tasks. For sleep staging, we follow~\citep{carter2025wav2sleep} and merge N1 and N2 into a single light sleep category, and predict four classes: awake, light sleep, deep sleep, and REM. For sleep-event detection, we perform 2-class classification for the four following event detection tasks: arousal, hypopnea, oxygen desaturation, and central apnea. We also consider three patient-level disease classification: coronary disease, diabetes, and hypertension. We report both AUC and AUPRC since all tasks are class-imbalanced.

For epoch-level evaluation, we consider three transfer learning settings: linear probing, full fine-tuning, and few-shot adaptation. In linear probing, we freeze the encoder and train a linear classifier on the full training split of each target dataset, then evaluate on the held-out test set. In full fine-tuning, we adopt a similar setting on data usage, and in addition to training a linear classifier, we also jointly optimize the encoder and classifier. In few-shot adaptation, we sample $K$ examples per class from the training split. In particular, we consider $K\in [1,5,50]$ in our main experiments. We keep the encoder frozen, train a linear classifier, and evaluate on the same test split across methods. 

For patient-level prediction, we segment each recording into non-overlapping 30-second epochs and extract an embedding for each epoch using the pre-trained encoder. For each pre-trained checkpoint, we pre-extract epoch embeddings for all recordings before training and evaluating on the downstream disease prediction tasks. During embedding extraction, we pad the epoch sequence length to 1200, corresponding to 10 hours. During downstream training, we consider two approaches of aggregating the epoch embeddings into a single patient-level representation: (i) we learn a lightweight aggregation module on top of epoch embeddings of each patient, or (ii) we use simple mean pooling over all epochs. We then feed the aggregated patient-level representation to a classifier for disease prediction.

Detailed configurations of evaluation tasks can be seen in Table~\ref{tab:appendix_downstream_settings}.

\begin{table*}[t]
  \centering
  \caption{\textbf{Pretraining configurations across different methods.}}
  \label{tab:pretrain_config}
  \begin{adjustbox}{width=\textwidth}
  \begin{tabular}{l *{8}{c}}
    \toprule[1.5pt]
    \textbf{Configuration} &
    \textbf{SleepFM} & \textbf{SimCLR} & \textbf{DINO} & \textbf{MAE} &
    \textbf{VQ-VAE} & \textbf{AR} & \textbf{\ours\ (SimCLR)} & \textbf{\ours\ (DINO)} \\
    \midrule

    Max Epoch & \multicolumn{8}{c}{30} \\
    Batch Size & 2560 & 3200 & 1024 & 9600 & 3200 & 3200 & 3200 & 1024 \\
    Base Learning Rate & 1.00E{-}04 & 1.00E{-}04 & 5.00E{-}05 & 3.00E{-}04 & 1.00E{-}04 & 1.00E{-}04 & 1.00E{-}04 & 5.00E{-}05 \\
    Optimizer & \multicolumn{8}{c}{AdamW} \\
    Opt.\ momentum $[\beta_1, \beta_2]$ & \multicolumn{8}{c}{$[0.9, 0.95]$} \\
    Weight Decay & 0.2 & 0.2 & 0.2 & 0.2 & 0.2 & 0.05 & 0.2 & 0.2 \\
    Gradient Clipping & -- & -- & -- & -- & 3 & -- & -- & 3 \\
    Learning Rate Scheduler & \multicolumn{8}{c}{LinearWarmup \& CosineDecay} \\
    Token window length & \multicolumn{8}{c}{64} \\

    \bottomrule[1.5pt]
  \end{tabular}
  \end{adjustbox}
\end{table*}

\begin{table*}[t]
  \centering
  \caption{\textbf{Training hyperparameters for sleep analysis tasks and disease prediction tasks.}}
  \label{tab:appendix_downstream_settings}
  \vspace{-2mm}
  \begin{adjustbox}{width=\textwidth}
  \footnotesize
  \setlength{\tabcolsep}{5pt}
  \begin{tabular}{l *{8}{c}}
    \toprule[1.5pt]
    \multirow{2}{*}{\textbf{Hyperparameter}} &
    \multicolumn{3}{c}{\textbf{Sleep Analysis Tasks}} &
    \multicolumn{5}{c}{\textbf{Disease Prediction}} \\
    \cmidrule(lr){2-4}\cmidrule(lr){5-9}
    & Linear Probing & Finetuning & Supervised &
      Mean Pool & MIL & LSTM & TopK & End2End \\
    \midrule
    Base learning rate
      & 0.1 & 1e{-4} & 1e{-3} & 1e{-2} & 5e{-3} & 5e{-3} & 5e{-3} & 1e{-3} \\

    Batch size
      & \multicolumn{3}{c}{800 per GPU} 
      & 256 per GPU
      & \multicolumn{3}{c}{128 per GPU} 
      & 16 per GPU \\

    Max step
      & \multicolumn{2}{c}{500} 
      & \multicolumn{6}{c}{--} \\

    Max epoch
      & \multicolumn{2}{c}{--} 
      & 5
      & \multicolumn{4}{c}{50} 
      & 5 \\

    Learning rate scheduler
      & \multicolumn{8}{c}{LinearWarmup \& CosineDecay} \\

    Optimizer
      & \multicolumn{8}{c}{AdamW} \\
    \bottomrule[1.5pt]
  \end{tabular}
  \end{adjustbox}
  \vspace{-2mm}
\end{table*}

\begin{figure}[!t]
    \centering
    \includegraphics[width=1\linewidth]{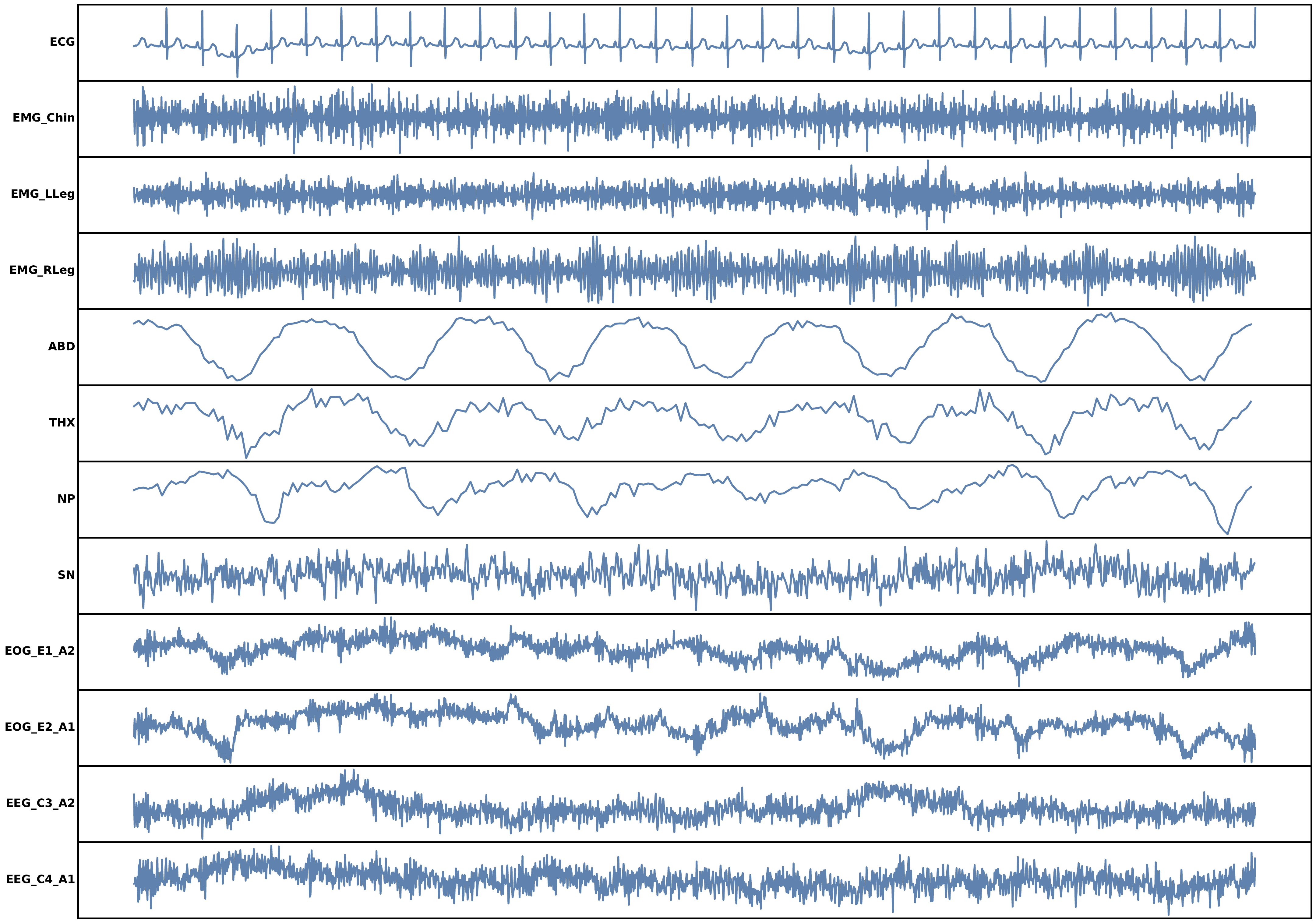}
    \caption{\textbf{Visualization of the pre-processed 12 channels of the sleep data used in our paper.} We pick a 30-sec epoch for demonstration.}
    \label{fig:sig_viz}
\end{figure}
\section{Details of \benchname}
\subsection{Dataset Statistics}
\label{appendix:data stats}

\begin{wraptable}{r}{0.5\textwidth}
  \centering
  \caption{\textbf{Comparison between \benchname and existing open-source sleep datasets.}}
  \label{tab:bench_vs_ssc}
  \vspace{-2mm}

  \begin{adjustbox}{width=\linewidth}
    \footnotesize
    \begin{tabular}{l *{2}{c}}
      \toprule[1.5pt]
      \textbf{Dataset} &
      \textbf{Num. Recordings} & \textbf{Num. Hours}\\
      \midrule
      \midrule
      SSC~\citep{kjaer2025stanford} & 17,467 & 163,000 \\
      \textsc{\benchname}          & 21,482 & 166,500 \\
      \bottomrule[1.5pt]
    \end{tabular}
  \end{adjustbox}
\end{wraptable}

To enable open benchmarking and our controlled explorations, we curated a large-scale benchmark \benchname, comprising nine diverse sleep datasets: SHHS~\citep{quan1997sleep}, NCHSDB~\citep{lee2022large}, WSC~\citep{young2009burden}, CCSHS~\citep{rosen2003prevalence}, CFS~\citep{redline1995familial}, MROS~\citep{blackwell2011associations}, MESA~\citep{chen2015racial}, CHAT~\citep{marcus2013randomized}, and SOF~\citep{spira2008sleep}. This consolidated corpus represents a significant variety of demographic profiles and sleep disorder prevalences, totaling 20M epochs of sleep recording data. Table~\ref{tab:data_stats} and Table~\ref{tab:epoch_stats} report the statistics of patients and 30-second epochs, across different datasets and data splits.

\begin{table*}[htbp]
  \centering
  \caption{\textbf{Distribution of sleep recordings across different datasets and splits.} Models are pretrained on pretrain datasets and evaluated on both pretrain and out-of-domain (OOD) datasets. The test split of pretrain data and all OOD data are unseen during pre-training.}
  \label{tab:data_stats}
  \begin{adjustbox}{width=\textwidth}
  \setlength{\tabcolsep}{10pt}
  \begin{tabular}{l rrrrr rrrr r}
    \toprule[1.5pt]
    \multirow{2}{*}{\textbf{Split}} &
    \multicolumn{5}{c}{\textbf{Pretrain Datasets}} &
    \multicolumn{4}{c}{\textbf{O.O.D. Datasets}} &
    \multirow{2}{*}{\textbf{Total}} \\
    \cmidrule(lr){2-6}\cmidrule(lr){7-10}
& SHHS & NCHSDB & WSC & CCSHS & CFS
  & MROS & MESA & CHAT & SOF & \\

    \midrule
    \midrule
    Train & 6,663 & 3,094 & 1,397 & 403 & 269 & 3,101 & 1,367 & 531 & 361 & 17,186 \\
    Valid &   811 &   392 &   182 &  59 &  34 &   385 &   174 &  66 &  45 &  2,148 \\
    Test  &   802 &   403 &   194 &  53 &  26 &   381 &   180 &  67 &  42 &  2,148 \\
    \midrule
    Total & 8,276 & 3,889 & 1,773 & 515 & 329 & 3,867 & 1,721 & 664 & 448 & 21,482 \\
    \bottomrule[1.5pt]
  \end{tabular}
  \end{adjustbox}
\end{table*}

\begin{table*}[t]
  \centering
  \caption{\textbf{Distribution of 30-second epochs across different datasets and splits.}}
  \label{tab:epoch_stats}
  \begin{adjustbox}{width=\textwidth}
  \setlength{\tabcolsep}{10pt}
  \begin{tabular}{l rrrrr rrrr r}
    \toprule[1.5pt]
    \multirow{2}{*}{\textbf{Split}} &
    \multicolumn{5}{c}{\textbf{Internal Pretrain Datasets}} &
    \multicolumn{4}{c}{\textbf{External Eval. Datasets}} &
    \multirow{2}{*}{\textbf{Total}}\\
    \cmidrule(lr){2-6}\cmidrule(lr){7-10}
& SHHS & NCHSDB & WSC & CCSHS & CFS
  & MROS & MESA & CHAT & SOF & \\

    \midrule
    \midrule
    Train & 5,919k & 2,892k & 1,270k & 432k & 255k & 3,015k & 1,312k & 562k & 341k & 15,997k \\
    Valid &   724k &   367k &   164k &  64k &  32k &   370k &   164k &  69k &  42k &  1,996k \\
    Test  &   711k &   376k &   175k &  57k &  24k &   372k &   172k &  70k &  38k &  1,995k \\
    \midrule
    Total & 7,354k & 3,634k & 1,609k & 553k & 311k & 3,757k & 1,648k & 701k & 421k & 19,989k \\
    \bottomrule[1.5pt]
  \end{tabular}
  \end{adjustbox}
\end{table*}

\begin{table*}[!htbp]
  \centering
  \caption{\textbf{Distributions of sleep event detection tasks across datasets used for sleep event evaluation.} For each sleep disorder event, we report its prevalence in each split of each used dataset.}
  \label{tab:appendix_disorder_distribution}
  \begin{adjustbox}{width=1.0\textwidth}
  \setlength{\tabcolsep}{12pt}
    \begin{tabular}{ll rrrrr}
    \toprule[1.5pt]
    Dataset & Split & Epochs & Arousal & Hypopnea & Oxygen Desat. & Central Apnea \\
    \midrule\midrule
    SHHS & Train & 5,919,217 & 17.85\% & 29.13\% & 23.46\% & 0.71\% \\
    SHHS & Valid & 724,056 & 17.73\% & 29.00\% & 23.71\% & 0.74\% \\
    SHHS & Test & 710,821 & 17.60\% & 29.00\% & 23.32\% & 0.71\% \\
    MROS & Train & 3,014,915 & 19.46\% & 12.27\% & 52.28\% & 1.32\% \\
    MROS & Valid & 370,378 & 19.55\% & 12.62\% & 52.65\% & 1.66\% \\
    MROS & Test & 371,873 & 19.77\% & 12.26\% & 51.93\% & 1.42\% \\
    \bottomrule[1.5pt]
    \end{tabular}
  \end{adjustbox}
\end{table*}
\subsection{Downstream Task Labels}
\textbf{Sleep Event Distribution.} We report the prevalence of sleep events of datasets used in main experiments, in the Table~\ref{tab:appendix_disorder_distribution}. The datasets exhibit substantial heterogeneity in event distributions, which enables the evaluation of the generalization capability of pre-trained models. During dataset splitting, we perform stratified sampling at the patient level to prevent information leakage and to keep event distributions consistent across splits. Specifically, we first randomly assign patients to the training, validation, and test splits. We then assign each epoch to the split of its corresponding patient via patient ID. As shown in Table~\ref{tab:appendix_disorder_distribution}, the data was divided into train, validate, and test sets with an approximate global ratio of 80:10:10 strictly on the subject level.

\begin{table}[htbp]
\centering
\caption{\textbf{Distribution of subject-level disease labels in the MROS dataset.} We filtered out patients with NaN values on the selected labels. Here we report the number and percentage of both negative and positive samples across splits.}
\label{tab:sleepbench_disease}
\resizebox{\textwidth}{!}{%
\setlength{\tabcolsep}{12pt}
\begin{tabular}{ll rr rr r}
\toprule[1.5pt]
\multirow{2}{*}{\textbf{Disease Category}} & \multirow{2}{*}{\textbf{Split}} & \multicolumn{2}{c}{\textbf{Negative}} & \multicolumn{2}{c}{\textbf{Positive}} & \multirow{2}{*}{\textbf{Total Samples}} \\
\cmidrule(lr){3-4} \cmidrule(lr){5-6}
 & & \textbf{Count} & \textbf{Ratio} & \textbf{Count} & \textbf{Ratio} & \\
\midrule
\midrule
\multirow{3}{*}{Coronary Disease} 
 & Train & 1,626 & 71.3\% & 654 & 28.7\% & 2,280 \\
 & Valid & 188 & 67.1\% & 92 & 32.9\% & 280 \\
 & Test  & 183 & 64.2\% & 102 & 35.8\% & 285 \\
\midrule
\multirow{3}{*}{Diabetes} 
 & Train & 1,987 & 86.9\% & 300 & 13.1\% & 2,287 \\
 & Valid & 247 & 88.2\% & 33 & 11.8\% & 280 \\
 & Test  & 248 & 87.0\% & 37 & 13.0\% & 285 \\
\midrule
\multirow{3}{*}{Hypertension} 
 & Train & 1,152 & 50.4\% & 1,134 & 49.6\% & 2,286 \\
 & Valid & 138 & 49.3\% & 142 & 50.7\% & 280 \\
 & Test  & 142 & 49.8\% & 143 & 50.2\% & 285 \\
\bottomrule[1.5pt]
\end{tabular}
}
\end{table}
\textbf{Disease Classification:} For more clinically valuable downstream tasks, we perform patient-level disease classification on the MROS dataset. We focused on three diagnosis tasks: Coronary Disease, Diabetes, and Hypertension. These labels are formed from disease-related variables provided MROS dataset. We set the disease label as positive, if the patient has self-reported related medication usage of diagnosis history. As detailed in Table~\ref{tab:sleepbench_disease}, the MROS dataset was split into training, validation, and testing sets. The three labels cover both class-balance and class-imbalance scenes, therefore more comprehensively evaluating the pre-trained model's ability on disease prediction tasks.

\subsection{Dataset Pre-processing Details} 
As mentioned in Sec.~\ref{dataset}, we utilize a standardized 12-channel montage covering 4 physiological modality groups:
\ding{182} \textit{Brain (EEG/EOG):} \texttt{C3-A2}, \texttt{C4-A1}, \texttt{E1-A2}, \texttt{E2-A1},
\ding{183} \textit{Respiration:} \texttt{Abdominal}, \texttt{Thorax}, \texttt{Nasal Pressure}, \texttt{Snore},
\ding{184} \textit{Cardiac:} \texttt{ECG}, and
\ding{185} \textit{Somatic:} \texttt{EMG-Chin}, \texttt{EMG-LLeg}, \texttt{EMG-RLeg}. The detailed channel descriptions are in Table~\ref{tab:channel_selection}.

For the pre-processing of the full-night recordings, we follow~\citep{carter2025wav2sleep}, and first resample each channel to a unified sampling frequency, as detailed in Table~\ref{tab:channel_selection}. Next, we conduct manual quality control by trimming prolonged wake periods at the beginning and end of the night, which also removes extreme artifacts introduced during sensor setup and removal. We then apply per-night z-score normalization to each channel, followed by clipping the normalized signals to the range $[-6,6]$ to suppress outliers. The clipping thresholds are chosen based on the empirical distribution of z-scored signals. For respiratory channels, we additionally apply area-dependent z-score normalization to reduce amplitude drift and improve cross-night consistency.

We follow prior work and segment each night into non-overlapping 30-second epochs, yielding roughly \textbf{20 million} epochs for self-supervised pre-training. We resample all channels in the segmented epochs to 64 Hz and zero-pad missing channels. The pre-processed data in \benchname is demonstrated in Fig.~\ref{fig:sig_viz}.

\label{appendix:channel details}
\begin{table*}[t]
  \centering
  \caption{\textbf{Channel configuration used for main experiments.} During post-processing, they are resampled to 64 Hz.}
  \label{tab:channel_selection}
  \vspace{-2mm}
  \begin{adjustbox}{width=\textwidth}
  \footnotesize
  \begin{tabular}{@{}l l l l r@{}}
    \toprule[1.5pt]
    \textbf{Group} & \textbf{Channel}  & \textbf{Full name} & \textbf{Electrode Detail} & \textbf{Preprocessing Frequency}\\
    \midrule
    \midrule
    \multirow{1}{*}{\textsc{\textbf{Cardiac}}} &
      \textbf{ECG} &  Electrocardiography & ECG left - ECG right & 128 Hz\\
    \midrule
    \multirow{3}{*}{\textsc{\textbf{Somatic}}} &
      \multirow{3}{*}{\textbf{EMG}} & \multirow{3}{*}{Electromyography} & EMG\_Chin & 64 Hz\\
 & & & EMG from left leg& 64 Hz\\
& & & EMG from right leg& 64 Hz\\
    \midrule
    \multirow{4}{*}{\textsc{\textbf{Resp.\ Effort}}} &
      \textbf{ABD} & Abdominal respiratory effort & & 8 Hz\\
    & \textbf{THX} & Thoracic respiratory effort && 8 Hz\\
    & \textbf{NP} & Nasal pressure & & 8 Hz\\
    & \textbf{SN} & Snore & & 32 Hz\\
    \midrule

    \multirow{4}{*}{\textsc{\textbf{Brain Activity}}} &
      \multirow{2}{*}{\textbf{EOG}} & \multirow{2}{*}{Electrooculography} & EOG E1 - A2& 64 Hz\\
    & & & EOG E2 - A1 & 64 Hz\\
     &
       \multirow{2}{*}{\textbf{EEG}} &  \multirow{2}{*}{Electroencephalography} & EEG C3 - A2& 64 Hz\\
    &  &  & EEG C4 - A1& 64 Hz\\
    \bottomrule[1.5pt]
  \end{tabular}
  \end{adjustbox}
  \vspace{-2mm}
\end{table*}

\section{Additional Results and Analysis}
\subsection{Additional Results of Sleep Analysis Tasks}
\label{appendix:main_downstream}

\textbf{\ours achieves state-of-the-art on sleep analysis tasks on in-domain evaluation set.} We also conduct evaluation on sleep analysis tasks on in domain dataset SHHS. As shown in Table~\ref{tab:shhs_dense_task}. \ours achieves the best performance under both linear probing and full fine-tuning. The same trend also holds on the out-of-domain cohorts reported in the main text, further supporting the superiority of our method. 

\begin{table*}[t]
  \centering
  \caption{\textbf{Sleep staging and sleep event detection results on SHHS}. \ours achieves overall best performance among all compared methods.}
  \label{tab:shhs_dense_task}
  \vspace{-2mm}
  \begin{adjustbox}{width=\textwidth}
  \footnotesize
  \setlength{\tabcolsep}{5pt}
  \begin{tabular}{c l *{5}{cc}}
    \toprule[1.5pt]
    \multirow{2}{*}{\textbf{Eval}} & \multirow{2}{*}{\textbf{Method}} &
    \multicolumn{2}{c}{\textbf{Sleep Staging}} &
    \multicolumn{2}{c}{\textbf{Arousal}} &
    \multicolumn{2}{c}{\textbf{Hypopnea}} &
    \multicolumn{2}{c}{\textbf{Ox. Desat.}} &
    \multicolumn{2}{c}{\textbf{Central Apnea}} \\
    \cmidrule(lr){3-4}\cmidrule(lr){5-6}\cmidrule(lr){7-8}\cmidrule(lr){9-10}\cmidrule(lr){11-12}
& & AUC$^\uparrow$ & AUPRC$^\uparrow$
  & AUC$^\uparrow$ & AUPRC$^\uparrow$
  & AUC$^\uparrow$ & AUPRC$^\uparrow$
  & AUC$^\uparrow$ & AUPRC$^\uparrow$
  & AUC$^\uparrow$ & AUPRC$^\uparrow$ \\

    \midrule
    \midrule
    \multicolumn{2}{c}{\textsc{Supervised ViT}} & 97.0 & 92.3 & 88.7 & 82.3 & 79.1 & 74.8 & 76.0 & 70.1 & 95.2 & 61.7 \\
    \midrule
    \multirow{8}{*}{\textbf{LP}} & \textsc{SleepFM} & 96.6 & 91.1 &  \underline{92.2} & 86.3 & 82.0 & 77.0 & \underline{80.6} & \underline{75.0} & 95.9 & 64.7 \\
     & \textsc{SimCLR} & 86.1 & 67.4 & 71.9 & 64.2 & 64.9 & 60.7 & 62.9 & 58.5 & 74.6 & 51.3 \\
     & \textsc{DINO} & 95.1 & 87.4 & 87.0 & 79.9 & 78.3 & 73.7 & 76.5 & 70.5 & 94.8 & 60.6 \\
     & \textsc{VQ-VAE} & 96.7 & 91.3 & 89.4 & 83.0 & 80.2 & 75.4 & 78.0 & 72.2 & 95.7 & 63.6 \\
     & \textsc{MAE} & \underline{96.9} & \underline{91.9} & 91.4 & 85.7 & \textbf{83.0} & \textbf{78.4} & 80.5 & 75.0 & \underline{96.7} & \underline{66.0} \\
     & \textsc{AR} & \underline{96.9} & 91.8 & 90.6 & \underline{88.4} & 79.0 & 73.3 & 78.6 & 72.9 & 94.4 & 61.2 \\
     \rowcolor{gray!15}
     & \textsc{\ours} & \textbf{97.5} & \textbf{93.2} & \textbf{94.0} & \textbf{89.3} & \underline{82.8} & \underline{78.1} & \textbf{81.3} & \textbf{75.6} & \textbf{96.9} & \textbf{66.5} \\
    \midrule

    \multirow{8}{*}{\textbf{\textsc{FT}}} & \textsc{SleepFM} & 97.0 & 92.0 & \underline{95.1} & \underline{91.0} & 86.7 & 83.3 & 82.5 & 77.1 & 97.1 & 68.6 \\
     & \textsc{SimCLR} & 95.4 & 88.3 & 85.8 & 78.4 & 80.3 & 76.3 & 75.5 & 69.6 & 95.7 & 64.1 \\
     & \textsc{DINO} & 97.1 & 92.4 & 93.8 & 89.3 & 86.2 & 83.0 & 81.8 & 76.1 & 97.4 & \underline{69.0} \\
     & \textsc{VQ-VAE} & \underline{97.5} & \underline{93.4} & 94.7 & 90.7 & 85.5 & 82.2 & 81.7 & 76.2 & 97.2 & 68.4 \\
     & \textsc{MAE} & 97.4 & 93.0 & 94.6 & 90.6 & \underline{87.1} & \underline{84.2} & \underline{83.2} & \underline{77.9} & \textbf{97.7} & 68.5 \\
     & \textsc{AR} & 97.4 & 93.1 & 94.2 & 90.0 & 86.4 & 83.3 & 82.4 & 77.1 & 97.2 & 68.2 \\
     \rowcolor{gray!15}
     & \textsc{\ours} & \textbf{97.9} & \textbf{94.3} & \textbf{95.7} & \textbf{92.1} & \textbf{87.6} & \textbf{84.6} & \textbf{83.7} & \textbf{78.6} & \underline{97.6} & \textbf{69.8} \\
    \bottomrule[1.5pt]
  \end{tabular}
  \end{adjustbox}
  \vspace{-2mm}
\end{table*}

\begin{table*}[t]
  \centering
  \caption{\textbf{Sleep staging and sleep event detection results on SHHS.} SimCLR can also be upgraded by our pre-training recipts and achieve better performance.}
  \label{tab:simclr_ours_dense_task}
  \vspace{-2mm}
  \begin{adjustbox}{width=\textwidth}
  \footnotesize
  \setlength{\tabcolsep}{5pt}
  \begin{tabular}{c l *{5}{cc}}
    \toprule[1.5pt]
    \multirow{2}{*}{Eval} & \multirow{2}{*}{Method} &
    \multicolumn{2}{c}{Sleep Staging} &
    \multicolumn{2}{c}{Arousal} &
    \multicolumn{2}{c}{Hypopnea} &
    \multicolumn{2}{c}{Ox. Desat.} &
    \multicolumn{2}{c}{Central Apnea} \\
    \cmidrule(lr){3-4}\cmidrule(lr){5-6}\cmidrule(lr){7-8}\cmidrule(lr){9-10}\cmidrule(lr){11-12}
& & AUC$^\uparrow$ & AUPRC$^\uparrow$
  & AUC$^\uparrow$ & AUPRC$^\uparrow$
  & AUC$^\uparrow$ & AUPRC$^\uparrow$
  & AUC$^\uparrow$ & AUPRC$^\uparrow$
  & AUC$^\uparrow$ & AUPRC$^\uparrow$ \\
    \midrule
    \midrule
    & \textsc{SimCLR} & 86.1 & 67.4 & 71.9 & 64.2 & 64.9 & 60.7 & 62.9 & 58.5 & 74.6 & 51.3 \\
    \rowcolor{gray!15}
     & \textsc{\ours}\ (\textsc{SimCLR}) & 97.3 & 92.9 & 92.9 & 87.4 & 82.2 & 77.2 & 80.4 & 74.7 & 96.3 & 64.8 \\
    \bottomrule[1.5pt]
  \end{tabular}
  \end{adjustbox}
  \vspace{-2mm}
\end{table*}

\label{appendix:main few-shot}
\textbf{\ours achieves state-of-the-art on few-shot adaptation tasks.} We conduct few-shot learning evaluations on the MROS dataset, on three sleep event detection tasks and the sleep staging task, across 1-shot, 5-shot, and 50-shot settings. \ours achieves the overall best performance under 1-shot and 5-shot evaluation (Tables~\ref{tab:mros_fewshot_k1} and~\ref{tab:mros_fewshot_k5}), and performs best on all tasks under 50-shot evaluation (Table~\ref{tab:mros_fewshot_k50}).

\begin{table*}[t]
  \centering
  \caption{\textbf{1-shot few-shot results on MROS.} \ours shows better adaptability to unseen out-of-domain data. }
  \label{tab:mros_fewshot_k1}
  \vspace{-2mm}
  \begin{adjustbox}{width=\textwidth}
  \footnotesize
  \setlength{\tabcolsep}{5pt}
  \begin{tabular}{c l *{4}{cc}}
    \toprule[1.5pt]
    \multirow{2}{*}{\textbf{Few-shot}} & \multirow{2}{*}{\textbf{Method}} &
    \multicolumn{2}{c}{\textbf{Sleep Staging}} &
    \multicolumn{2}{c}{\textbf{Arousal}} &
    \multicolumn{2}{c}{\textbf{Hypopnea}} &
    \multicolumn{2}{c}{\textbf{Ox. Desat.}} \\
    \cmidrule(lr){3-4}\cmidrule(lr){5-6}\cmidrule(lr){7-8}\cmidrule(lr){9-10}
    & & AUC$^\uparrow$ & AUPRC$^\uparrow$
      & AUC$^\uparrow$ & AUPRC$^\uparrow$
      & AUC$^\uparrow$ & AUPRC$^\uparrow$
      & AUC$^\uparrow$ & AUPRC$^\uparrow$ \\
    \midrule
    \midrule
    \multirow{8}{*}{\textsc{$k$=1}}
      & \textsc{SleepFM} & 59.6 & 32.3 & \underline{56.6} & \underline{53.8} & 49.3 & 49.8 & 55.2 & 54.1 \\
      & \textsc{SimCLR}  & 50.8 & 25.1 & 51.2 & 50.6 & 49.9 & 50.0 & 50.9 & 50.8 \\
      & \textsc{DINO}    & 67.0 & 41.6 & 52.7 & 51.2 & 45.8 & 48.7 & \textbf{60.2} & \textbf{58.9} \\
      & \textsc{MAE}    & 63.0 & 36.6 & 49.3 & 50.0 & 49.1 & 49.9 & 54.4 & 53.4 \\
      & \textsc{VQ-VAE}  & 62.6 & 33.7 & 56.0 & 53.3 & \underline{50.8} & \underline{50.3} & 55.8 & 54.7 \\
      & \textsc{AR}      & \underline{67.6} & \underline{43.4} & 48.0 & 49.4 & 47.4 & 49.4 & 49.3 & 49.4 \\
      \rowcolor{gray!15}
      & \textsc{\ours} & \textbf{75.9} & \textbf{48.9} & \textbf{61.4} & \textbf{57.0} & \textbf{55.4} & \textbf{51.8} & \underline{56.6} & \underline{55.7} \\
    \bottomrule[1.5pt]
  \end{tabular}
  \end{adjustbox}
  \vspace{-2mm}
\end{table*}

\begin{table*}[t]
  \centering
  \caption{\textbf{5-shot few-shot results on MROS.} \ours shows better adaptability to unseen out-of-domain data. }
  \label{tab:mros_fewshot_k5}
  \vspace{-2mm}
  \begin{adjustbox}{width=\textwidth}
  \footnotesize
  \setlength{\tabcolsep}{5pt}
  \begin{tabular}{c l *{4}{cc}}
    \toprule[1.5pt]
    \multirow{2}{*}{\textbf{Few-shot}} & \multirow{2}{*}{\textbf{Method}} &
    \multicolumn{2}{c}{\textbf{Sleep Staging}} &
    \multicolumn{2}{c}{\textbf{Arousal}} &
    \multicolumn{2}{c}{\textbf{Hypopnea}} &
    \multicolumn{2}{c}{\textbf{Ox. Desat.}} \\
    \cmidrule(lr){3-4}\cmidrule(lr){5-6}\cmidrule(lr){7-8}\cmidrule(lr){9-10}
    & & AUC$^\uparrow$ & AUPRC$^\uparrow$
      & AUC$^\uparrow$ & AUPRC$^\uparrow$
      & AUC$^\uparrow$ & AUPRC$^\uparrow$
      & AUC$^\uparrow$ & AUPRC$^\uparrow$ \\
    \midrule
    \midrule
    \multirow{8}{*}{\textsc{$k$=5}}
      & \textsc{SleepFM} & 71.8 & 45.3 & 55.5 & 53.0 & 49.3 & 49.8 & \underline{59.8} & \underline{58.6} \\
      & \textsc{SimCLR}  & 53.0 & 27.1 & 54.2 & 52.2 & 50.0 & 50.0 & 50.7 & 50.5 \\
      & \textsc{DINO}    & \underline{82.8} & 55.8 & 56.8 & 53.3 & 47.7 & 49.2 & 57.8 & 56.1 \\
      & \textsc{MAE}    & 82.6 & 59.2 & 56.7 & 53.7 & 49.7 & 49.9 & \textbf{62.9} & \textbf{61.2} \\
      & \textsc{VQ-VAE}  & 78.1 & 51.8 & 52.2 & 51.1 & \underline{51.1} & \underline{50.4} & 58.9 & 57.6 \\
      & \textsc{AR}      & 81.9 & \underline{60.1} & \underline{57.1} & \underline{54.0} & 50.9 & 50.4 & 55.1 & 53.9 \\
      \rowcolor{gray!15}
      & \textsc{\ours} & \textbf{89.2} & \textbf{70.1} & \textbf{67.9} & \textbf{61.3} & \textbf{55.9} & \textbf{52.0} & 56.4 & 55.7 \\
    \bottomrule[1.5pt]
  \end{tabular}
  \end{adjustbox}
  \vspace{-2mm}
\end{table*}

\begin{table*}[t]
  \centering
  \caption{\textbf{50-shot few-shot results on MROS.} \ours shows better adaptability to unseen out-of-domain data on all tasks. }
  \label{tab:mros_fewshot_k50}
  \vspace{-2mm}
  \begin{adjustbox}{width=\textwidth}
  \footnotesize
  \setlength{\tabcolsep}{5pt}
  \begin{tabular}{c l *{4}{cc}}
    \toprule[1.5pt]
    \multirow{2}{*}{\textbf{Few-shot}} & \multirow{2}{*}{\textbf{Method}} &
    \multicolumn{2}{c}{\textbf{Sleep Staging}} &
    \multicolumn{2}{c}{\textbf{Arousal}} &
    \multicolumn{2}{c}{\textbf{Hypopnea}} &
    \multicolumn{2}{c}{\textbf{Ox. Desat.}} \\
    \cmidrule(lr){3-4}\cmidrule(lr){5-6}\cmidrule(lr){7-8}\cmidrule(lr){9-10}
    & & AUC$^\uparrow$ & AUPRC$^\uparrow$
      & AUC$^\uparrow$ & AUPRC$^\uparrow$
      & AUC$^\uparrow$ & AUPRC$^\uparrow$
      & AUC$^\uparrow$ & AUPRC$^\uparrow$ \\
    \midrule
    \midrule
    \multirow{8}{*}{\textsc{$k$=50}}
      & \textsc{SleepFM} & 85.9 & 64.0 & 72.3 & 65.8 & 56.3 & 52.3 & 61.7 & 60.3 \\
      & \textsc{SimCLR}  & 56.6 & 29.1 & 55.8 & 53.1 & 50.6 & 50.2 & 49.5 & 49.6 \\
      & \textsc{DINO}    & 87.4 & 67.0 & 72.1 & 64.5 & 57.2 & 52.7 & 63.5 & 61.3 \\
      & \textsc{MAE}    & \underline{92.1} & \underline{79.6} & \underline{76.5} & \underline{68.9} & \underline{60.2} & \underline{53.6} & \underline{66.8} & \underline{65.1} \\
      & \textsc{VQ-VAE}  & 85.3 & 63.3 & 68.6 & 62.3 & 51.5 & 50.5 & 65.4 & 63.7 \\
      & \textsc{AR}      & 90.8 & 75.4 & 72.0 & 64.8 & 58.8 & 53.2 & 63.0 & 61.4 \\
      \rowcolor{gray!15}
      & \textsc{\ours} & \textbf{94.5} & \textbf{82.5} & \textbf{83.9} & \textbf{77.3} & \textbf{62.9} & \textbf{54.9} & \textbf{68.4} & \textbf{66.7} \\
    \bottomrule[1.5pt]
  \end{tabular}
  \end{adjustbox}
  \vspace{-2mm}
\end{table*}

\begin{table*}[t]
  \centering
  \caption{\textbf{Comparison with domain-specific pretraining methods.} OSF outperforms baselines tailored for physiological signals.}
  \label{tab:full_domain_specific_baselines}
  \vspace{-2mm}
  \begin{adjustbox}{width=\textwidth}
  \footnotesize
  \setlength{\tabcolsep}{5pt}
  \begin{tabular}{l *{8}{c}}
    \toprule[1.5pt]
    \multirow{3}{*}{\textbf{Method}} &
    \multicolumn{4}{c}{\textbf{SHHS}} &
    \multicolumn{4}{c}{\textbf{MROS}} \\
    \cmidrule(lr){2-5}\cmidrule(lr){6-9}
    & \multicolumn{2}{c}{\textbf{Sleep Staging}} & \multicolumn{2}{c}{\textbf{Hypopnea}} &
      \multicolumn{2}{c}{\textbf{Sleep Staging}} & \multicolumn{2}{c}{\textbf{Hypopnea}} \\
    \cmidrule(lr){2-3}\cmidrule(lr){4-5}\cmidrule(lr){6-7}\cmidrule(lr){8-9}
    & AUC $^\uparrow$ & AUPRC $^\uparrow$ & AUC $^\uparrow$ & AUPRC $^\uparrow$
    & AUC $^\uparrow$ & AUPRC $^\uparrow$ & AUC $^\uparrow$ & AUPRC $^\uparrow$ \\
    \midrule

    \textsc{TS2Vec}~\citep{yue2022ts2vec} & 90.2 & 76.7 & 68.6 & 64.3 & 86.9 & 67.3 & 61.9 & 54.5 \\
    \textsc{ST-MEM}~\citep{na2024guiding} & 97.0 & 92.2 & 81.3 & 76.3 & 96.1 & 87.5 & 75.4 & 60.0 \\
    \textsc{LSM-2}~\citep{xu2025lsm} & 95.9 & 89.6 & 80.3 & 75.4 & 95.8 & 86.7 & 74.6 & 60.1 \\
    \textsc{PedSleepMAE}~\citep{pandey2024pedsleepmae} & 89.2 & 73.4 & 64.8 & 61.1 & 84.0 & 59.0 & 59.4 & 53.4 \\
    \textsc{SleepGPT}~\citep{huang2026unified} & 96.2 & 90.0 & 71.3 & 66.4 & 95.4 & 85.7 & 65.1 & 55.7 \\
    \rowcolor{gray!15}
    \textsc{\ours} & \textbf{97.5} & \textbf{93.2} & \textbf{82.8} & \textbf{78.1}& \textbf{97.3} & \textbf{90.4} & \textbf{77.7} & \textbf{62.2} \\

    \bottomrule[1.5pt]
  \end{tabular}
  \end{adjustbox}
  \vspace{-2mm}
\end{table*}

\begin{wraptable}{r}{0.45\textwidth}
  \centering
  \caption{\textbf{Disease prediction results on SHHS.} \ours outperforms the baseline method on most metrics.}
  \label{tab:shhs_disease}
  \vspace{-2mm}
  \begin{adjustbox}{width=\linewidth}
  \footnotesize
  \setlength{\tabcolsep}{4pt}
  \begin{tabular}{l cc cc}
    \toprule[1.5pt]
    \multirow{2}{*}{\textbf{Method}} &
    \multicolumn{2}{c}{\textbf{Diabetes}} &
    \multicolumn{2}{c}{\textbf{Hypertension}} \\
    \cmidrule(lr){2-3}\cmidrule(lr){4-5}
    & AUC$^\uparrow$ & AUPRC$^\uparrow$
    & AUC$^\uparrow$ & AUPRC$^\uparrow$ \\
    \midrule
    \midrule
    \textsc{SleepFM} & 79.5 & 56.5 & 76.7 & \textbf{76.1} \\
    \textsc{\ours}  & \textbf{79.8} & \textbf{58.1} & \textbf{76.8} & 74.7 \\
    \bottomrule[1.5pt]
  \end{tabular}
  \end{adjustbox}
  \vspace{-2mm}
\end{wraptable}

\textbf{OSF outperforms physiological-signal-specific pre-training methods.}
To assess whether methods tailored to physiological signals transfer effectively to sleep foundation modeling, we further compare \ours with five recent domain-specific baselines: TS2Vec~\citep{yue2022ts2vec}, ST-MEM~\citep{na2024guiding}, LSM-2~\citep{xu2025lsm}, PedSleepMAE~\citep{pandey2024pedsleepmae}, and SleepGPT~\citep{huang2026unified}.
We pre-train PedSleepMAE on \benchname, evaluate SleepGPT using its released checkpoint, and adapt the remaining methods to our benchmark following their original settings when applicable.
As shown in Table~\ref{tab:full_domain_specific_baselines}, \ours consistently achieves the strongest performance on all of the downstream tasks on MROS dataset.

\textbf{\ours improves in-domain disease prediction.}
We further evaluate in-domain disease prediction on SHHS using two additional outcomes, \textit{hypertension} and \textit{diabetes}.
As shown in Table~\ref{tab:shhs_disease}, \ours consistently outperforms the compared baselines, further supporting the effectiveness of our pre-training strategy for clinically relevant downstream prediction tasks.

\begin{table*}[t]
\centering
\caption{\textbf{Confidence intervals for sleep staging and sleep event detection.} OSF achieves the best overall performance among all compared methods with statistical significance.}
\label{tab:mros_dense_task_ci}
\vspace{-4pt}
\begin{adjustbox}{width=\textwidth}
\scriptsize
\setlength{\tabcolsep}{3pt}
\begin{tabular}{l *{4}{cc}}
\toprule[1.5pt]
\multirow{2}{*}{\textbf{Method}} &
\multicolumn{2}{c}{\textbf{Sleep Staging}} &
\multicolumn{2}{c}{\textbf{Arousal}} &
\multicolumn{2}{c}{\textbf{Hypopnea}} &
\multicolumn{2}{c}{\textbf{Ox. Desat.}} \\
\cmidrule(lr){2-3}\cmidrule(lr){4-5}\cmidrule(lr){6-7}\cmidrule(lr){8-9}
& AUC$^\uparrow$ & AUPRC$^\uparrow$
& AUC$^\uparrow$ & AUPRC$^\uparrow$
& AUC$^\uparrow$ & AUPRC$^\uparrow$
& AUC$^\uparrow$ & AUPRC$^\uparrow$ \\
\midrule
\midrule

\textsc{SleepFM} &
96.4 $\pm$ 0.04 & 87.7 $\pm$ 0.15 &
\underline{90.3 $\pm$ 0.12} & \underline{84.9 $\pm$ 0.17} &
75.5 $\pm$ 0.20 & 60.6 $\pm$ 0.17 &
80.9 $\pm$ 0.14 & 80.3 $\pm$ 0.15 \\

\textsc{SimCLR} &
81.4 $\pm$ 0.11 & 56.8 $\pm$ 0.21 &
73.6 $\pm$ 0.21 & 67.8 $\pm$ 0.21 &
58.4 $\pm$ 0.25 & 53.1 $\pm$ 0.10 &
66.8 $\pm$ 0.18 & 65.0 $\pm$ 0.17 \\

\textsc{DINO} &
93.6 $\pm$ 0.06 & 81.4 $\pm$ 0.18 &
82.2 $\pm$ 0.17 & 75.7 $\pm$ 0.21 &
72.7 $\pm$ 0.22 & 59.5 $\pm$ 0.16 &
78.7 $\pm$ 0.14 & 78.2 $\pm$ 0.15 \\

\textsc{VQ-VAE} &
95.8 $\pm$ 0.05 & 86.7 $\pm$ 0.16 &
86.2 $\pm$ 0.15 & 80.3 $\pm$ 0.20 &
75.1 $\pm$ 0.21 & 60.9 $\pm$ 0.17 &
79.9 $\pm$ 0.14 & 79.5 $\pm$ 0.15 \\

\textsc{MAE} &
\underline{96.5 $\pm$ 0.04} & \underline{88.7 $\pm$ 0.14} &
89.7 $\pm$ 0.12 & 84.3 $\pm$ 0.17 &
\underline{77.6 $\pm$ 0.19} & \underline{61.9 $\pm$ 0.17} &
\underline{81.2 $\pm$ 0.13} & \underline{80.8 $\pm$ 0.14} \\

\textsc{Autoregression} &
96.0 $\pm$ 0.04 & 87.6 $\pm$ 0.15 &
87.8 $\pm$ 0.13 & 81.9 $\pm$ 0.18 &
72.1 $\pm$ 0.22 & 58.9 $\pm$ 0.15 &
79.0 $\pm$ 0.14 & 78.5 $\pm$ 0.15 \\

\rowcolor{gray!15}
\textsc{\ours} &
\textbf{97.3 $\pm$ 0.04} & \textbf{90.4 $\pm$ 0.14} &
\textbf{92.8 $\pm$ 0.10} & \textbf{88.3 $\pm$ 0.15} &
\textbf{77.7 $\pm$ 0.19} & \textbf{62.2 $\pm$ 0.18} &
\textbf{81.5 $\pm$ 0.13} & \textbf{81.0 $\pm$ 0.15} \\

\bottomrule[1.5pt]
\end{tabular}
\end{adjustbox}
\vspace{2pt}
\end{table*}

\begin{wraptable}{r}{0.5\columnwidth}
  \centering
    \caption{\textbf{Additional sleep staging evaluation metrics.} \ours remains the best.}
  \label{tab:stage_metrics}
  \vspace{-2mm}
  \begin{adjustbox}{width=\linewidth}
  \footnotesize
  \setlength{\tabcolsep}{4pt}
  \begin{tabular}{l ccc}
    \toprule[1.5pt]
    \textbf{Method} & \textbf{Acc. (micro)} & \textbf{F1 (weighted)} & \textbf{Cohen Kappa} \\
    \midrule
    \midrule
    \textsc{SleepFM} & 85.2 & 85.0 & \underline{79.9} \\
    \textsc{SimCLR}  & 63.2 & 59.5 & 25.2 \\
    \textsc{DINO}    & 79.4 & 78.8 & 63.9 \\
    \textsc{VQ-VAE}  & 83.6 & 83.4 & 74.6 \\
    \textsc{MAE}     & \underline{85.4} & \underline{85.2} & 79.6 \\
    \textsc{AR}      & 84.0 & 83.7 & 75.7 \\
    \rowcolor{gray!15}
    \textsc{\ours}   & \textbf{87.0} & \textbf{86.9} & \textbf{82.8} \\
    \bottomrule[1.5pt]
  \end{tabular}
  \end{adjustbox}
  \vspace{-2mm}
\end{wraptable}

\textbf{Confidence intervals.}
We additionally report confidence intervals for the main results to enable a more reliable comparison across methods. As shown in Table~\ref{tab:mros_dense_task_ci}, the confidence intervals show that the performance gains of \ours over competing baselines are consistent across tasks, further supporting the robustness of our improvements.

\textbf{Additional evaluation metrics for sleep staging.}
Since several downstream tasks are label-imbalanced, we use AUROC and AUPRC as the primary threshold-independent metrics.
For completeness, we also report Accuracy and weighted F1 for sleep staging results shown in the main table, as shown in Table~\ref{tab:stage_metrics}.
The results are consistent with the main findings, with \ours remaining the strongest overall method.

\begin{table}[t]
\centering
\caption{\textbf{Robustness analysis on demographic subgroups.} \ours achieves better performance and deliver fairer outcome.}
\label{tab:subgroup_eval}
\vspace{-2mm}

\begin{minipage}[t]{0.48\linewidth}
\centering
\textbf{Subgroup of CFS}
\vspace{1mm}

\begin{adjustbox}{width=\linewidth}
\scriptsize
\setlength{\tabcolsep}{3pt}
\begin{tabular}{l l cc}
\toprule[1.5pt]
\textbf{Subgroup} & \textbf{Setting} & \textbf{Sleep Staging} & \textbf{Hypopnea} \\
\midrule
\midrule
\multirow{2}{*}{Young}

& \textsc{SleepFM} & 97.9 & \textbf{92.2} \\
& \textsc{\ours}    & \textbf{98.1} & 91.0 \\
\midrule
\multirow{2}{*}{Old}

& \textsc{SleepFM} & 96.8 & 83.6 \\
& \textsc{\ours}    & \textbf{97.6} & \textbf{85.3} \\
\midrule
\multirow{2}{*}{Disparity}

& \textsc{SleepFM} & 0.78 & 6.08 \\
& \textsc{\ours}    & \textbf{0.35} & \textbf{4.03} \\
\bottomrule[1.5pt]
\end{tabular}
\end{adjustbox}
\end{minipage}
\hfill
\begin{minipage}[t]{0.48\linewidth}
\centering
\textbf{Subgroup of MESA}
\vspace{1mm}

\begin{adjustbox}{width=\linewidth}
\scriptsize
\setlength{\tabcolsep}{3pt}
\begin{tabular}{l l cc}
\toprule[1.5pt]
\textbf{Subgroup} & \textbf{Setting} & \textbf{Sleep Staging} & \textbf{Hypopnea} \\
\midrule
\midrule
\multirow{2}{*}{Male}

& \textsc{SleepFM} & 94.6 & 74.5 \\
& \textsc{\ours}    & \textbf{95.5} & \textbf{77.4} \\
\midrule
\multirow{2}{*}{Female}

& \textsc{SleepFM} & 94.2 & 76.8 \\
& \textsc{\ours}    & \textbf{96.1} & \textbf{77.9} \\
\midrule
\multirow{2}{*}{Disparity}
& \textsc{SleepFM} & \textbf{0.28} & 1.63 \\
& \textsc{\ours}    & 0.42 & \textbf{0.35} \\
\bottomrule[1.5pt]
\end{tabular}
\end{adjustbox}
\end{minipage}

\vspace{-2mm}
\end{table}
\textbf{Demographic analysis.}
We further evaluate subgroup generalization within cohorts using age subgroups in CFS dataset and gender subgroups in MESA dataset.
For each pretrained model, we fine-tune and evaluate the downstream classifier separately within each subgroup.
As shown in Tables~\ref{tab:subgroup_eval}, \ours outperforms SleepFM in most subgroup-task settings and shows lower performance variation across demographic groups.
These results suggest that the benefits of \ours extend beyond cohort-level averages to subgroup-level evaluation within heterogeneous sleep cohorts.

\subsection{Additional Analysis Studies}
\label{appendix:main_ablation}

\textbf{Channel masking consistently improves downstream performance across cohorts and tasks.} Here, we report additional experiments on more downstream tasks and data cohorts. In Table~\ref{tab:appendix_ablation_on_aug}, we report results across 3 sleep disorder event detection tasks and sleep staging tasks, on both the out-of-domain dataset MROS and the in-domain dataset SHHS. Under these broader evaluation settings, we observe that channel masking remains a strong augmentation strategy, compared to crop or time-masking strategies. Other findings in Sec.~\ref{sec:what ssl works} and Sec.~\ref{sec:ablation_aug} also hold. These results suggest that encouraging channel-invariant representations is important for improving downstream performance.

\begin{table*}[t]
  \centering
  \caption{\textbf{Controlled study of masking strategies for self-supervised pretraining on sleep data.} We report linear-probing results on sleep analysis tasks (\textbf{AUC}) on SHHS and MROS datasets. Channel masking as an augmentation consistently improves representation quality.}
  \label{tab:appendix_motivation_augmentation}
  \begin{adjustbox}{width=\textwidth}
  \begin{tabular}{l cc | cccc cccc}
    \toprule[1.5pt]
    \textbf{Method} & \multicolumn{2}{c}{\textbf{Mask Strategy}} &
    \multicolumn{4}{c}{\textbf{SHHS }} &
    \multicolumn{4}{c}{\textbf{MROS }} \\
    \cmidrule(lr){2-3}\cmidrule(lr){4-7}\cmidrule(lr){8-11}
    & Time & Channel
  & Sleep Staging & Arousal & Hypopnea & Ox.\ Desat.
  & Sleep Staging & Arousal & Hypopnea & Ox.\ Desat. \\

    \midrule
    \midrule
    \multirow{3}{*}{\textsc{SimCLR}}
      & \cmark & \xmark & 86.1 & 71.9 & 64.9 & 62.9 & 81.4 & 73.6 & 58.4 & 66.8 \\
    & \xmark & \cmark & 96.5 & 90.1 & 79.0 & 75.5 & 94.8 & 86.5 & 73.3 & 79.1 \\
    \grayrow
    & \cmark & \cmark & \textbf{97.3} & \textbf{92.9} & \textbf{82.2} & \textbf{80.4}
               & \textbf{96.7} & \textbf{90.8} & \textbf{75.6} & \textbf{81.1} \\
    \midrule

    \multirow{3}{*}{\textsc{DINO}}
      & \cmark & \xmark & 95.1 & 87.0 & 78.3 & 76.5 & 93.6 & 82.2 & 72.7 & 78.6 \\
    & \xmark & \cmark & 97.4 & 93.9 & 82.8 & 80.1 & 96.4 & 90.5 & 77.1 & 80.3 \\
    \grayrow
    & \cmark & \cmark & \textbf{97.5} & \textbf{94.0} & \textbf{82.8} & \textbf{81.3}
               & \textbf{97.3} & \textbf{92.8} & \textbf{77.7} & \textbf{81.5} \\
    \bottomrule[1.5pt]
  \end{tabular}
  \end{adjustbox}
\end{table*}

\begin{table*}[t]
  \centering
  \caption{\textbf{Ablation studies between augmentation from the vision domain (crop) and our masking-based strategies.} We show linear probing results (\textbf{AUC$^\uparrow$}) here. Vision domain's augmentation works for sleep data, but not necessary to be added.}
  \label{tab:appendix_ablation_on_aug}
  \begin{adjustbox}{width=\textwidth}
  \begin{tabular}{l ccc | cccc cccc}
    \toprule[1.5pt]
    \textbf{Method} & \textbf{Crop} & \multicolumn{2}{c}{\textbf{Masking}} &
    \multicolumn{4}{c}{\textbf{SHHS }} &
    \multicolumn{4}{c}{\textbf{MROS }} \\
    \cmidrule(lr){3-4}\cmidrule(lr){5-8}\cmidrule(lr){9-12}
&  & Time & Channel
  & Sleep Staging & Arousal & Hypopnea & Ox.\ Desat.
  & Sleep Staging & Arousal & Hypopnea & Ox.\ Desat. \\

    \midrule
    \midrule
    \multirow{3}{*}{\textsc{SimCLR}}
      & \cmark & \xmark & \xmark & 95.2 & 86.7 & 79.2 & 77.1 & 94.7 & 83.8 & 75.0 & 80.2 \\
          & \xmark & \xmark & \cmark & 96.5 & 90.1 & 79.0 & 75.5 & 94.8 & 86.5 & 73.3 & 79.1 \\
    & \cmark & \cmark & \cmark & 97.1 & 91.9 & \textbf{83.4} & 80.4 & 96.6 & 89.7 & \textbf{79.3} & \textbf{81.5} \\
    \grayrow
    & \xmark & \cmark & \cmark & \textbf{97.3} & \textbf{92.9} & 82.2 & \textbf{80.4} & \textbf{96.7} & \textbf{90.8} & 75.6 & 81.1 \\
    \midrule

    \multirow{3}{*}{\textsc{DINO}}
      & \cmark & \xmark & \xmark & 95.6 & 88.1 & 81.0 & 78.5 & 95.4 & 85.7 & 77.2 & 80.6 \\
      & \xmark    & \xmark & \cmark & 97.4 & 93.9 & 82.8 & 80.1 & 96.4 & 90.5 & 77.1 & 80.3 \\
    & \cmark & \cmark & \cmark & 97.0 & 91.4 & 82.1 & 79.8 & 96.6 & 89.3 & 77.3 & 81.0 \\
    \grayrow
    & \xmark & \cmark & \cmark & \textbf{97.5} & \textbf{94.0} & \textbf{82.8} & \textbf{81.3} & \textbf{97.3} & \textbf{92.8} & \textbf{77.7} & \textbf{81.5} \\
    \bottomrule[1.5pt]
  \end{tabular}
  \end{adjustbox}
\end{table*}

\begin{wraptable}{r}{0.4\textwidth}
  \centering
  \caption{\textbf{Heart-rate prediction results on SHHS and MROS}. Reconstruction-based SSL objectives help the regression task.}
  \label{tab:hr_pred}
  \begin{adjustbox}{width=\linewidth}
  \begin{tabular}{l rr rr}
    \toprule[1.5pt]
    \multirow{2}{*}{\textbf{Method}} &
    \multicolumn{2}{c}{\textbf{MROS}} &
    \multicolumn{2}{c}{\textbf{SHHS}} \\
    \cmidrule(lr){2-3}\cmidrule(lr){4-5}
    & MAE$^\downarrow$ & RMSE$^\downarrow$
    & MAE$^\downarrow$ & RMSE$^\downarrow$ \\
    \midrule
    \midrule
    \textsc{SleepFM}        & 3.38 & 5.00 & 3.52 & 5.06 \\
    \textsc{SimCLR}         & 8.11 & 10.09 & 8.14 & 10.06 \\
    \textsc{DINO}           & 7.39 & 9.26 & 7.40 & 9.23 \\
    \textsc{VQ-VAE}         & 4.03 & 5.79 & 3.50 & 5.22 \\
    \textsc{MAE}            & \underline{2.33} & \underline{3.99} & \underline{2.04} & \underline{3.63} \\
    \textsc{AR}             & \textbf{2.17} & \textbf{3.91} & \textbf{1.89} & \textbf{3.56} \\
    \textsc{\ours}   & 3.22 & 4.94 & 3.42 & 4.94 \\
    \midrule
    \textsc{Mean}           & 7.97 & 10.30 & 7.97 & 10.15 \\
    \bottomrule[1.5pt]
  \end{tabular}
  \end{adjustbox}
\end{wraptable}

\textbf{Our identified pre-training recipes also apply to other pre-training methods.} To verify whether our pre-training recipes generalize to other self-supervised learning methods, we adopt our identified pre-training recipes to the original SimCLR objective, yielding an upgraded variant (\ours~(SimCLR)). We evaluate both SimCLR and \ours~(SimCLR) models on the in-domain SHHS cohort. As shown in Table~\ref{tab:simclr_ours_dense_task},
the upgraded SimCLR (\ours~(SimCLR)) consistently outperforms the original SimCLR by a large margin across all tasks.

\begin{table*}[t]
\centering
\caption{\textbf{Ablation on normalization strategies.} Per-night z-score normalization leads to better performance.}
\label{tab:normalization_ablation}
\vspace{-2mm}
\begin{adjustbox}{width=\textwidth}
\footnotesize
\setlength{\tabcolsep}{5pt}
\begin{tabular}{l cc cc cc cc}
\toprule[1.5pt]
\multirow{3}{*}{\textbf{Normalization}} &
\multicolumn{4}{c}{\textbf{SHHS}} &
\multicolumn{4}{c}{\textbf{MROS}} \\
\cmidrule(lr){2-5}\cmidrule(lr){6-9}
&
\multicolumn{2}{c}{\textbf{Sleep Staging}} &
\multicolumn{2}{c}{\textbf{Hypopnea}} &
\multicolumn{2}{c}{\textbf{Sleep Staging}} &
\multicolumn{2}{c}{\textbf{Hypopnea}} \\
\cmidrule(lr){2-3}\cmidrule(lr){4-5}\cmidrule(lr){6-7}\cmidrule(lr){8-9}
& AUC$^\uparrow$ & AUPRC$^\uparrow$
& AUC$^\uparrow$ & AUPRC$^\uparrow$
& AUC$^\uparrow$ & AUPRC$^\uparrow$
& AUC$^\uparrow$ & AUPRC$^\uparrow$ \\
\midrule
\midrule
Min-Max & 96.3 & 90.4 & 80.6 & 76.0 & 95.8 & 86.4 & 77.1 & 62.1 \\
Per-Night Z-Score & \textbf{97.5} & \textbf{93.2} & \textbf{82.8} & \textbf{78.1} & \textbf{97.3} & \textbf{90.4} & \textbf{77.7} & \textbf{62.2} \\
\bottomrule[1.5pt]
\end{tabular}
\end{adjustbox}
\vspace{-2mm}
\end{table*}
\textbf{Ablation on data normalization strategy.}
We further compare our per-night per-channel z-score normalization with min-max normalization, which is commonly used in prior physiological self-supervised learning studies.
Using the same model architecture and training protocol, we pre-train a variant with min-max normalized inputs and evaluate it on SHHS dataset and MROS dataset for sleep staging and hypopnea detection.
As shown in Table~\ref{tab:normalization_ablation}, per-night z-score normalization achieves better performance across both datasets and tasks, suggesting that it better handles cross-subject and cross-cohort variation in sleep signal scale.

\begin{figure}[t]
    \centering
    \begin{minipage}[t]{0.48\linewidth}
        \centering
        \includegraphics[width=\linewidth]{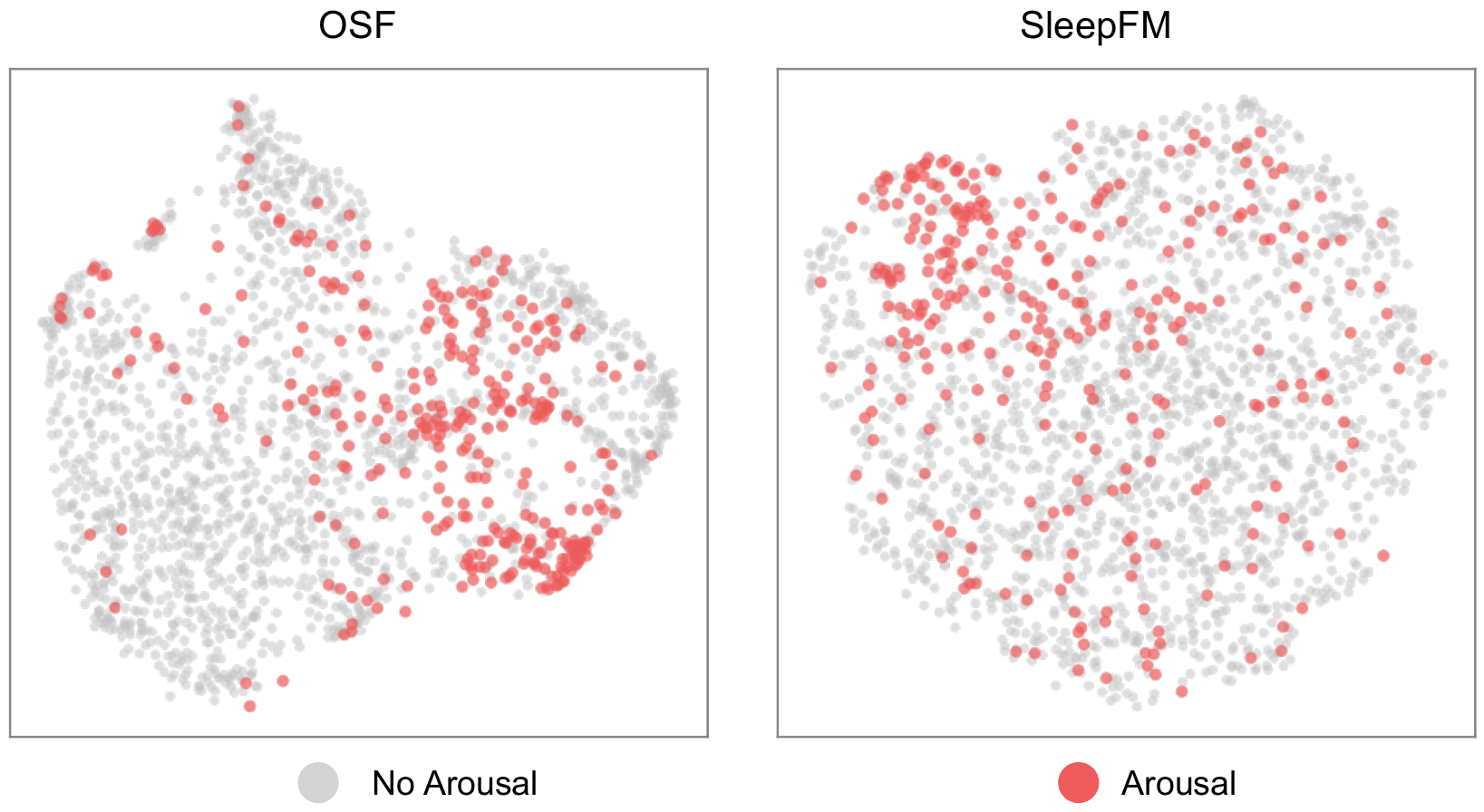}
        \vspace{-1mm}
        \centerline{\textbf{(a)} Arousal}
    \end{minipage}
    \hfill
    \begin{minipage}[t]{0.48\linewidth}
        \centering
        \includegraphics[width=\linewidth]{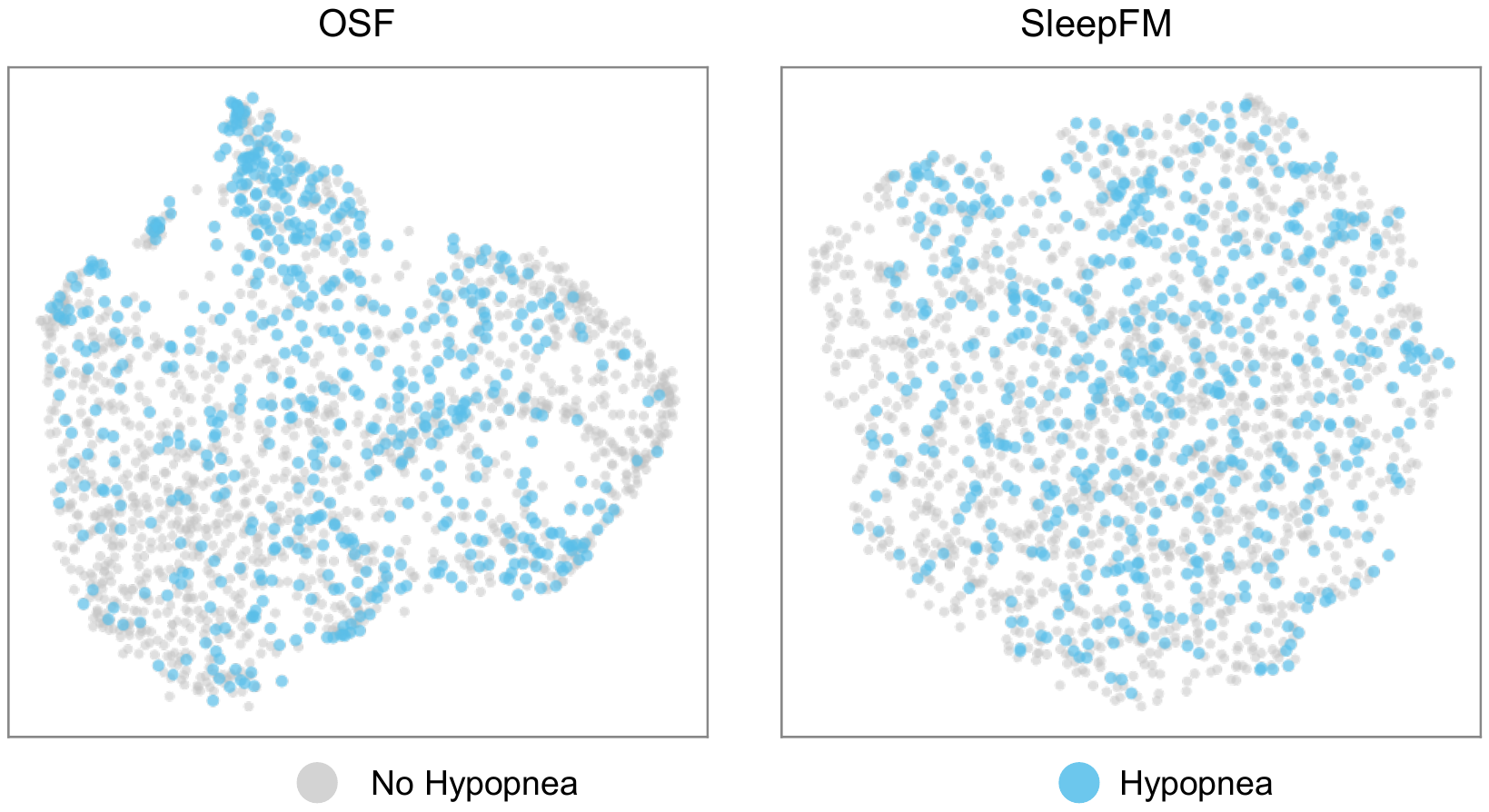}
        \vspace{-1mm}
        \centerline{\textbf{(b)} Hypopnea}
    \end{minipage}

    \caption{\textbf{UMAP visualization of learned representations.} \ours produces more separable embeddings, indicating stronger alignment between the embedding geometry and sleep event labels, including arousal and hypopnea events.}
    \label{fig:umap_events}
\end{figure}

\textbf{Additional UMAP visualizations on imbalanced downstream tasks.}
UMAP is computed from 2,000 randomly sampled epochs from the unseen SHHS test split.
We visualize embeddings for hypopnea and arousal events, which are substantially more imbalanced than sleep staging. As shown in Fig.~\ref{fig:umap_events}, although rare-event tasks show less clean separation, \ours generally exhibits clearer task-related structure than the baseline.

\label{appendix:other learnings}
\textbf{Reconstruction objectives transfer better on regression tasks on sleep data.}
As shown in Table~\ref{tab:mros_dense_task}, reconstruction-based methods such as MAE, VQ-VAE, and AR can be adapted to sleep signals with minimal modification and achieve reasonable performance; MAE even outperforms \sleepfm \ in many cases. In contrast, vanilla SimCLR and DINO lag behind both reconstruction-based methods and SleepFM. This indicates that invariance-based objectives require domain-appropriate augmentations to work well on sleep data. We also designed a heart-rate prediction task to test each pre-trained model's capability in regression, and found that MAE and AR achieve good performance, as shown in Table~\ref{tab:hr_pred}. 
\begin{figure}[t]
    \centering
    \includegraphics[width=1\linewidth]{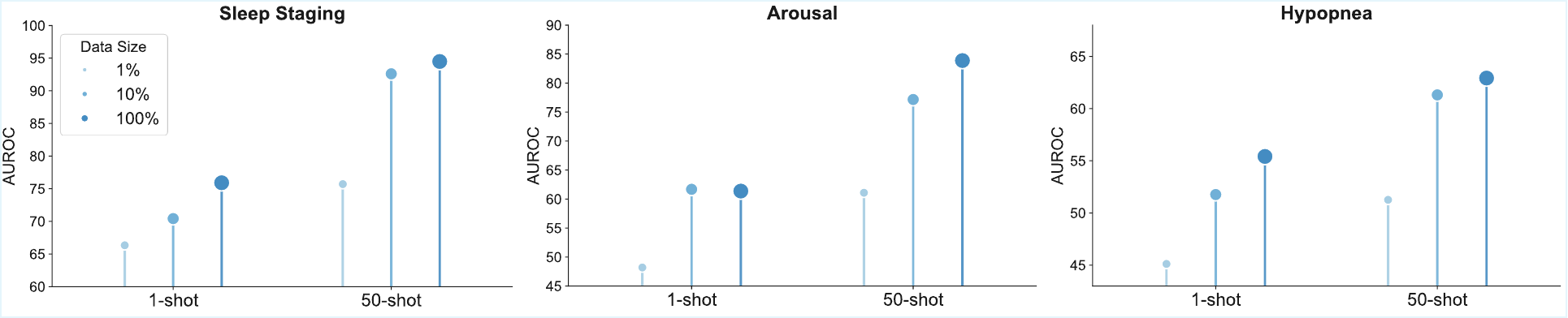}
    \caption{\textbf{Scaling performance on few-shot adaptation.} \ours scales on the pre-training sample size.}
    \label{fig:scaling_fewshot}
\end{figure}

\section{Additional Results on Scaling Analysis}
\label{appendix:scaling}
\subsection{Scaling across Pre-training Sample Scale and Model Capacity}
\label{appendix:data scaling}

\begin{table*}[t]
  \centering
  \caption{\textbf{Detailed data scaling results (linear probing) for \ours\  evaluated on \textbf{SHHS}.} All methods are trained with ViT-85M. Large model capacities or better pretraining strategies enable scalability.}
  \label{tab:appendix_scaling_ours_dino_shhs}
  \vspace{-2mm}
  \begin{adjustbox}{width=\textwidth}
  \footnotesize
  \begin{tabular}{@{}l r *{4}{cc}@{}}
    \toprule[1.5pt]
    \multirow{2}{*}{\textbf{Pretrain Set}} &
    \multirow{2}{*}{\textbf{Data Pct}} &
    \multicolumn{2}{c}{\textbf{Sleep Staging}} &
    \multicolumn{2}{c}{\textbf{Arousal}} &
    \multicolumn{2}{c}{\textbf{Hypopnea}} &
    \multicolumn{2}{c}{\textbf{Ox. Desat.}} \\
    \cmidrule(lr){3-4}\cmidrule(lr){5-6}\cmidrule(lr){7-8}\cmidrule(lr){9-10}
& & AUC$^\uparrow$ & AUPRC$^\uparrow$
  & AUC$^\uparrow$ & AUPRC$^\uparrow$
  & AUC$^\uparrow$ & AUPRC$^\uparrow$
  & AUC$^\uparrow$ & AUPRC$^\uparrow$ \\
    \midrule
    \midrule
    \textsc{single-src} & 1\%   & 91.8 & 79.8 & 74.9 & 67.2 & 70.0 & 65.2 & 68.5 & 62.7 \\
    \textsc{single-src} & 10\%  & 96.3 & 90.4 & 90.2 & 84.1 & 79.4 & 74.8 & 77.6 & 71.6 \\
    \textsc{single-src} & 100\% & 97.5 & 93.3 & 94.1 & 89.5 & 83.1 & 78.4 & 81.4 & 75.9 \\
    \midrule
    \textsc{multi-src}  & 1\%   & 92.8 & 81.9 & 78.0 & 69.9 & 72.0 & 67.1 & 70.3 & 64.5 \\
    \textsc{multi-src}  & 10\%  & 96.8 & 91.6 & 91.9 & 86.3 & 80.2 & 75.4 & 78.8 & 73.1 \\
    \textsc{multi-src}  & 100\% & 97.5 & 93.2 & 94.0 & 89.3 & 82.8 & 78.1 & 81.3 & 75.6 \\
    \bottomrule[1.5pt]
  \end{tabular}
  \end{adjustbox}
  \vspace{-2mm}
\end{table*}

\begin{table*}[t]
  \centering
  \caption{\textbf{Detailed data scaling results (linear probing) for \ours\  evaluated on \textbf{MROS}.} All methods are trained with ViT-85M. Large model capacities or better pretraining strategies enable scalability.}
  \label{tab:appendix_scaling_ours_dino_mros}
  \vspace{-2mm}
  \begin{adjustbox}{width=\textwidth}
  \footnotesize
  \begin{tabular}{@{}l r *{4}{cc}@{}}
    \toprule[1.5pt]
    \multirow{2}{*}{\textbf{Pretrain Set}} &
    \multirow{2}{*}{\textbf{Data Pct}} &
    \multicolumn{2}{c}{\textbf{Sleep Staging}} &
    \multicolumn{2}{c}{\textbf{Arousal}} &
    \multicolumn{2}{c}{\textbf{Hypopnea}} &
    \multicolumn{2}{c}{\textbf{Ox. Desat.}} \\
    \cmidrule(lr){3-4}\cmidrule(lr){5-6}\cmidrule(lr){7-8}\cmidrule(lr){9-10}
& & AUC$^\uparrow$ & AUPRC$^\uparrow$
  & AUC$^\uparrow$ & AUPRC$^\uparrow$
  & AUC$^\uparrow$ & AUPRC$^\uparrow$
  & AUC$^\uparrow$ & AUPRC$^\uparrow$ \\

    \midrule
    \midrule
    \textsc{single-src} & 1\%   & 88.9 & 71.3 & 75.4 & 68.7 & 63.5 & 55.3 & 74.5 & 73.5 \\
    \textsc{single-src} & 10\%  & 94.8 & 84.5 & 84.7 & 78.3 & 71.3 & 58.7 & 78.1 & 77.6 \\
    \textsc{single-src} & 100\% & 97.0 & 89.5 & 92.1 & 87.4 & 75.9 & 60.9 & 80.8 & 80.4 \\
    \midrule
    \textsc{multi-src}  & 1\%   & 90.5 & 74.8 & 77.6 & 70.8 & 66.8 & 56.6 & 76.4 & 75.7 \\
    \textsc{multi-src}  & 10\%  & 96.1 & 87.8 & 88.4 & 82.8 & 73.9 & 60.1 & 79.8 & 79.5 \\
    \textsc{multi-src}  & 100\% & 97.3 & 90.4 & 92.8 & 88.3 & 77.7 & 62.2 & 81.5 & 81.0 \\
    \bottomrule[1.5pt]
  \end{tabular}
  \end{adjustbox}
  \vspace{-2mm}
\end{table*}

\begin{table*}[t]
  \centering
  \caption{\textbf{Detailed data scaling results (linear probing) for \sleepfm\ evaluated on \textbf{SHHS}.} All methods are trained with ViT-85M. Large model capacities or better pretraining strategies enable scalability.}
  \label{tab:appendix_scaling_sleepfm_shhs}
  \vspace{-2mm}
  \begin{adjustbox}{width=\textwidth}
  \footnotesize
  \begin{tabular}{@{}l r *{4}{cc}@{}}
    \toprule[1.5pt]
    \multirow{2}{*}{\textbf{Pretrain Set}} &
    \multirow{2}{*}{\textbf{Data Pct}} &
    \multicolumn{2}{c}{\textbf{Sleep Staging}} &
    \multicolumn{2}{c}{\textbf{Arousal}} &
    \multicolumn{2}{c}{\textbf{Hypopnea}} &
    \multicolumn{2}{c}{\textbf{Ox. Desat.}} \\
    \cmidrule(lr){3-4}\cmidrule(lr){5-6}\cmidrule(lr){7-8}\cmidrule(lr){9-10}
& & AUC$^\uparrow$ & AUPRC$^\uparrow$
  & AUC$^\uparrow$ & AUPRC$^\uparrow$
  & AUC$^\uparrow$ & AUPRC$^\uparrow$
  & AUC$^\uparrow$ & AUPRC$^\uparrow$ \\
    \midrule
    \midrule
    \textsc{single-src} & 1\%   & 93.7 & 84.0 & 84.2 & 76.0 & 76.3 & 71.2 & 74.4 & 68.7 \\
    \textsc{single-src} & 10\%  & 95.1 & 87.5 & 88.1 & 81.1 & 79.1 & 74.2 & 77.6 & 71.7 \\
    \textsc{single-src} & 100\% & 96.2 & 90.1 & 91.8 & 85.6 & 81.5 & 76.3 & 80.0 & 74.3 \\
    \midrule
    \textsc{multi-src}  & 1\%   & 94.3 & 85.7 & 84.0 & 75.5 & 76.8 & 72.0 & 75.3 & 69.8 \\
    \textsc{multi-src}  & 10\%  & 95.1 & 87.7 & 88.7 & 81.9 & 78.7 & 73.8 & 77.3 & 71.3 \\
    \textsc{multi-src}  & 100\% & 96.7 & 91.1 & 92.2 & 86.3 & 82.0 & 77.0 & 80.6 & 75.0 \\
    \bottomrule[1.5pt]
  \end{tabular}
  \end{adjustbox}
  \vspace{-2mm}
\end{table*}

\begin{table*}[t]
  \centering
  \caption{\textbf{Detailed data scaling results (linear probing) for SleepFM evaluated on \textbf{MROS}.} All methods are trained with ViT-85M. Large model capacities or better pretraining strategies enable scalability.}
  \label{tab:appendix_scaling_sleepfm_mros}
  \vspace{-2mm}
  \begin{adjustbox}{width=\textwidth}
  \footnotesize
  \begin{tabular}{@{}l r *{4}{cc}@{}}
    \toprule[1.5pt]
    \multirow{2}{*}{\textbf{Pretrain Set}} &
    \multirow{2}{*}{\textbf{Data Pct}} &
    \multicolumn{2}{c}{\textbf{Sleep Staging}} &
    \multicolumn{2}{c}{\textbf{Arousal}} &
    \multicolumn{2}{c}{\textbf{Hypopnea}} &
    \multicolumn{2}{c}{\textbf{Ox. Desat.}} \\
    \cmidrule(lr){3-4}\cmidrule(lr){5-6}\cmidrule(lr){7-8}\cmidrule(lr){9-10}
& & AUC$^\uparrow$ & AUPRC$^\uparrow$
  & AUC$^\uparrow$ & AUPRC$^\uparrow$
  & AUC$^\uparrow$ & AUPRC$^\uparrow$
  & AUC$^\uparrow$ & AUPRC$^\uparrow$ \\
    \midrule
    \midrule
    \textsc{single-src} & 1\%   & 91.2 & 77.2 & 80.7 & 73.9 & 69.1 & 57.9 & 78.3 & 77.9 \\
    \textsc{single-src} & 10\%  & 92.9 & 80.5 & 82.6 & 76.5 & 70.4 & 58.3 & 78.4 & 77.9 \\
    \textsc{single-src} & 100\% & 96.0 & 86.8 & 89.2 & 83.4 & 74.2 & 60.0 & 80.0 & 79.6 \\
    \midrule
    \textsc{multi-src}  & 1\%   & 92.3 & 79.1 & 81.2 & 74.6 & 68.3 & 57.4 & 77.8 & 77.3 \\
    \textsc{multi-src}  & 10\%  & 94.4 & 82.9 & 85.0 & 78.6 & 70.9 & 58.5 & 78.3 & 78.0 \\
    \textsc{multi-src}  & 100\% & 96.4 & 87.7 & 90.3 & 84.9 & 75.5 & 60.6 & 80.9 & 80.3 \\
    \bottomrule[1.5pt]
  \end{tabular}
  \end{adjustbox}
  \vspace{-2mm}
\end{table*}

\textbf{Scaling pre-training data leads to larger gains in few-shot adaptation.} Besides training with the full training split of the downstream task, we also tested the sample efficiency brought by scaling the pre-training sample size. Particularly, we run 1-shot, 5-shot, and 50-shot adaptation experiments to test whether increasing the pre-training sample size improves few-shot transfer. We conduct these experiments on MROS. As shown in Fig.~\ref{fig:scaling_fewshot}, in both the 1-shot and 50-shot settings, models pre-trained with more data consistently outperform those pre-trained with less data.

\textbf{Additional results on the effect of sample size scaling.} We pre-train \ours and SleepFM with a ViT-85M backbone using different fractions of the pre-training corpus. As shown in Tables~\ref{tab:appendix_scaling_ours_dino_mros} and~\ref{tab:appendix_scaling_ours_dino_shhs}, we conduct more comprehensive evaluations on sleep staging task and three sleep event detection tasks. The downstream performance of \ours improves consistently as we increase the percentage of pre-training data. We observe a similar scaling trend for SleepFM when using the same 85M backbone, as shown in Tables~\ref{tab:appendix_scaling_sleepfm_mros} and~\ref{tab:appendix_scaling_sleepfm_shhs}. The models pre-trained with larger parameter size demonstrate good scaling behaviors, which motivates us to jointly scale both model size and sample size. 

\begin{wrapfigure}{r}{0.35\columnwidth}
    \centering
    \vspace{-2mm}
    \includegraphics[width=\linewidth]{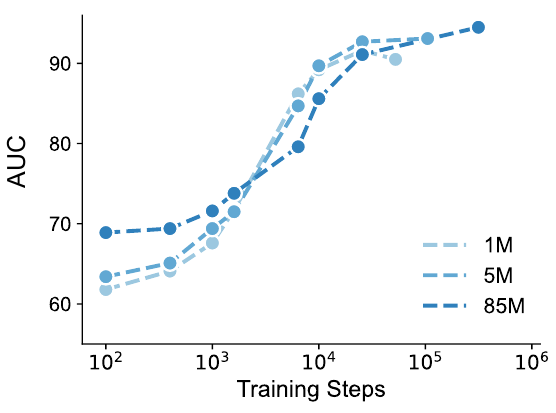}
    \caption{\textbf{Scaling on pretraining steps.} \ours scales on training computation.}
    \label{fig:scaling_on_step}
    \vspace{-2mm}
\end{wrapfigure}
\textbf{Single-source pre-training also exhibits scaling with sample size.} To further verify the generalization of scalability on sample size, we pre-train models on with different percentage of the SHHS cohort. As shown in Tables~\ref{tab:appendix_scaling_ours_dino_mros},  Tables~\ref{tab:appendix_scaling_ours_dino_shhs}, Tables~\ref{tab:appendix_scaling_sleepfm_mros}, and Tables~\ref{tab:appendix_scaling_sleepfm_shhs}, models pre-trained on the single-source SHHS cohort also show clear scaling behavior. When we increase the fraction of SHHS used for pre-training, downstream performance improves consistently.

\begin{figure}[!t]
\centering
\begin{minipage}[t]{0.54\columnwidth}
    \centering
    \includegraphics[width=\linewidth, trim={0cm 0cm 0cm 0cm}, clip]{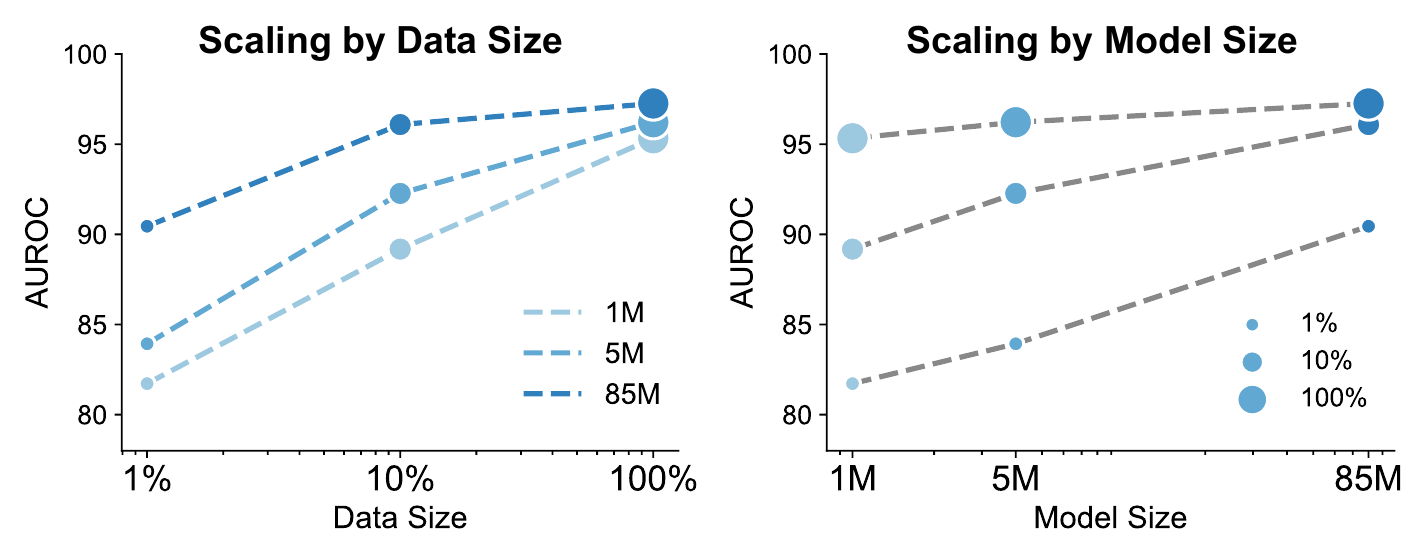}
    \caption{\textbf{Scaling performance on linear probing sleep staging.} \ours scales on both the model capacity and the pre-training sample size. }
    \label{fig:scaling-ours_stage}
\end{minipage}\hfill
\begin{minipage}[t]{0.44\columnwidth}
    \centering
    \includegraphics[width=\linewidth, trim={0cm 0cm 0cm 0cm}, clip]{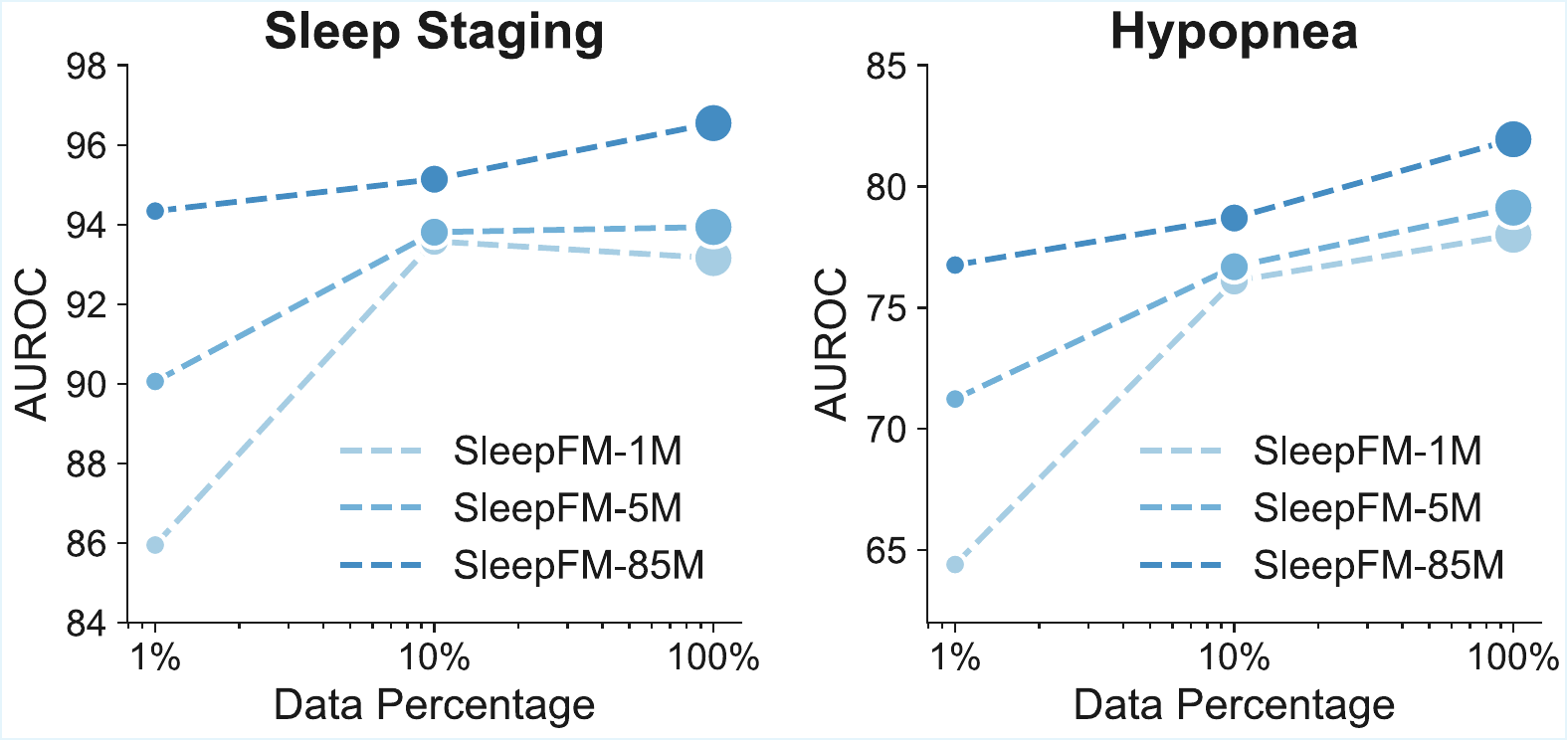}
    \caption{\textbf{Improved scaling performance of SleepFM.} Increasing model capacity helps weaker pre-training methods utilize more training samples effectively.}
    \label{fig:sleepfm_scale_by_improve_model_size}
\end{minipage}
\vspace{-8pt}
\end{figure}

\begin{table*}[t]
\centering
\caption{\textbf{Ablation on channel masking ratios.} A moderate masking ratio works the best.}
\label{tab:mask_ratio_ablation_full}
\vspace{-2mm}
\begin{adjustbox}{width=\textwidth}
\footnotesize
\setlength{\tabcolsep}{5pt}
\begin{tabular}{l cc cc cc cc}
\toprule[1.5pt]
\multirow{3}{*}{\textbf{Masking Ratio}} &
\multicolumn{4}{c}{\textbf{SHHS}} &
\multicolumn{4}{c}{\textbf{MROS}} \\
\cmidrule(lr){2-5}\cmidrule(lr){6-9}
&
\multicolumn{2}{c}{\textbf{Sleep Staging}} &
\multicolumn{2}{c}{\textbf{Hypopnea}} &
\multicolumn{2}{c}{\textbf{Sleep Staging}} &
\multicolumn{2}{c}{\textbf{Hypopnea}} \\
\cmidrule(lr){2-3}\cmidrule(lr){4-5}\cmidrule(lr){6-7}\cmidrule(lr){8-9}
& AUC$^\uparrow$ & AUPRC$^\uparrow$
& AUC$^\uparrow$ & AUPRC$^\uparrow$
& AUC$^\uparrow$ & AUPRC$^\uparrow$
& AUC$^\uparrow$ & AUPRC$^\uparrow$ \\
\midrule
\midrule
0.1 & 96.9 & 91.9 & 81.7 & 77.0 & 95.9 & 87.2 & 77.4 & 61.5 \\
\grayrow
\textbf{0.5 (main setting)} & \textbf{97.5} & \textbf{93.2} & \textbf{82.8} & \textbf{78.1} & \textbf{97.3} & \textbf{90.4} & \textbf{77.7} & \textbf{62.2} \\
0.9 & 95.4 & 88.7 & 79.5 & 74.5 & 95.3 & 85.9 & 75.0 & 60.3 \\
\bottomrule[1.5pt]
\end{tabular}
\end{adjustbox}
\vspace{-2mm}
\end{table*}

\label{appendix:model scaling}
\begin{table*}[t]
  \centering
  \caption{\textbf{Model size scaling results (linear probing) on \textbf{SHHS} and \textbf{MROS}.} Large models show better performances.}
  \label{tab:appendix_mode_size_scaling}
  \vspace{-2mm}
  \begin{adjustbox}{width=\textwidth}
  \footnotesize
  \begin{tabular}{@{}l l l *{4}{cc}@{}}
    \toprule[1.5pt]
    \multirow{2}{*}{\textbf{Dataset}} &
    \multirow{2}{*}{\textbf{Model}} &
    \multirow{2}{*}{\textbf{Enc Size}} &
    \multicolumn{2}{c}{\textbf{Sleep Staging}} &
    \multicolumn{2}{c}{\textbf{Arousal}} &
    \multicolumn{2}{c}{\textbf{Hypopnea}} &
    \multicolumn{2}{c}{\textbf{Ox. Desat.}} \\
    \cmidrule(lr){4-5}\cmidrule(lr){6-7}\cmidrule(lr){8-9}\cmidrule(lr){10-11}
& & & AUC$^\uparrow$ & AUPRC$^\uparrow$
  & AUC$^\uparrow$ & AUPRC$^\uparrow$
  & AUC$^\uparrow$ & AUPRC$^\uparrow$
  & AUC$^\uparrow$ & AUPRC$^\uparrow$ \\

    \midrule
    \midrule
    \multirow{6}{*}{\textsc{SHHS}} &
      \multirow{3}{*}{\textsc{\ours}} &
      ViT-1M  & 96.2 & 90.2 & 89.2 & 82.7 & 78.1 & 72.9 & 76.7 & 70.5 \\
    & & ViT-5M  & 96.9 & 91.8 & 91.8 & 86.2 & 80.2 & 75.1 & 78.5 & 72.5 \\
    & & ViT-85M & 97.5 & 93.2 & 94.0 & 89.3 & 82.8 & 78.1 & 81.3 & 75.6 \\
    \cmidrule(lr){2-11}
    & \multirow{3}{*}{\textsc{SleepFM}} &
      ViT-1M  & 93.2 & 82.6 & 85.9 & 78.0 & 78.0 & 72.7 & 76.1 & 69.7 \\
    & & ViT-5M  & 93.9 & 84.4 & 89.4 & 82.6 & 79.1 & 73.1 & 77.9 & 71.8 \\
    & & ViT-85M & 96.6 & 91.1 & 92.2 & 86.3 & 82.0 & 77.0 & 80.6 & 75.0 \\

    \midrule

    \multirow{6}{*}{\textsc{MROS}} &
      \multirow{3}{*}{\textsc{\ours}} &
      ViT-1M  & 95.3 & 86.1 & 86.0 & 80.1 & 70.7 & 58.4 & 78.7 & 78.3 \\
    & & ViT-5M  & 96.2 & 88.1 & 88.9 & 83.6 & 72.5 & 59.3 & 79.5 & 79.1 \\
    & & ViT-85M & 97.3 & 90.4 & 92.8 & 88.3 & 77.7 & 62.2 & 81.5 & 81.0 \\
    \cmidrule(lr){2-11}
    & \multirow{3}{*}{\textsc{SleepFM}} &
      ViT-1M  & 92.0 & 75.8 & 81.6 & 75.2 & 68.3 & 57.2 & 76.1 & 69.7 \\
    & & ViT-5M  & 93.4 & 78.8 & 85.0 & 78.7 & 69.1 & 57.4 & 78.3 & 77.5 \\
    & & ViT-85M & 96.4 & 87.7 & 90.3 & 84.9 & 75.5 & 60.6 & 80.9 & 80.3 \\

    \bottomrule[1.5pt]
  \end{tabular}
  \end{adjustbox}
  \vspace{-2mm}
\end{table*}

\textbf{Comprehensive results on model-size scaling.} 
We pre-train \ours and SleepFM on the full pre-training corpus, with different parameters of the backbone: 1M, 5M, 85M. We then evaluate these checkpoints on both the in-domain SHHS cohort and the out-of-domain MROS cohort, on three sleep event detection tasks and the sleep staging task. As shown in Table~\ref{tab:appendix_mode_size_scaling}, the scaling trend holds across all tasks and both cohorts, therefore, increasing model size leads to improved downstream performance.

\subsection{Additional Results of Multi-Source Data Mixture}
\label{appendix:multi-source}

Here we report detailed results comparing multi-source data mixture pre-training with single-source pre-training. We pre-train two groups of models on SHHS only, as well as on data sampled from the full pre-training corpus. For each group, we consider using 1\%, 10\%, 100\% of the given pre-training cohorts. As shown in Tables~\ref{tab:appendix_scaling_ours_dino_mros}, Tables~\ref{tab:appendix_scaling_ours_dino_shhs}, Tables~\ref{tab:appendix_scaling_sleepfm_mros}, and Tables~\ref{tab:appendix_scaling_sleepfm_shhs}, across four sleep analysis tasks and three data scales, multi-source pre-training consistently outperforms single-source pre-training, over all data percentages.

\subsection{Jointly Scale Model and Sample Size is Needed}

We further examine how model capacity and pre-training sample size jointly influence downstream performance. We pre-train both \ours and the baseline \sleepfm with different model sizes, on different sample size, and evaluate them on four sleep analysis tasks. As shown in Fig.~\ref{fig:sleepfm_scale_by_improve_model_size}, we find that increasing model capacity can improve the utilization of additional pre-training data. In particular, while SleepFM with its original model capacity shows only marginal gains as the pre-training sample size grows, scaling its each backbone encoder to 85M parameters yields substantially larger improvements. Motivated by this observation, we adopt a ViT-85M backbone when benchmarking different pre-training strategies and pre-train on the full multi-source data.

\subsection{Scaling with Pre-training Computation}
We further study whether \ours benefits from increased pre-training computation.
Specifically, we evaluate few-shot sleep staging performance across different numbers of pre-training steps using models with 1M, 5M, and 85M parameters.
As shown in Fig.~\ref{fig:scaling_on_step}, all models improve rapidly during the first $10^4$ steps.
After that, smaller models tend to saturate, while the 85M model continues to improve with additional pre-training.
These results suggest that \ours can benefit from increased pre-training computation when paired with sufficient model capacity.

\section{Additional Results on Handling Missing Channels}

\label{appendix:missing channel inference}

\begin{table*}[t]
  \centering
  \caption{\textbf{Linear probing results on \textbf{SHHS} under realistic missing-channel settings.} \textsc{\ours}\ (\textsc{DINO}) is more robust to missing channels compared to existing sleep FM.}
  \label{tab:ablation_channel_setting_shhs}
  \vspace{-2mm}
  \begin{adjustbox}{width=\textwidth}
  \footnotesize
  \setlength{\tabcolsep}{5pt}
  \begin{tabular}{ccc l *{4}{cc}}
    \toprule[1.5pt]
    \multirow{2}{*}{\textbf{Brain}} & \multirow{2}{*}{\textbf{Resp.}} & \multirow{2}{*}{\textbf{ECG}}& \multirow{2}{*}{\textbf{Method}} &
    \multicolumn{2}{c}{\textbf{Sleep Staging}} &
    \multicolumn{2}{c}{\textbf{Arousal}} &
    \multicolumn{2}{c}{\textbf{Hypopnea}} &
    \multicolumn{2}{c}{\textbf{Ox. Desat.}} \\
    \cmidrule(lr){5-6}\cmidrule(lr){7-8}\cmidrule(lr){9-10}\cmidrule(lr){11-12}
    & & & & AUC$^\uparrow$ & AUPRC$^\uparrow$
      & AUC$^\uparrow$ & AUPRC$^\uparrow$
      & AUC$^\uparrow$ & AUPRC$^\uparrow$
      & AUC$^\uparrow$ & AUPRC$^\uparrow$ \\
    \midrule
    \midrule

    \multirow{2}{*}{\textbf{\cmark}} & \multirow{2}{*}{\textbf{\xmark}} & \multirow{2}{*}{\textbf{\xmark}} &
      \textsc{SleepFM} & 97.0 & 92.0 & 92.2 & 86.2 & 75.0 & 69.7 & 73.8 & 67.5 \\

    & & & \textsc{\ours} & 97.5 & 93.4 & 93.7 & 88.9 & 75.9 & 70.5 & 75.3 & 69.0 \\
    \midrule
        \multirow{2}{*}{\textbf{\xmark}} & \multirow{2}{*}{\textbf{\cmark}} & \multirow{2}{*}{\textbf{\cmark}} &
      \textsc{SleepFM} & 86.7 & 69.1 & 87.3 & 80.3 & 82.3 & 77.5 & 80.1 & 74.5 \\
    & & & \textsc{\ours} & 85.2 & 66.6 & 87.4 & 80.7 & 82.8 & 78.4 & 80.1 & 74.4 \\
 \midrule
    \multirow{2}{*}{\textbf{\cmark}} & \multirow{2}{*}{\textbf{\xmark}} & \multirow{2}{*}{\textbf{\cmark}}  &
      \textsc{SleepFM} & 96.7 & 91.3 & 92.3 & 86.3 & 76.1 & 70.6 & 76.4 & 70.6 \\

    & & & \textsc{\ours} & 97.5 & 93.3 & 94.0 & 89.2 & 76.4 & 70.9 & 76.7 & 70.8 \\
    \midrule

    \multirow{2}{*}{\textbf{\xmark}} & \multirow{2}{*}{\textbf{\cmark}} & \multirow{2}{*}{\textbf{\xmark}}  &
      \textsc{SleepFM} & 86.6 & 68.7 & 87.0 & 79.8 & 82.8 & 78.2 & 80.1 & 74.3 \\

   & & & \textsc{\ours} & 85.5 & 67.0 & 86.9 & 80.2 & 84.1 & 80.3 & 80.3 & 74.7 \\

    \bottomrule[1.5pt]
  \end{tabular}
  \end{adjustbox}
  \vspace{-2mm}
\end{table*}

\textbf{\ours better handles missing channel inference problems on in-domain data corpus as well.} To validate the robustness of the conclusion in Sec.~\ref{sec:missing channel}, We report additional results for inference on the SHHS cohort on the four realistic missing channel settings. As shown in Table~\ref{tab:ablation_channel_setting_shhs}, across four sleep analysis tasks, \ours consistently outperforms \sleepfm, indicating that \ours learns more robust representations under inference-time channel missingness.

\label{appendix:pretrain with less channel}
\textbf{Additional results on pre-training with fewer channels.} Here we report detailed results for pre-training only with breathing channels and ECG channel. We compare \ours against SleepFM under the same channel subset. As shown in Table~\ref{tab:appendix_monitor_type_ablation}, across both the in-domain SHHS cohort and the out-of-domain MROS cohort, on most sleep analysis tasks, \ours makes better use of the available channels and achieves higher downstream performance.

\begin{table*}[t]
  \centering
  \caption{\textbf{Robustness of pre-training with fewer channels.} We pre-train the two models and evaluate them using only ECG and respiratory signals, \textsc{\ours} better utilizes these channels.}
  \label{tab:appendix_monitor_type_ablation}
  \begin{adjustbox}{width=\textwidth}
  \begin{tabular}{l l cc cc cc cc}
    \toprule[1.5pt]
    \multirow{2}{*}{\textbf{Dataset}} &
    \multirow{2}{*}{\textbf{Method}} &
    \multicolumn{2}{c}{\textbf{Sleep Staging}} &
    \multicolumn{2}{c}{\textbf{Arousal}} &
    \multicolumn{2}{c}{\textbf{Hypopnea}} &
    \multicolumn{2}{c}{\textbf{Ox.\ Desat.}} \\
    \cmidrule(lr){3-4}\cmidrule(lr){5-6}\cmidrule(lr){7-8}\cmidrule(lr){9-10}
    & &
AUC$^\uparrow$ & AUPRC$^\uparrow$ &
AUC$^\uparrow$ & AUPRC$^\uparrow$ &
AUC$^\uparrow$ & AUPRC$^\uparrow$ &
AUC$^\uparrow$ & AUPRC$^\uparrow$ \\

    \midrule
    \midrule
    \multirow{2}{*}{SHHS}
    & \textsc{SleepFM} & 82.1 & 61.1 & 84.4 & 77.1 & 79.5 & 74.5 & 78.1 & 72.6 \\
    & \textsc{\ours}          & 81.5 & 60.4 & 84.8 & 77.8 & 81.5 & 77.0 & 78.6 & 73.3 \\

    \midrule

    \multirow{2}{*}{MROS}
    & \textsc{SleepFM} & 83.2 & 59.0 & 82.7 & 76.8 & 71.3 & 58.6 & 79.7 & 78.9 \\
    & \textsc{\ours}          & 83.2 & 59.4 & 83.3 & 77.6 & 77.6 & 62.4 & 80.8 & 80.1 \\

    \bottomrule[1.5pt]
  \end{tabular}
  \end{adjustbox}
\end{table*}

\section{Additional Ablation Results}

\textbf{Additional results of ablation studies on channel masking ratio.} 
We further ablate the channel masking ratio by pre-training two additional variants with masking ratios of 0.1 and 0.9, while keeping all other settings unchanged.
As shown in Table~\ref{tab:mask_ratio_ablation_full}, the default ratio of 0.5 achieves the best overall linear probing performance across SHHS and MROS.
This suggests that moderate channel masking provides a better balance between task difficulty and preserved input information: weak masking may make the pretext task less informative, whereas overly aggressive masking can remove too much training signal and hurt downstream transfer.

\end{document}